\documentclass[sigconf]{acmart}
\AtBeginDocument{%
  }

\setcopyright{rightsretained}
\copyrightyear{2026}

\acmYear{2026}
\acmDOI{10.1145/3786335.3813149}

\acmConference[ACM CAIS '26]{ACM Conference on AI and Agentic Systems}{May 26--29, 2026}{San Jose, CA, USA}
\acmBooktitle{ACM Conference on AI and Agentic Systems (ACM CAIS '26), May 26--29, 2026, San Jose, CA, USA}
\acmISBN{979-8-4007-2415-2/26/05}

\newcommand{\best}[1]{{%
  \setlength{\fboxsep}{1.5pt}%
  \setlength{\fboxrule}{1.2pt}%
  \fcolorbox{black!60!black}{white!8}{\textbf{#1}}%
}}

\newcommand{\worst}[1]{\setlength{\fboxsep}{1pt}\fcolorbox{red!60!black}{white!8}{#1}}
\newcommand{\up}{\ensuremath{\,{\scriptstyle\uparrow}}}
\newcommand{\dn}{\ensuremath{\,{\scriptstyle\downarrow}}}
\newcommand{\ck}{\ensuremath{\,{\scriptstyle\checkmark}}}

\begin{document}

\title{Context, Reasoning, and Hierarchy: A Cost-Performance Study of Compound LLM Agent Design in an Adversarial POMDP}

\author{Igor Bogdanov}
\email{igorbogdanov@cmail.carleton.ca}
\orcid{0009-0008-6606-189X}
\affiliation{%
  \institution{Carleton University}
  \city{Ottawa}
  \state{Ontario}
  \country{Canada}
}

\author{Chung-Horng Lung}
\email{chlung@sce.carleton.ca}
\orcid{0000-0002-5662-490X}
\affiliation{%
  \institution{Carleton University}
  \city{Ottawa}
  \state{Ontario}
  \country{Canada}
}

\author{Thomas Kunz}
\email{tkunz@sce.carleton.ca}
\orcid{0000-0002-6241-778X}
\affiliation{%
  \institution{Carleton University}
  \city{Ottawa}
  \state{Ontario}
  \country{Canada}
}

\author{Jie Gao}
\email{jie.gao6@carleton.ca}
\orcid{0000-0001-6095-2968}
\affiliation{%
  \institution{Carleton University}
  \city{Ottawa}
  \state{Ontario}
  \country{Canada}
}

\author{Adrian Taylor}
\email{Adrian.Taylor@forces.gc.ca}
\orcid{0000-0002-3785-6270}
\affiliation{%
  \institution{Defence R\&D Canada}
  \city{Ottawa}
  \state{Ontario}
  \country{Canada}
}

\author{Marzia Zaman}
\email{Marzia@cistel.com}
\orcid{0000-0002-0610-0470}
\affiliation{%
  \institution{Cistel Technology}
  \city{Ottawa}
  \state{Ontario}
  \country{Canada}
}

\renewcommand{\shortauthors}{Bogdanov et al.}

\begin{abstract}
  Deploying compound LLM agents in adversarial, partially observable sequential environments requires navigating several interacting design dimensions: (1) what the agent sees, (2) how it reasons, and (3) how tasks are decomposed across components. Yet practitioners lack guidance on which design choices improve performance versus merely increase inference costs. We present a controlled study of compound LLM agent design in CybORG CAGE-2, a cyber defense environment modeled as a Partially Observable Markov Decision Process (POMDP). Reward is non-positive, so all configurations operate in a failure-mitigation mode and errors compound over time. Our evaluation spans five model families, six models, and twelve configurations (3,475 episodes) with token-level cost accounting. We systematically vary context representation (raw observations vs. a deterministic, programmatic environment state-tracking layer with compressed history), deliberation (self-questioning, self-critique, and self-improvement tools, with optional chain-of-thought prompting), and hierarchical decomposition (monolithic ReAct vs. delegation to specialized sub-agents). We find that: (1) Programmatic state abstraction delivers the largest returns per token spent (RPTS), improving mean return by up to 76\% over raw observations. (2) Distributing deliberation tools across a hierarchy degrades performance relative to hierarchy alone for all five model families, reaching up to 3.4$\times$ worse mean return while using 1.8-2.7$\times$ more tokens. We call this destructive interaction pattern a \emph{deliberation cascade}. (3) Hierarchical decomposition without deliberation tools achieves the best absolute performance for most models, and context engineering is generally more cost-effective than deliberation. These findings suggest a design principle for structured adversarial POMDPs: invest in programmatic infrastructure and clean task decomposition rather than deeper per-agent reasoning, as these strategies can interfere when combined.

\end{abstract}



\maketitle

\section{Introduction}\label{sec:intro}

Practitioners build compound LLM agents by composing three design dimensions: \emph{context engineering} \cite{karpathy2025} (what the agent sees), \emph{deliberation} (reasoning depth), and \emph{hierarchical decomposition} (task distribution). While often assumed to be additive, these choices interact destructively in adversarial, partially observable sequential environments. We present a controlled empirical study of these interactions within a compound LLM agent defending a network in the CybORG CAGE-2 POMDP. By systematically ablating context, reasoning, and hierarchy across five model families and six models, we measure both task performance and token cost, revealing that \emph{what the agent sees} is a more reliable lever than \emph{how deeply it thinks}: deterministic programmatic context yields large gains at near-zero marginal cost, whereas distributing deliberation across a hierarchy often degrades performance while inflating token consumption. We term this failure mode a \emph{deliberation cascade}. While bounded hierarchy often achieves the best \emph{absolute} return, context engineering delivers the best \emph{returns per token}, making it the most cost-effective first investment.

\paragraph{The Empirical Gap} Three gaps motivate this work. (1) Multi-agent research emphasizes \emph{topology} (wiring) over internal agent design~\cite{kim2025}, leaving the interaction between internal configuration and hierarchy underexplored. (2) While context engineering is widely endorsed~\cite{karpathy2025}, controlled evidence on the cost-performance trade-offs of specific context components in sequential settings is limited. (3) The deliberation and multi-agent literatures remain disconnected, and we show that combining them can introduce failure modes invisible to either in isolation.

\paragraph{Research Question} Our central question is: \textbf{which compound-agent design dimensions deliver performance gains per token invested, and how do they interact when composed across a multi-agent hierarchy?} We decompose this into: \textbf{RQ1} (Context): value of programmatic abstraction vs.\ raw observations; \textbf{RQ2} (Reasoning): impact of deliberation tools in monolithic vs.\ hierarchical agents; and \textbf{RQ3} (Composition): when decomposition helps or hurts.

\paragraph{Contributions} We present four contributions at the intersection of architectural design and system optimization. (1) \textbf{Cost-effective context engineering.} We show that a deterministic state-tracking layer reduces cumulative penalty by 52-76\% relative to raw observations for four of six models, dominating raw observation context configurations at near-zero marginal cost. (2) \textbf{Identification of deliberation cascades.} We demonstrate that enabling deliberation tools across a hierarchy degrades performance in all six models (up to $3.4{\times}$ worse return) while doubling token costs, producing cascading uncertainty. (3) \textbf{Three-axis Pareto analysis.} We conduct a controlled ablation of context, reasoning, and hierarchy across five model families (72~pairs, 3,475~episodes), constructing cost-performance frontiers that consistently place programmatic context on the efficient frontier. (4) \textbf{Multi-model validation.} We show that while qualitative effects (context helps, distributed reasoning hurts) are robust, quantitative magnitudes vary, validating multi-model evaluation as essential for compound AI design.

\paragraph{Scope} This paper is a static architectural design-space study: it asks what compound-agent architecture to build at deployment time, before any runtime adaptation. We scope our claims to structured adversarial POMDPs.

\section{Background \& Problem Definition}\label{sec:background}
We study compound LLM agents operating in an adversarial, partially observable sequential decision problem and evaluate designs jointly on (1)~task return and (2)~token cost. The agent architecture and the multi-dimensional ablation study mirror our engineering ladder for solving CAGE-2 with an LLM agent: we began with raw environment observations, then introduced deterministic state tracking and context engineering to make the observations actionable, decomposed the task into a hierarchy as context grew, and finally added deliberation tools to facilitate better decision-making within each agent. This section defines the environment and objectives, specifies the execution scaffold and initialization scope, and establishes tokens as the cost primitive.

\paragraph{CybORG CAGE-2}\label{sec:cage2}
We evaluate on CybORG CAGE-2~\cite{cage2,standen2021cyborg}, an adversarial POMDP modelling network defense. A defender protects a 13-host network against an attacker that follows a scripted, non-adaptive multi-stage kill chain~\cite{kiely2023cage} over $T{=}30$ steps. The attacker does not respond to the defender's actions, but host attributes, processes, and the progression of the kill chain vary stochastically across runs. The defender chooses from five actions (\textsc{Monitor}, \textsc{Analyse}, \textsc{Remove}, \textsc{Restore}, \textsc{Decoy}) with asymmetric costs. Reward $r_t \le 0$ penalizes compromise and intervention; we report episodic return $G=\sum_{t=1}^{T} r_t$ (closer to zero is better). Partial observability requires sustaining situational awareness under noisy indicators. (See Appendix~\ref{app:cage2} for full details.)

\paragraph{Configuration-Driven ReAct Scaffold}\label{sec:react}
Agents follow a ReAct~\cite{yao2023react} loop, iterating between deliberation and tool use until emitting an answer. We separate a reusable ReAct Agent engine (I/O, parsing) from declarative YAML "personality" configurations (prompts, tools). This separation ensures reproducibility and allows architectural variants to be defined as configuration changes rather than code modifications.
\paragraph{Knowledge-Free Initialization}\label{sec:knowledge-free}
To isolate architectural effects from domain engineering, we impose a knowledge-free scope at $t{=}0$. Agents receive only a one-sentence role instruction and a compact action reference table. No network topology, attacker kill-chain details, host-value assignments, or defense heuristics are provided, and the prompt never mentions the environment by name. Performance gains arise from model internal knowledge, runtime context, and deliberation. Models' internal knowledge may include information about CAGE-2 acquired during pre-training. To minimize the consequences of this exposure, no benchmark-specific cues are provided at runtime. However, influence from pre-training may still be present. All LLMs receive an identical set of prompts and tools.

\paragraph{Tokens as Cost Primitive}\label{sec:tokens}
We use total number of tokens per episode (prompt + completion) as the primary cost metric, mapping directly to billed usage and correlating with latency. We use raw token counts because they do not depend on provider pricing. Appendix~\ref{app:tokens} reports prompt/completion splits by model and configuration for pricing-weighted reinterpretation. We instrument all LLM calls, aggregating prompt and completion tokens across the Planner and sub-agents. Pairing return with token cost enables our central evaluative question: which design choices deliver the largest \textbf{returns per token spent} (RPTS), and which inflate the cost.

\begin{table*}[t]
    \caption{System Modularity and Configuration Layers. The agent architecture is built on four functional pillars, ordered from the system's top-level decision structure down to its internal reasoning mechanisms.}
    \label{tab:system-layers}
    \centering
    \small
    \begin{tabular}{l p{3.5cm} p{5.5cm} p{4.5cm}}
      \toprule
      \textbf{System Layer} & \textbf{Core Aspects} & \textbf{Responsibility} & \textbf{Configurability} \\
      \midrule
      \textbf{1. Hierarchy} & Planner, Analyst, ActionChooser, JSON Contracts & Distributes tasks: Planner (decides), Analyst (assesses), ActionChooser (ranks). & Role definitions, tool availability, and I/O contracts in YAML. \\
      \midrule
      \textbf{2. Infrastructure} & Env Model, State Machine, History Log, Action Validator & Maintains deterministic belief state (status, history, decoys) and validates actions via regex. & Fixed deterministic backbone (Python). \\
      \midrule
      \textbf{3. Context} & \texttt{\{network\_status\}}, \texttt{\{history\}}, \texttt{\{observation\}} & Transforms raw observations into structured summaries and compressed logs. & YAML templates define which context blocks are injected. \\
      \midrule
      \textbf{4. Reasoning} & ReAct Engine, Tools (Question, Critique, Improve), CoT & Executes the reasoning loop, tool calls, and optional deliberation steps. & Boolean flags toggle deliberation tools and CoT injection. \\
      \bottomrule
    \end{tabular}
  \end{table*}
\section{Compound Agent System Design}\label{sec:system}
Our system couples a deterministic backbone with an LLM decision engine and spans four layers (Table~\ref{tab:system-layers}). 
(1) The \textbf{hierarchy layer} allows a Planner to delegate tasks to Analyst and ActionChooser sub-agents via strict JSON contracts. (2) A \textbf{deterministic infrastructure layer} maintains a primitive environment model, history of interactions, and validates actions without model calls. (3) The \textbf{context engineering layer} comprises injections that extend user prompts and connect environment model state and action history by converting them into structured summaries like \texttt{\{network\_status\}}. (4) The \textbf{reasoning layer} follows the ReAct pattern~\cite{yao2023react}, optionally executing deliberation within ReAct loop.

\begin{figure}[htbp]
  \centering
  \includegraphics[width=\linewidth]{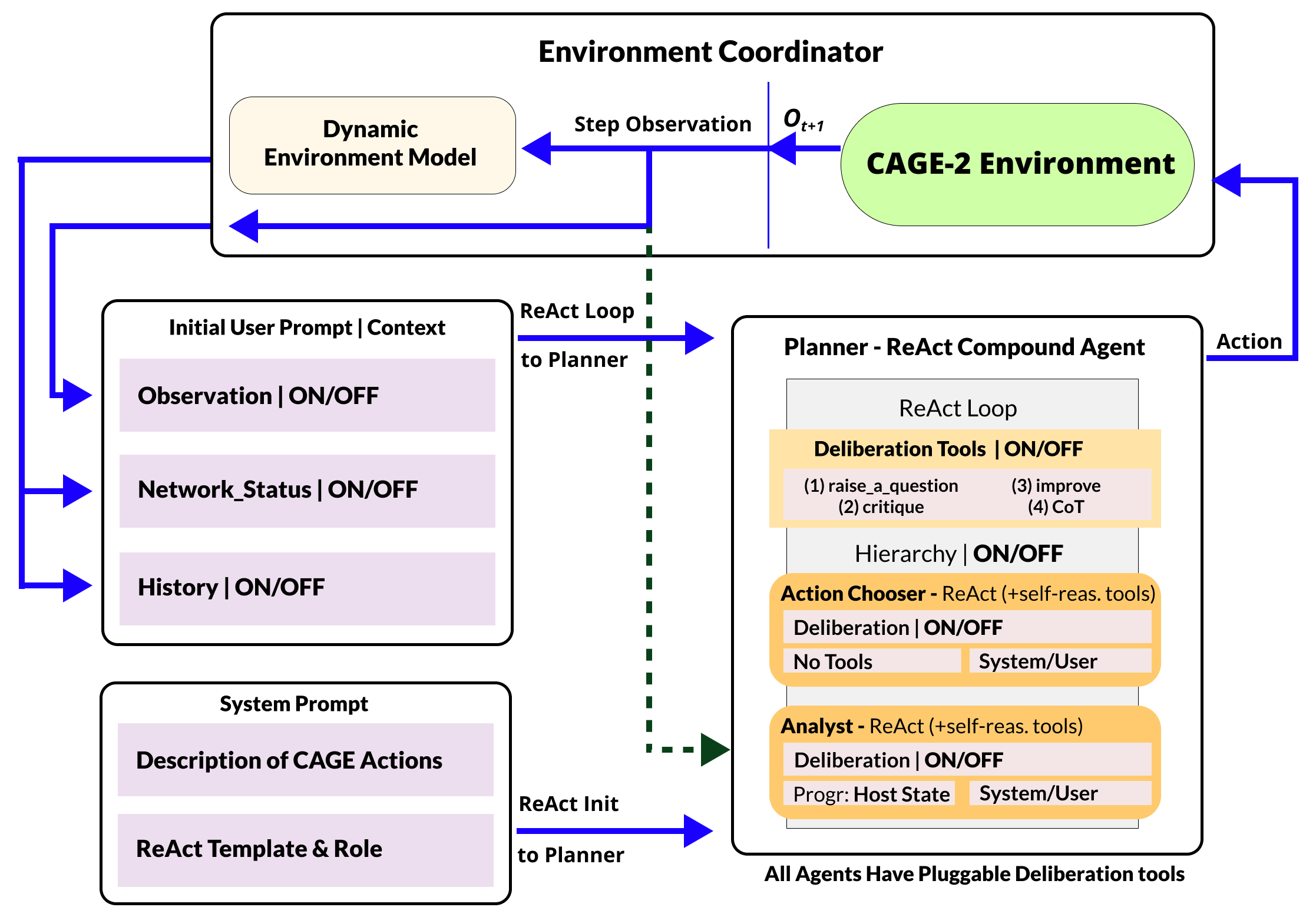}
  \caption{End-to-end system architecture. The deterministic layer (left) compiles structured context from CybORG observations and assembles the agent prompt. The Planner (right) executes a ReAct loop, optionally delegating to Analyst and ActionChooser sub-agents, before emitting a validated action back to the environment.}
  \Description{System diagram showing the deterministic infrastructure on the left (ReAct template, knowledge, prompt construction, network status and history) feeding into the Planner on the right, which contains deliberation tools and optional sub-agents (ActionChooser, Analyst). The Planner emits an action to the CAGE-2 environment, which returns the next observation.}
  \label{fig:system-architecture}
\end{figure}

\paragraph{Decision Cycle}\label{sec:decision-cycle}
At each of the $T{=}30$ steps, an Agent Coordinator receives the raw CybORG observation, updates the deterministic environment model, and inserts the chosen context bundle into the Planner prompt. It then starts \textbf{a fresh step-level Planner instance}, validates the emitted action, retries on invalid outputs, and submits the validated action to the environment (Figure~\ref{fig:system-architecture}). Step-level instantiation ensures no hidden conversational state accumulates across steps. All inter-step continuity is explicit in the deterministic state structures and compiled context.

\paragraph{Reliability Mechanisms}\label{sec:reliability}
In an adversarial sequential environment, every invalid action is a wasted step during which the attacker advances unopposed. We therefore treat reliability as load-bearing infrastructure implementing the following:
(1)~\textbf{Action validation and retry:} the Planner's output is parsed against the CybORG action schema using regex-based validation; invalid outputs trigger up to three retries with the parsing error injected as feedback.
(2)~\textbf{Safe fallback:} if all retries fail, the system defaults to \textsc{Monitor}, preserving observability without risking a misapplied intervention.
(3)~\textbf{Sub-agent output validation:} the ActionChooser's JSON undergoes tiered parsing (direct parse $\rightarrow$ normalization $\rightarrow$ optional repair prompt). Any repair calls are included in token accounting. These mechanisms ensure format compliance. They are designed to minimize harm when invoked.

\subsection{Layer 1: Hierarchical Decomposition}\label{sec:hierarchy-axis}
The system can run either as a \emph{monolithic Planner} that directly emits an environment action, or as a fixed three-agent hierarchy consistent with the centralized multi-agent system (MAS) topology described by Kim et al.~\cite{kim2025}, in which an orchestrator coordinates bounded sub-agents through structured communication. The three-agent split separates three functions: strategic decision-making, localized perception, and bounded action candidate ranking to accommodate increasing context size. The \textbf{Planner} retains sole authority over environment actions but may delegate to two sub-agents whose outputs are advisory. The \textbf{Analyst} assesses a single host by comparing its current state against the effective baseline and returns a structured JSON assessment (status, anomalies, suspected compromise). The \textbf{ActionChooser} receives the Planner's situational summary and, when available, the \textbf{Analyst's} report, returning a ranked list of up to three candidate actions with confidence labels in strict JSON. Sub-agents cannot issue environment actions. The purpose of this split is to simplify the Planner's decision by constraining what each sub-agent produces. All three agents share the same engine-personality separation logic, ensuring the observed hierarchy effects reflect role decomposition.

\subsection{Layer 2: Deterministic Environment Model}\label{sec:env-model}
To provide in-context situational awareness, the system maintains a fully deterministic environment model that transforms raw CybORG observations into structured network state description and action history. Critically, this model is computed exclusively from the agent's own observations and past actions. The system stores environment \emph{baseline state} upon the first observation and creates a host-indexed data structure. This layer embeds domain-informed observation processing. The engineering choices that form the model shape what the agent perceives. Although the layer prescribes no action-selection decisions (no threat rubrics, no host priorities, and no response heuristics), it still carries inductive bias that may influence the agent's reasoning.

\paragraph{Dynamic Environment Model} The model data structure comprises a dictionary where each host obtains a \textbf{status}: \{\texttt{baseline}, \texttt{changed}, \texttt{unknown}, \texttt{analysed at step~$n$}\}, and a \textbf{history record}, an ordered, arrow-delimited record of all actions applied to each host (e.g., \texttt{Analyse\,$\rightarrow$\,Remove\,$\rightarrow$\,Restore}). This concise single-host record provides the Planner with additional intervention memory without requiring it to parse a full transcript.

\paragraph{Model update mechanism.}
On each step, the model compares the current observation against the saved \emph{baseline state} using signature-based comparison of stable fields (process and service identity), ignoring volatile fields (e.g., transient connections) that produce false positives~\cite{kiely2023cage}. The system deterministically updates host status based on the comparison and prior actions (\textsc{Restore} $\rightarrow$ \texttt{baseline}; \textsc{Remove} $\rightarrow$ \texttt{unknown}; \textsc{Analyse} $\rightarrow$ \texttt{analysed at step~$n$}). Deployed decoys are registered as baseline overrides so that expected decoy processes are incorporated into the effective baseline and only genuinely new processes remain visible as anomalous.

\subsection{Layer 3: Context Engineering}\label{sec:context-axis}
\paragraph{Context Injections}
Three context injections feed the Planner's initial prompt via \textbf{placeholders}: \texttt{\{observation\}}, \texttt{\{history\}}, and \texttt{\{network\_status\}}. \textbf{\texttt{\{observation\}}} is the raw CybORG dictionary: a verbose, noisy dump of per-host process tables, network connections, and service states that the LLM must parse unaided. \textbf{\texttt{\{network\_status\}}} is a compact JSON list of only non-baseline hosts, each annotated with status (\texttt{changed}, \texttt{unknown}, \texttt{analysed}), recency (\texttt{Current} vs.\ \texttt{Past}), and the action history applied to that host. It collapses to a single sentence when all hosts are healthy. \textbf{\texttt{\{history\}}} is a compressed action log that folds consecutive quiet steps (\textsc{Monitor}, no state change) into ranges while preserving full detail for intervention steps, controlling prompt growth over the 30-step episode. \texttt{\{history\}} thus serves a dual function: it provides temporal context for intervention sequencing and supplies the Planner's own prior reasoning through programmatically extracted justifications, creating a compressed log of past decisions with explanations.

\paragraph{The Initial User Prompt} The initial user prompt template itself contributes nothing beyond a step counter and a closing question. All environment understanding comes from the content injected into these \textbf{placeholders}. Each sub-agent also receives a role-specific initial prompt and context components provided by the Planner. The Analyst receives the target hostname and is asked to assess the host's situation and the ActionChooser receives a situational JSON from the Planner containing the target host, threat description, severity level, and relevant context from prior steps. Sub-agents see neither the full network status nor the episode history. The Planner decides what to provide, enforcing limited context per role. All structured injections are fully deterministic and their cost is limited to the marginal tokens they add to the prompt (examples in Appendix~\ref{app:agent-definitions}). 

\subsection{Deliberation Tools}\label{sec:reasoning-axis}
To further increase decision-making capabilities each ReAct agent in the hierarchy supports four cumulative levels of deliberation. The deliberation tools implement a self-questioning, self-critique, and self-refinement cycle inspired by Self-Refine~\cite{madaan2023selfrefine} and self-interrogation techniques~\cite{press2023selfask}; the CoT injection follows Wei et al.~\cite{wei2022cot} and Kojima et al.~\cite{kojima2022zeroshotcot}. Unlike cross-episode reflection (e.g., Reflexion~\cite{shinn2023reflexion}), all deliberation occurs within a single step and carries no memory to future steps.

The following three \emph{deliberation tools} and an explicit chain-of-thought (CoT) prompt injection are executed as additional ReAct turns inside one loop \emph{within a single environment step}:
(1)~\textbf{question}: the agent questions its initial reasoning before committing to an action~\cite{press2023selfask}. (2)~\textbf{critique} (includes~\#1): the agent generates an explicit critique of its response~\cite{madaan2023selfrefine}. (3)~\textbf{improve} (includes~\#1-2): the agent revises its action in light of the critique~\cite{madaan2023selfrefine}. (4)~\textbf{COT} (includes~\#1-3): an explicit chain-of-thought instruction~\cite{wei2022cot,kojima2022zeroshotcot} is added to the system prompt, providing reasoning scaffolding on top of the tools.

\section{Experimental Methodology}\label{sec:methodology}

This section describes the models, evaluation protocol, metrics, and controlled ablation design that produce the evidence base for our findings. Table~\ref{tab:setup-summary} provides an at-a-glance summary.

\begin{table}[t]
  \caption{Experimental overview, models evaluated, and evaluation hyperparameters.}
  \label{tab:setup-summary}
  \centering
  \small
  \begin{tabular}{p{0.33\linewidth} p{0.55\linewidth}}
    \toprule
    \textbf{Component} & \textbf{Description} \\
    \midrule
    \textbf{Models: 5 families} & Grok, Llama, Devstral, Qwen, Gemini \\
    \textbf{Axis 1: Context} & 6 configs varying \texttt{\{obs\}}, \texttt{\{hist\}}, \texttt{\{net\}} \\
    \textbf{Axis 2: Deliberation} & 4 levels: question, critique, improve, COT \\
    \textbf{Axis 3: Hierarchy} & Delegation vs.\ delegation + deliberation \\
    \textbf{Default Context} & \texttt{hist+net}: network + history \\
    \textbf{Scale} & 72 exp.; 3{,}475 episodes; 283.9M tokens \\
    \midrule
    \textbf{Models} & Grok, Llama, Devstral, Qwen, G2.5FL, G3FP
    (6 models from 5 families; episode counts in Appendix~\ref{app:episode-counts}) \\
    \midrule
    \textbf{Eval Config} & \\
    Instances per config & 10 (standard) \\
    Runs per instance & 5 \\
    Episode length & 30 steps \\
    Decoding & Deterministic (greedy, temp=0) \\
    \midrule
    \textbf{Metrics} & \\
    Mean episode return & $G=\sum_{t=1}^{T} r_t \le 0$; closer to zero is better \\
    Total tokens/episode & Prompt + completion; main cost primitive \\
    RPTS & Returns per token spent;\\
    Standard deviation & Episode return variability \\
    Validity guard & Replication across 5 families, 6 Models \\
    \midrule
    \textbf{Reference Rewards} & Official CAGE-2 Leaderboard~\cite{cage2} \\
    Top DRL agent & $-3.47$ \\
    Simple heuristic & $-58.83$ \\
    Random Agent & $-154.06$ \\
    Sleeping (no-op) & $-218.65$ \\
    \bottomrule
  \end{tabular}
\end{table}

\paragraph{Models}\label{sec:models}

We evaluate six models from five contemporary model families (Table~\ref{tab:setup-summary}) accessed via OpenRouter (Grok, Llama, Devstral, Qwen) and Google Cloud Gemini API. We adopt a full-coverage design: all six models are evaluated on all 12~configurations across three experimental axes (72 unique model-configuration pairs). This safeguards against reliance on single-model results. All models use deterministic decoding (temperature~0). All models receive identical prompt templates and tool definitions. No per-model tuning is performed.

\paragraph{Evaluation Protocol}\label{sec:eval-protocol}

Each configuration is evaluated over multiple  \emph{containerized agent instances} and multiple \emph{runs} per instance. Each run is launched with a unique seed and fresh agent state. The standard allocation is $10$~instances $\times$ $5$~runs $=50$~episodes per pair. G3FP uses a reduced 25-episode budget. Several G2.5FL/Qwen configs use extended batches (75-100) to resolve uncertainty on key comparisons. These allocation differences do not affect qualitative conclusions, which are validated across all six models (Appendix~\ref{app:episode-counts}). Step-level fresh instantiation ensures token accounting is precise and isolated.

\paragraph{Three-Axis Design with Shared Anchor configuration}\label{sec:three-axis}

The experimental design varies three axes. (1)~\textbf{Axis~1 (Context):} 6 monolithic Planner configurations varying \texttt{\{obs\}}, \texttt{\{hist\}}, and \texttt{\{net\}} placeholders (Table~\ref{tab:system-layers}). (2)~\textbf{Axis~2 (Deliberation):} 4 cumulative levels adding \texttt{+question}, \texttt{+critique}, \texttt{+improve}, and \texttt{+COT} tools to the monolithic Planner. (3)~\textbf{Axis~3 (Hierarchy):} 2 configs: \texttt{hier-base} (delegation to Analyst/ActionChooser, deliberation tools OFF) and \texttt{hier-delib} (delegation + deliberation tools ON on all agents, CoT inactive). Axes~2 and~3 share a common default, the \texttt{hist+net} configuration (structured state + compressed history, no raw obs, no deliberation, no delegation), which emerged from our engineering ladder as the default compound-agent context setting. Axis~1 treats this same configuration as one of six context variants.

\paragraph{Metrics and Statistical Reporting}\label{sec:metrics}

We report \textbf{mean episode return} (sum of rewards $r_t \le 0$, closer to zero is better) as the primary performance metric, alongside \textbf{total tokens per episode} (prompt + completion) as the cost primitive. To jointly evaluate cost and performance, we define \textbf{returns per token spent} (RPTS) as the mean return improvement over the observation-only baseline per kilotoken consumed:
\[
  \text{RPTS} = \frac{G_{\text{config}} - G_{\text{obs}}}{\text{KTok}_{\text{config}}}
\]
where $G_{\text{obs}}$ is the per-model mean return under the \texttt{obs} configuration and $\text{KTok}$ is total tokens (thousands) per episode. RPTS identifies which design choices deliver the largest gains per unit of inference budget; negative values indicate configurations that perform worse than unstructured observation alone. To capture tail risk in this failure-mitigation setting, we report \textbf{standard deviation} and \textbf{catastrophic failure rate} (fraction of episodes $< -150$). We adopt \emph{multi-model replication} as the primary validity guard: findings must replicate across the majority of the tested models.

\paragraph{Reward-Scale Reference Points}\label{sec:reward-scale}
To ground the reward scale, we reference the public CAGE-2 leaderboard~\cite{cage2} (30-step, B\_line setting). Reference reward values are listed in Table~\ref{tab:setup-summary}.

\section{Results}\label{sec:results}

All configurations follow the protocol and metrics defined in Section~\ref{sec:methodology}. Table~\ref{tab:full-results} presents the complete results matrix. Appendix~\ref{app:statistical} reports additional statistical support, including 95\% confidence intervals for mean returns and paired difference confidence intervals for key comparisons. These analyses preserve the qualitative direction of the main findings while highlighting model-dependent uncertainty. We organize the analysis into four subsections corresponding to context (\textbf{RQ1}), the interaction between deliberation and hierarchy (\textbf{RQ2}, \textbf{RQ3}), cost-performance frontiers, and robustness.

\begin{table*}[t]
  \centering
  \caption{Full results matrix. Mean episode return $\pm$ standard deviation
    (Ret~$\pm$~SD) and kilotokens per episode (KTok = tokens $\times 10^3$) across all 72 pairs,
    grouped by experimental axis.
    \underline{Underline}\up\,= best within axis;
    \best{box}\ck\,= best overall per model;
    $\downarrow$\,= worst within axis;
    \worst{box}\dn\,= worst overall per model.
    The shared anchor
    configuration hist+net$^\dagger$ serves as baseline for Axes~2 and~3.}
  \label{tab:full-results}
  \footnotesize
  \begin{tabular}{l rr rr rr rr rr rr}
    \toprule
     & \multicolumn{2}{c}{Grok} & \multicolumn{2}{c}{Llama} & \multicolumn{2}{c}{Devstral} & \multicolumn{2}{c}{Qwen} & \multicolumn{2}{c}{G2.5FL} & \multicolumn{2}{c}{G3FP} \\
    \cmidrule(lr){2-3}\cmidrule(lr){4-5}\cmidrule(lr){6-7}\cmidrule(lr){8-9}\cmidrule(lr){10-11}\cmidrule(lr){12-13}
    Config & Ret~$\pm$~SD & KTok & Ret~$\pm$~SD & KTok & Ret~$\pm$~SD & KTok & Ret~$\pm$~SD & KTok & Ret~$\pm$~SD & KTok & Ret~$\pm$~SD & KTok \\
    \midrule
    \multicolumn{13}{l}{\textit{Axis~1: Context}} \\
    obs & $-98.4$$\pm$70 & 17.7 & \worst{$-214.7$$\pm$22\dn} & 8.8 & \worst{$-155.1$$\pm$65\dn} & 22.2 & \worst{$-218.2$$\pm$20\dn} & 10.3 & $-214.7$$\pm$23 & 68.1 & $-96.8$$\pm$71 & 8.0 \\
    obs+hist & $-89.9$$\pm$75 & 34.6 & $-137.4$$\pm$63 & 19.2 & $-133.9$$\pm$86 & 26.4 & $-93.4$$\pm$72 & 19.0 & $-172.9$$\pm$69 & 125.2 & $-76.3$$\pm$67 & 15.4 \\
    obs+hist+net & $-81.9$$\pm$78 & 33.5 & $-102.6$$\pm$73 & 30.6 & $-85.3$$\pm$77 & 25.1 & $-69.0$$\pm$53 & 18.7 & \underline{$-147.8$$\pm$70}\up & 94.0 & $-82.6$$\pm$64 & 19.4 \\
    obs+net & \underline{$-47.0$$\pm$40}\up & 20.9 & \best{$-51.4$$\pm$20\ck} & 12.3 & \underline{$-72.6$$\pm$48}\up & 15.4 & $-63.1$$\pm$26 & 11.4 & $-200.0$$\pm$41 & 79.4 & $-113.7$$\pm$68 & 9.0 \\
    network & $-86.3$$\pm$31 & 20.2 & $-68.7$$\pm$37 & 13.0 & $-93.3$$\pm$44 & 14.9 & $-109.4$$\pm$59 & 10.6 & \worst{$-215.4$$\pm$19\dn} & 94.4 & \worst{$-136.4$$\pm$49\dn} & 7.9 \\
    hist+net$^\dagger$ & \worst{$-112.9$$\pm$79\dn} & 29.1 & $-57.1$$\pm$54 & 28.6 & $-79.3$$\pm$77 & 21.3 & \underline{$-61.5$$\pm$52}\up & 16.4 & $-208.7$$\pm$39 & 81.7 & \underline{$-52.0$$\pm$58}\up & 18.0 \\
    \midrule
    \multicolumn{13}{l}{\textit{Axis~2: Deliberation (cumulative, on anchor$^\dagger$)}} \\
    +question & $-66.5$$\pm$62\dn & 55.6 & $-104.1$$\pm$84\dn & 84.1 & $-53.9$$\pm$42 & 60.4 & $-92.3$$\pm$69 & 45.7 & $-206.2$$\pm$37\dn & 104.8 & $-100.6$$\pm$57\dn & 30.7 \\
    +critique & $-44.9$$\pm$45 & 74.7 & $-93.4$$\pm$77 & 97.8 & $-62.8$$\pm$58 & 106.9 & $-93.6$$\pm$67\dn & 68.9 & \best{$-128.6$$\pm$95\ck} & 118.2 & $-66.4$$\pm$63 & 41.0 \\
    +improve & $-53.4$$\pm$47 & 154.0 & \underline{$-75.0$$\pm$57}\up & 115.3 & $-80.6$$\pm$70\dn & 153.1 & $-92.4$$\pm$55 & 115.9 & $-168.9$$\pm$60 & 182.9 & $-64.0$$\pm$45 & 58.1 \\
    +COT & \underline{$-42.9$$\pm$33}\up & 144.5 & $-100.8$$\pm$70 & 131.3 & \underline{$-40.9$$\pm$31}\up & 157.4 & \underline{$-55.6$$\pm$43}\up & 162.4 & $-157.0$$\pm$70 & 225.7 & \underline{$-29.9$$\pm$20}\up & 75.1 \\
    \midrule
    \multicolumn{13}{l}{\textit{Axis~3: Hierarchy (on anchor$^\dagger$, structured context)}} \\
    hier-base & \best{$-24.0$$\pm$27\ck} & 141.9 & \underline{$-69.5$$\pm$60}\up & 87.7 & \best{$-37.8$$\pm$37\ck} & 97.0 & \best{$-28.6$$\pm$37\ck} & 79.6 & \underline{$-183.1$$\pm$62}\up & 120.6 & \best{$-16.1$$\pm$3\ck} & 56.4 \\
    hier-delib & $-40.4$$\pm$26\dn & 364.1 & $-108.0$$\pm$74\dn & 158.1 & $-127.4$$\pm$72\dn & 257.7 & $-30.1$$\pm$33\dn & 209.9 & $-186.4$$\pm$58\dn & 270.5 & $-23.6$$\pm$12\dn & 104.8 \\
    \bottomrule
  \end{tabular}
\end{table*}

\subsection{Finding 1: Programmatic State Abstraction Delivers the Largest Gains Per Token}\label{sec:finding1}
Context engineering ablations reveal that a deterministic layer compiling observations into structured summaries delivers the largest return on token investment.
\begin{figure}[htbp]
    \centering
    \includegraphics[width=\linewidth]{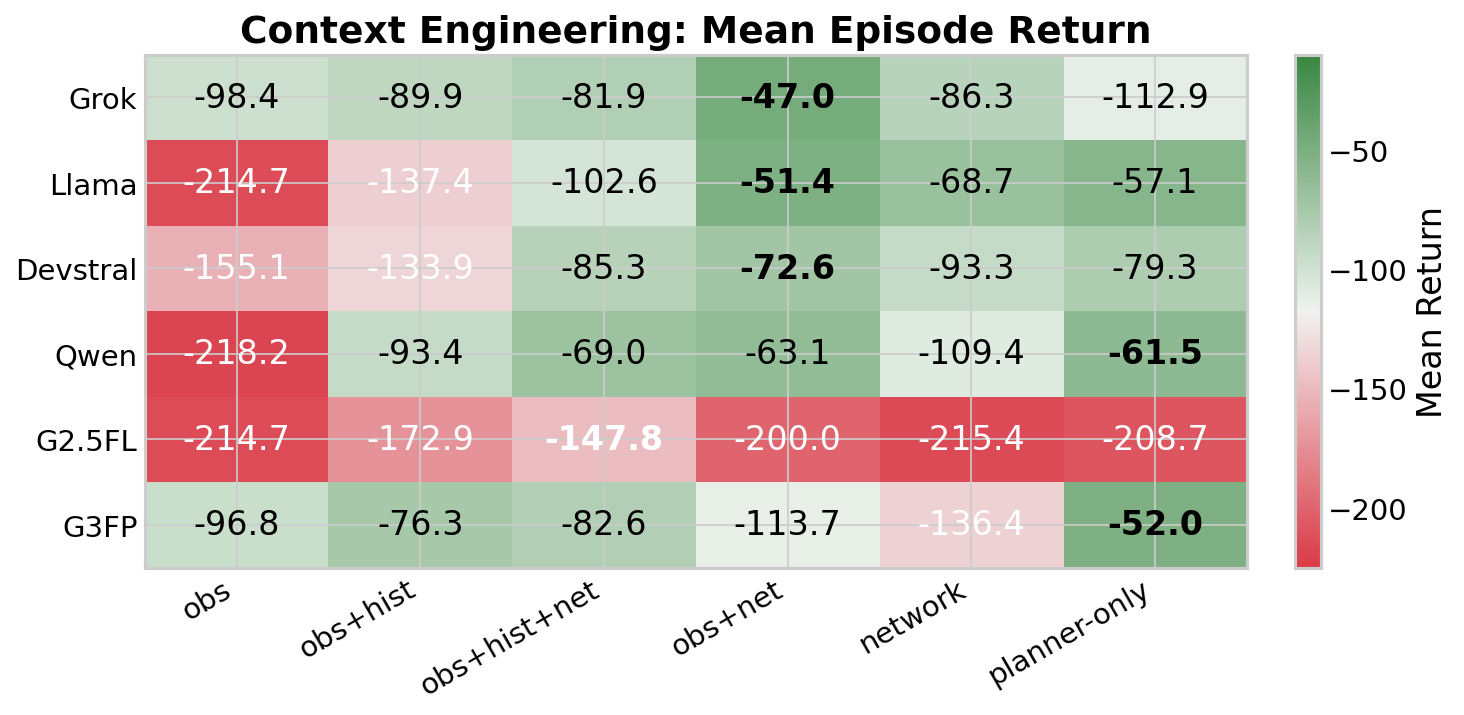}
    \caption{Context engineering heatmap. Each cell shows mean episode return for one model-context pair. Darker shading indicates worse (more negative) returns. The \texttt{obs+net} and \texttt{hist+net}(Planner-Only) columns are consistently strong across models, while \texttt{obs} alone is typically poor.}
    \Description{Heatmap with six model rows and six context configuration columns. Cell colour intensity encodes mean return.}
    \label{fig:context-heatmap}
\end{figure}
\subsubsection{Raw observation alone is ineffective.}
Feeding unprocessed CybORG dictionaries (\texttt{obs}) yields the worst or near-worst performance for five of six models (leftmost column in Figure~\ref{fig:context-heatmap}, Table~\ref{tab:full-results}). Llama, Qwen, and G2.5FL fall below $-214$ mean return, approaching the Sleeping (no-op) agent ($-219$), with catastrophic failure rates ($<-150$) reaching 96-98\% (Figure~\ref{fig:failure-heatmap}). The raw format's verbosity and noise overwhelm the planner.

\subsubsection{Programmatic backbone enables raw data utility.}
Augmenting raw observations with the deterministic \texttt{\{network\_status\}} layer (\texttt{obs}~$\to$~\texttt{obs+net}) transforms performance without additional LLM calls. Llama improves by 76\% ($-214.7 \to -51.4$), Qwen by 71\%, and Devstral by 53\% (Table~\ref{tab:full-results}). The contrast is visible in Figure~\ref{fig:context-heatmap}, where the \texttt{obs+net} column is consistently lighter than \texttt{obs}.

\subsubsection{Less-but-structured often beats more-but-unstructured.}
The \texttt{hist+net} anchor configuration (structured state + history, no raw obs) matches or beats the maximum-information \texttt{obs+hist+net} for four models (Llama, G3FP, Qwen, Devstral). Adding raw observations to a clean state summary often dilutes the signal; removing them improves Llama's return by 44\% (compare the \texttt{hist+net} and \texttt{obs+hist+net} columns in Figure~\ref{fig:context-heatmap}).

\subsubsection{History provides complementary temporal context.}
Adding compressed action history to raw observations (\texttt{obs}$\to$\texttt{obs+hist}) improves all six models, with Llama gaining 36\% and Qwen 57\% (Table~\ref{tab:full-results}). However, history's marginal value depends on what other context is present. When network status is already available, adding history (\texttt{obs+net} vs.\ \texttt{obs+hist+net}) hurts four of six models (Table~\ref{tab:full-results}), suggesting that \texttt{\{network\_status\}} already encodes much of the decision-relevant temporal signal.

\subsection{Finding 2: Deliberation Destructively Interferes with Hierarchy}\label{sec:finding2}
The intersection of Axes~2 and~3 reveals the study's central failure mode: deliberation that helps monolithic agents can degrade hierarchical systems.

\begin{figure}[htbp]
    \centering
    \includegraphics[width=\linewidth]{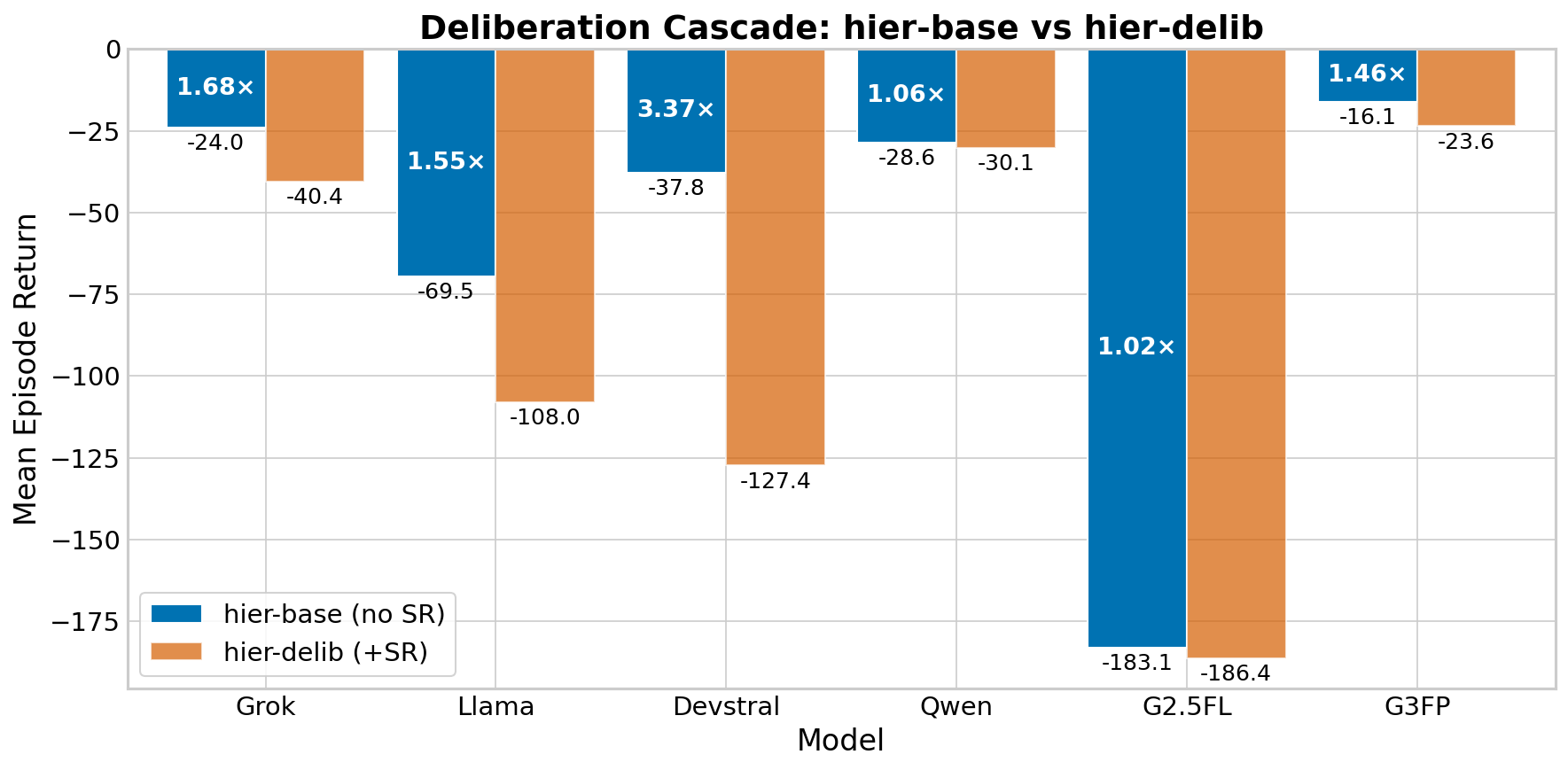}
    \caption{Deliberation cascade effect. Paired bars show mean episode return for \texttt{hier-base} (dark) vs.\ \texttt{hier-delib} (light) across six models. Degradation ratio annotated above each pair. Mean return decreases for all six models.}
    \Description{Paired bar chart comparing hier-base and hier-delib across six models with degradation ratios.}
    \label{fig:compound-metacognition}
\end{figure}

\subsubsection{Monolithic deliberation is model-dependent}\label{sec:mono-sr}
Adding reasoning tools to the monolithic Planner produces model-dependent effects (Table~\ref{tab:full-results}, Axis~2). Explicit chain-of-thought (\texttt{+COT}) is the best deliberation level for four models (Grok, Devstral, Qwen, G3FP), often improving on the anchor configuration. However, patterns are non-monotonic, and stronger baselines (e.g., Llama on anchor configuration) can be destabilized by additional turns. Token costs scale steeply: \texttt{+COT} consumes $3$-$10\times$ more tokens than the anchor.

\subsubsection{Hierarchy without deliberation wins}\label{sec:hier-best}
The \texttt{hier-base} configuration (delegation only, no deliberation tools) achieves the best or near-best absolute performance for four of six models (G3FP at $-16.1$, Grok at $-24.0$, approaching the top published DRL result of $-3.47$~\cite{cardiffuni_cage2_agent}; dark bars in Figure~\ref{fig:compound-metacognition}). The benefit comes from \emph{task decomposition}, bounded Analyst assessments and ActionChooser rankings, rather than deeper deliberation. Llama is the exception, degrading by 22\% compared to monolithic \texttt{hist+net} ($-57.1 \to -69.5$).

\begin{figure*}[htbp]
  \centering
  \includegraphics[width=0.9\linewidth]{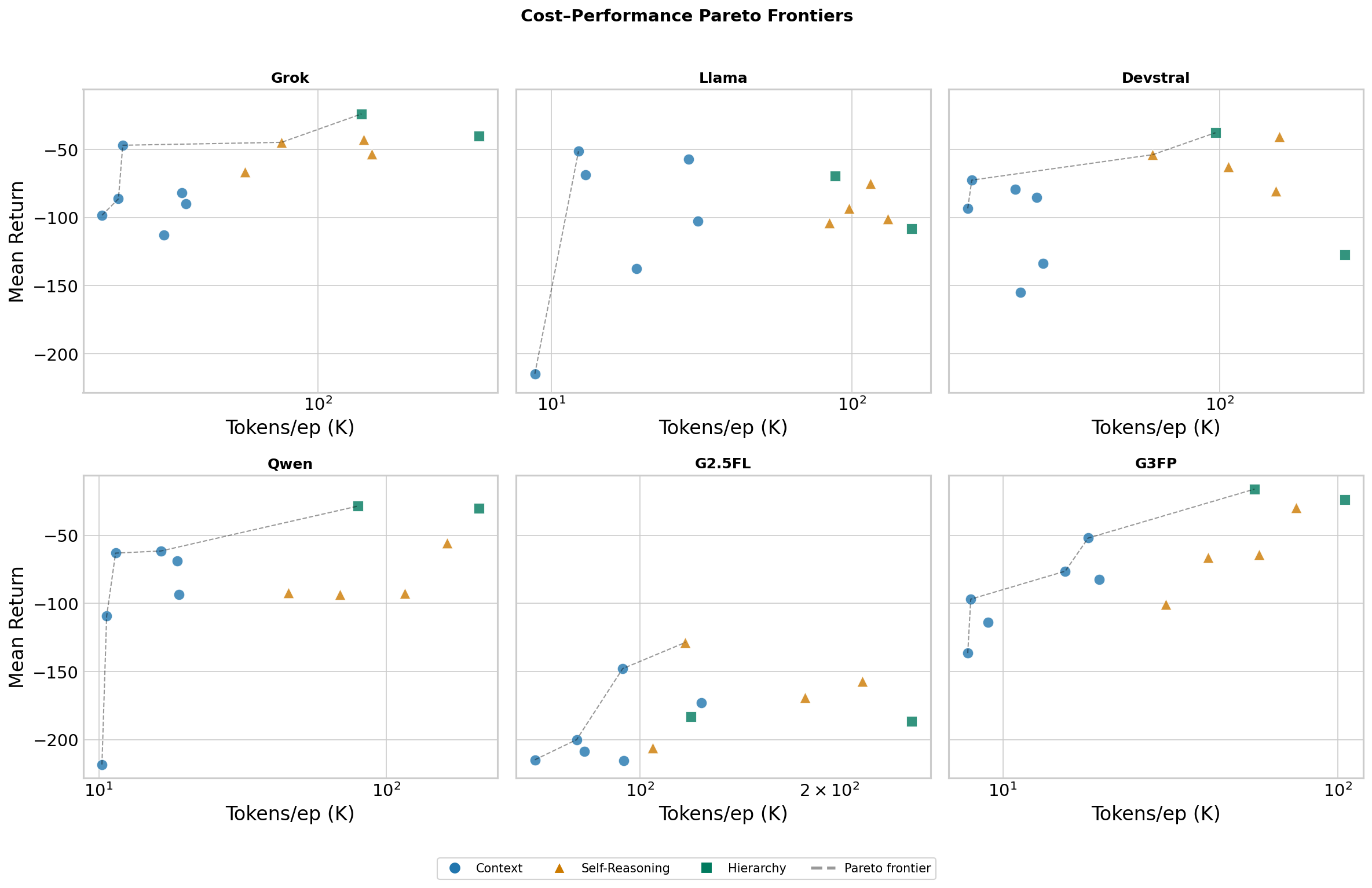}
  \caption{Cost-performance Pareto frontiers. Points shaped by axis (circles: context, triangles: deliberation, squares: hierarchy). Context configurations occupy the low-cost efficient region; \texttt{hier-delib} is high-cost and dominated.}
  \Description{Six scatter subplots with tokens on x-axis and mean return on y-axis. Points shaped by axis. Dashed Pareto frontier lines connect non-dominated configurations.}
  \label{fig:pareto-scatter}
\end{figure*}

\subsubsection{Deliberation cascade: the destructive interaction}\label{sec:compound-meta}
Enabling deliberation tools on all hierarchical agents (\texttt{hier-delib}) decreases mean return relative to \texttt{hier-base} for all six models (Table~\ref{tab:full-results}, Figure~\ref{fig:compound-metacognition}). Devstral worsens by $3.37\times$ ($-37.8 \to -127.4$), Grok by $1.68\times$, and Llama by $1.55\times$, while token costs typically double ($1.8$-$2.7\times$).

This compositional failure arises from \emph{deliberation cascade}: independent deliberation loops in the Analyst, ActionChooser, and Planner amplify uncertainty without an arbitration protocol. Moreover, for four of six models (Grok, Devstral, Qwen, G3FP), \texttt{hier-base} matches or outperforms the best monolithic deliberation at comparable or lower token cost, achieving through task decomposition what deliberation attempts through deeper reasoning. The degradation under \texttt{hier-delib} is not attributable to token overhead per se: \texttt{hier-base} consumes comparable tokens to monolithic \texttt{+COT} (e.g., Grok: 141.9K vs.\ 144.5K) yet achieves substantially better returns ($-24.0$ vs.\ $-42.9$), confirming the loss is structural.

\paragraph{Mechanistic evidence: passivity amplification under distributed deliberation}
Auditing a matched episode trace from Devstral run reveals a recurring pattern of \textbf{passivity amplification}: under low-severity contexts with multiple active decoys, the deliberative ActionChooser converges on "avoid redundant interventions" and defers remediation until evidence becomes unambiguous. For example: at step 24 User4 has been recently analyzed and instrumented with multiple decoys. The ActionChooser's deliberation prioritizes passive observing: \texttt{Monitor} (0.9) $>$ \texttt{Analyse User4} (0.8) $>$ \texttt{Restore User4} (0.7). The Planner adopts the top recommendation. On the next step, the Analyst flags \texttt{severity=high} with concrete anomalies (unusual outbound traffic, unknown processes), and the Planner overrides and restores User4, with a critical delay as in a failure-mitigation regime, even single-step deferral prolongs high-penalty states.
Distributed deliberation overestimates immediate "stability" cues under partial observability, yielding high-confidence deferral until evidence becomes overwhelming (See Appendix~\ref{app:trajectories}).

\begin{table}[htbp]
  \centering
  \caption{RPTS across all configurations
    (Section~\ref{sec:metrics}). Higher is better; negatives = worse than \texttt{obs}.
    \textbf{Bold}\,= best per model.}
  \label{tab:rpts}
  \small
  \begin{tabular}{l rrrrrr}
    \toprule
    Config & Grok & Llama & Devstral & Qwen & G2.5FL & G3FP \\
    \midrule
    \multicolumn{7}{l}{\textit{Axis~1: Context}} \\
    obs+hist     & 0.25 & 4.03 & 0.80 & 6.57 & 0.33 & 1.33 \\
    obs+hist+net & 0.49 & 3.66 & 2.78 & 7.98 & 0.71 & 0.73 \\
    obs+net      & \textbf{2.46} & \textbf{13.28} & \textbf{5.36} & \textbf{13.60} & 0.19 & $-1.88$ \\
    network      & 0.60 & 11.23 & 4.15 & 10.26 & $-0.01$ & $-5.01$ \\
    hist+net$^\dagger$ & $-0.50$ & 5.51 & 3.56 & 9.55 & 0.07 & \textbf{2.49} \\
    \midrule
    \multicolumn{7}{l}{\textit{Axis~2: Deliberation}} \\
    +question    & 0.57 & 1.32 & 1.68 & 2.75 & 0.08 & $-0.12$ \\
    +critique    & 0.72 & 1.24 & 0.86 & 1.81 & \textbf{0.73} & 0.74 \\
    +improve     & 0.29 & 1.21 & 0.49 & 1.09 & 0.25 & 0.56 \\
    +COT         & 0.38 & 0.87 & 0.73 & 1.00 & 0.26 & 0.89 \\
    \midrule
    \multicolumn{7}{l}{\textit{Axis~3: Hierarchy}} \\
    hier-base     & 0.52 & 1.66 & 1.21 & 2.38 & 0.26 & 1.43 \\
    hier-delib      & 0.16 & 0.67 & 0.11 & 0.90 & 0.10 & 0.70 \\
    \bottomrule
  \end{tabular}
\end{table}

\subsection{Finding 3: Context Engineering Dominates the Cost-Performance Frontier}\label{sec:pareto}

We construct per-model Pareto frontiers over all twelve configurations (tokens per episode vs.\ mean return, Figure~\ref{fig:pareto-scatter}). Three patterns emerge. (1) Across per-model Pareto frontiers, at least one context configuration is efficient for every model, typically \texttt{obs+net} or \texttt{hist+net}. (2) Deliberation configurations reach the frontier for only a minority of models and are often dominated by context or hierarchy. (3) \texttt{hier-base} sets the high-performance frontier where hierarchy helps, while \texttt{hier-delib} is consistently dominated by \texttt{hier-base}, combining higher token cost with worse mean return.
\begin{figure*}[htbp]
  \centering
  \includegraphics[width=0.8\linewidth]{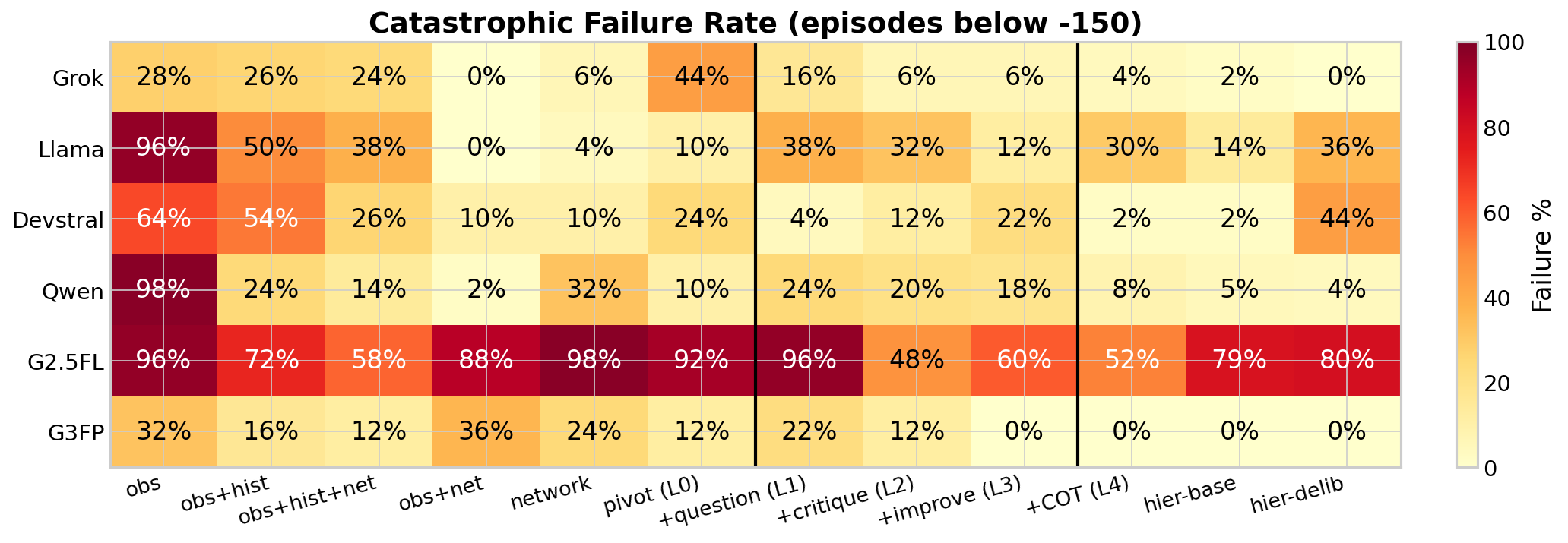}
  \caption{Catastrophic failure rate (return $< -150$) by model and configuration. G2.5FL fails across all configurations; context engineering reduces catastrophic rates for most other models.}
  \Description{Heatmap with six model rows and twelve configuration columns showing catastrophic failure percentages. G2.5FL row is uniformly dark.}
  \label{fig:failure-heatmap}
  \end{figure*}
\paragraph{Return per Token Spent}
Table~\ref{tab:rpts} reports RPTS for all non-baseline configurations. Context configurations dominate: \texttt{obs+net} achieves the highest RPTS for Grok, Llama, Devstral, and Qwen, while \texttt{hist+net} leads for G3FP. G2.5FL is the exception, with \texttt{+critique} performing best, consistent with its difficulty exploiting structured context. Hierarchy often improves absolute return but is less token-efficient: \texttt{hier-base} consumes substantially more tokens than \texttt{obs+net}, yielding lower RPTS despite stronger mean return. \texttt{hier-delib} remains high-cost and low-efficiency relative to \texttt{hier-base}. As a pricing sensitivity check, we re-weighted token costs using provider-specific input/output pricing ratios (Appendix~\ref{app:tokens}); this did not reverse any qualitative conclusion, although it narrowed the relative cost advantage of context over hierarchy.

\subsection{Robustness, Variance, and Tail Risk}\label{sec:robustness}
Qualitative effects (context helps, distributed deliberation hurts) are consistent across models, and magnitudes vary by $2$-$10\times$. Standard deviations (SD) of episode return, reported alongside means in Table~\ref{tab:full-results}, reveal how each design axis affects outcome variability. Context engineering compresses both mean and variance. Programmatic state abstraction reduces SD alongside mean return (e.g., Grok: 70 to 40, Devstral: 65 to 48) and reduces catastrophic failure ($<-150$) from $>90\%$ under \texttt{obs} to $<10\%$ in the strongest cases (Figure~\ref{fig:failure-heatmap}). Bounded hierarchy further tightens outcomes: G3FP under \texttt{hier-base} achieves SD\,$=$\,3, and Grok's SD falls from 79 to 27. Conversely, \texttt{hier-delib} increases variance: Devstral's SD rises from 37 to 72 and its catastrophic rate jumps from 2\% to 44\%, consistent with the deliberation cascade. G2.5FL fails catastrophically ($>48\%$) across all configurations, suggesting a capability floor for the structured I/O compliance the architecture requires; positive claims do not depend on G2.5FL. Hierarchy provides limited or negative value for two models. Llama worsens under hierarchy ($-57.1 \to -69.5$, 22\%) with SD remaining high at 60. G2.5FL improves only marginally (12\%; $-208.7 \to -183.1$). The remaining four models improve substantially (52-79\%). Multi-model evaluation is thus essential: a Llama-only study would conclude hierarchy hurts, whereas a Grok-only study would assert it is important.

\section{Discussion}\label{sec:discussion}

Our findings share a unifying theme: in adversarial sequential POMDPs, the value of a design choice is determined by \emph{information flow} through the system rather than per-component merit. We distill three design principles (\textbf{RQ1}-\textbf{RQ3}).

\paragraph{Principle 1: Invest in deterministic infrastructure before LLM reasoning (RQ1).}\label{sec:principles}
The programmatic state-tracking layer delivers the largest consistent gains per token by shifting the LLM from perception-plus-reasoning to reasoning-over-state. Knowledge-free agents rely solely on this scaffolding, so gains come from \emph{presentation} and \emph{uncertainty compression}, not domain expertise. Context engineering also compresses tail risk: catastrophic failure (return $< -150$, roughly the Random Agent level) drops from $>90\%$ under \texttt{obs} to $<10\%$ under \texttt{obs+net} or \texttt{hist+net} in the strongest cases (Figure~\ref{fig:failure-heatmap}), making it the most reliable lever for average and worst case.

\paragraph{Principle 2: Decompose into bounded specialists, not reflective generalists (\textbf{RQ2}, \textbf{RQ3}).}
Hierarchy without deliberation (\texttt{hier-base}) achieves best or near-best absolute performance for four of six models. The benefit is \emph{interface constraints}: the Analyst gives a bounded assessment and the ActionChooser a ranked list, turning an open-ended generation problem into a verifiable decision.

\paragraph{Principle 3: Do not distribute deliberation without an uncertainty-resolution protocol (\textbf{RQ2}, \textbf{RQ3}).}
Enabling deliberation across all hierarchical agents degrades performance and inflates cost. Independent critique loops create \emph{cascading uncertainty}. When a sub-agent critiques its own answer, it introduces excessive qualifications. The consuming agent cannot distinguish these from genuine warnings about the environment, so caution accumulates through the hierarchy. If deliberation is needed across a hierarchy, centralize it or use explicit mediation (e.g., confidence gating, aggregation rules).

\paragraph{Model dependence and practical ordering.}\label{sec:model-dependence}
Effect directions (context helps, distributed deliberation hurts) hold across models, but magnitudes vary by $2$-$10\times$ and some effects flip (e.g., Llama harmed by hierarchy). Single-model studies would thus contradict each other. The best first step differs by model (context for weak raw-observation handling, hierarchy for strong baselines). The observed Pareto ordering follows our engineering trajectory: (1)~context is the most reliable lever; (2)~bounded hierarchy sets the ceiling; (3)~monolithic deliberation is capability-conditional; (4)~distributed deliberation is dominated. Our results complement topology-focused scaling~\cite{kim2025} by showing that internal configuration determines whether decomposition helps or hurts, revealing failure modes invisible to topology-only analyses. Each layer builds on the previous, and the value of hierarchy and deliberation is conditional on the quality of infrastructure and context.

\paragraph{Transferability.}
The effect directions (Principles~1--3) replicate across six models from five families, making them directional starting points rather than hard prescriptions; specific magnitudes are tied to CAGE-2 and the models tested. A practitioner should: (1)~add structured programmatic context (environment model, history tracking, observation decomposition, baseline state); (2)~compare monolithic configurations against raw observations; (3)~test bounded hierarchy if feasible; and (4)~avoid distributing deliberation tools across sub-agents by default. The state-tracking layer is likely to transfer where observations are structured enough that a program can track what has changed and why, and most steps are routine so history compresses without losing important information.


\section{Related Work}\label{sec:related}

Our study sits at the intersection of four research streams that prior work typically addresses in isolation.

\paragraph{Multi-Agent Architecture and Scaling}
Recent work categorizes coordination protocols~\cite{qian2024macnet,tran2025survey}, derives topology scaling laws~\cite{kim2025}, and proposes structured communication formats~\cite{benkhaled2026g2cp,tang2025statedelta}. These studies vary wiring while treating nodes as fixed; we hold wiring fixed and ablate \emph{internal} configuration, revealing failure modes like deliberation cascades that are invisible to topology-only analyses.

\paragraph{Autonomous Cyber Defense}
CybORG CAGE-2~\cite{cage2,kiely2023cage} has been addressed with RL~\cite{bates2023rewardshaping}, model-based planning~\cite{hammar2024cpomcp}, and particle filtering~\cite{le2025bfp}. LLM-based defenders~\cite{mohammadi2025llm,castro2025llm} are newer but lack controlled architectural ablations. We provide the first cost-performance analysis of compound LLM design decisions in CAGE-2.

\paragraph{Context Engineering}
Context design is a critical lever~\cite{karpathy2025}, supported by tooling ecosystems~\cite{langchain2025} and algorithmic context evolution~\cite{zhang2025ace}. We contribute an orthogonal, controlled ablation of context \emph{composition} (raw vs.\ structured) in a POMDP, showing that deterministic state abstraction outperforms raw observations at near-zero marginal cost.

\paragraph{Deliberation and Self-Critique}
Intra-step deliberation techniques, such as chain-of-thought~\cite{wei2022cot,kojima2022zeroshotcot}, self-interrogation~\cite{press2023selfask}, and self-refinement~\cite{madaan2023selfrefine}, operate within a single inference call, unlike cross-episode methods such as Reflexion~\cite{shinn2023reflexion}. Recent work shows that self-correction without external feedback can be harmful in monolithic agents~\cite{huang2024selfcorrect}, and Renze and Guven~\cite{renze2024selfreflection} decompose reflection components but evaluate only monolithic settings. No prior work studies deliberation \emph{distributed} across a hierarchy. Our deliberation cascade finding extends the single-agent self-correction limitation to compound systems, identifying a compositional failure mode invisible to either literature in isolation.


\section{Limitations and Future Work}\label{sec:limitations}
Our claims are scoped to \emph{structured adversarial POMDPs} where deterministic state abstraction is feasible. We use a single environment (fixed topology, scripted adversary, 30-step horizon) and a single three-agent hierarchy; alternative topologies may exhibit different deliberation cascade dynamics. Token counts proxy cost but do not capture latency or pricing, and knowledge-free prompts isolate architectural effects without fully disentangling pretrained priors. Deliberation tools are tested only in cumulative activation order; independent activation may yield different interaction patterns. Our model selection spans mid-tier and efficiency-focused families; frontier-scale models may respond differently. Our design space is static. Priorities for future work include independent ablation of individual deliberation tools to isolate which components drive the cascade, selective deliberation placement within the hierarchy (e.g., enabling deliberation on only one sub-agent), and evaluation on frontier-scale models to test whether the observed effects persist at higher capability levels. Extending ablations to diverse environments and designing inter-agent uncertainty arbitration protocols, such as confidence gating or calibrated aggregation, are also important next steps.

\section{Conclusion}\label{sec:conclusion}

We presented a controlled cost-performance study of compound LLM agent design in an adversarial, partially observable sequential environment (CybORG CAGE-2). Across a three-axis ablation of context representation (6 configurations), deliberation depth (4 cumulative levels), and hierarchical decomposition (2 configurations), we evaluated 72 model-configuration pairs spanning five model families, totaling 3{,}475 episodes and 283.9M tokens.

Three conclusions emerge. First (\textbf{RQ1}), \emph{context engineering dominates}: deterministic programmatic state abstraction yields the largest and most consistent gains per token, while raw observations alone are destabilizing. Second (\textbf{RQ2}), \emph{hierarchy can substitute for deliberation}: bounded specialist decomposition (\texttt{hier-base}) achieves the best absolute performance for most models through strict I/O contracts rather than deeper per-agent reasoning. Third (\textbf{RQ3}), \emph{deliberation is not modular}: distributing deliberation tools across a hierarchy (\texttt{hier-delib}) produces a \emph{deliberation cascade} that degrades returns while increasing token expenditure.

A cross-cutting finding reinforces these conclusions: qualitative effects hold across all six models, but magnitudes vary by $2$-$10\times$ and some reverse sign (e.g., Llama is harmed by hierarchy), validating multi-model evaluation as essential for compound AI research. These results suggest a practical starting point for structured adversarial POMDPs, mirroring the trajectory we followed: build deterministic infrastructure to deliver clean structured context, add bounded hierarchy when models can exploit delegation, and treat deliberation as a costly capability-conditional option rather than a universal upgrade. This is not a universal prescription, as magnitudes are environment-dependent, and practitioners should validate this ordering in their own settings. More broadly, our findings suggest that the science of compound AI systems requires studying topology, node internals, and the interaction effects that arise when individually sensible components are composed. The deliberation cascade identified in this study is one such interaction effect, and designing inter-agent uncertainty arbitration protocols to prevent it is a promising direction for future work. Reproducibility details and ethics considerations are in Appendix~\ref{sec:reproducibility}. The archived artifact is available at \url{https://doi.org/10.5281/zenodo.19908100}; the development repository is available at \url{https://github.com/isbogdanov/agent-design-study}.

\bibliographystyle{ACM-Reference-Format}
\bibliography{bib}

\appendix

  \noindent\textbf{Appendix organization.} Appendix~\ref{sec:reproducibility} provides reproducibility details and ethics considerations. Appendix~\ref{app:agent-definitions} lists the complete YAML definitions for the Planner, Analyst, and ActionChooser. Appendix~\ref{app:sr-tools} documents the deliberation tool schemas and activation flags. Appendix~\ref{app:results} reports complete results, cross-axis comparisons, distributional analysis, and token cost data for all 72 model--configuration pairs. Appendix~\ref{app:statistical} provides 95\% confidence intervals and paired mean-return difference confidence intervals. Appendix~\ref{app:tokens} breaks down token consumption and prompt/completion shifts. Appendix~\ref{app:trajectories} provides illustrative trajectory excerpts for the deliberation cascade failure mode. Appendix~\ref{app:cage2} summarizes CAGE-2 environment details, network topology, and instance difficulty. Appendix~\ref{app:episode-counts} lists the evaluated episode counts per model and configuration. 

\section{Reproducibility \& Ethics}\label{sec:reproducibility}
\paragraph{Ethics.}
All authors have read and adhere to the ACM Code of Ethics\footnote{https://www.acm.org/code-of-ethics}. All experiments run within the simulated CybORG CAGE-2 environment~\cite{cage2}; no real networks, live attack infrastructure, or human-subject data are involved. The work is strictly defensive in scope. LLM tools were used for language polishing and data processing scripts; all design decisions, analyses, and claims are authored by the research team.

\paragraph{Reproducibility.}

All models use deterministic decoding (temperature 0 or provider minimum).
We release the source code, exact YAML configuration snapshots, episode-allocation
metadata, container specification, API-key template, and experiment runner needed to
rerun the evaluated variants. The paper reports the aggregate results, token accounting, and episode counts used for the main
claims, while the artifact provides the implementation and configuration snapshots
needed for inspection and selected reruns. We rely on cross-model replication of
qualitative effects rather than single-model statistical significance.

\subsection{Artifact Availability and Scope}

The artifact supporting this paper is archived on Zenodo at
\url{https://doi.org/10.5281/zenodo.19908100}. The development repository is
available at \url{https://github.com/isbogdanov/agent-design-study}. Detailed
build, configuration, and execution instructions are provided in the artifact
README.

The artifact contains the agent implementation, experiment runner,
container specification, API-key template, and configuration snapshots needed to
rerun evaluated variants. The main implementation is in
\texttt{agent\_base/}, including the CybORG and LLM-agent coordinators,
provider-connector configuration, logging utilities, and the YAML-defined
Planner, Analyst, and ActionChooser agents. \texttt{exp\_configs/} contains the
twelve paper configurations (six context, four deliberation, two hierarchy),
each a self-contained YAML snapshot; switching conditions requires changing
\texttt{definitions\_source} in \texttt{experiment\_agent\_eval.yaml}.

Experiments are launched via \texttt{run\_experiment.py} (10 instances $\times$ 5 runs = 50 episodes per pair by default). The Dockerfile builds a Python
environment with CybORG CAGE-2 and dependencies; LLM access is
supplied through the provided \texttt{.env.template}.

A run creates an \texttt{experiments/} directory with the copied configuration,
per-instance reports, aggregate summaries, and token-usage logs. Full raw
LLM-provider transcripts are not bundled due to storage size; original run seeds
are also not included, so the artifact supports executable reruns rather than
bit-for-bit log regeneration. Original logs may be available upon request.
The primary reproducibility target is to enable inspection of the implementation
and configuration snapshots, and rerunning of selected configurations. 

\section{Agent Definitions}\label{app:agent-definitions}

This appendix provides the complete YAML definitions for all three agents
(Planner, Analyst, ActionChooser), demonstrating the near-zero-knowledge
starting point described in Section~\ref{sec:knowledge-free}. Each agent is
defined by three files: \texttt{core.yaml} (identity, model binding, tool
flags), \texttt{initial\_prompt.yaml} (per-step prompt template), and
\texttt{persistent\_knowledge.yaml} (domain knowledge). A shared
\texttt{common\_knowledge.yaml} applies to all agents. All remaining knowledge
files, \texttt{reflection\_knowledge.yaml}, \\\texttt{reflection\_examples.yaml},
and \texttt{examples.yaml}, are confirmed empty for every agent in every
experiment reported in this paper; the learning subsystem that would populate
them is disabled throughout. The boolean flags
\texttt{include\_tool\_raise\_a\_question}, \texttt{include\_tool\_critique\_the\_answer},
and \\\texttt{include\_tool\_improve\_based\_on\_critique} in each agent's\\
\texttt{core.yaml} are set to \texttt{false} in the baseline configuration and
toggled to \texttt{true} cumulatively for the deliberation axis
(Section~\ref{sec:reasoning-axis}). The YAML shown below reflects the baseline
(anchor) configuration; deliberation variants differ only in these three flags
and in the optional \texttt{include\_COT\_instruction} injection.

\subsection{Shared Configuration}

\paragraph{common\_knowledge.yaml.}
\begin{verbatim}
, 
reflection_knowledge:
\end{verbatim}
This file is empty, confirming that no shared domain heuristics are injected
across agents.

\subsection{Planner}

\paragraph{planner/core.yaml.}
\begin{verbatim}
agent_type: "ReAct"
include_tool_raise_a_question: false
include_tool_critique_the_answer: false
include_tool_improve_based_on_critique: false
include_COT_instruction: false

add_examples: false
add_reflection_examples: true

system_message: |
  You are the strategic decision-maker for the
  Blue team's cybersecurity defense. Your role is
  to select ONE action per step that best defends
  the network.

  You must balance immediate threats with strategic
  positioning, considering that you can only act
  once per step.

  You MUST follow your <reflection_rules> during
  each thought phase.

  Provide a final answer to the question asked
  by the USER.

COT_instruction: |
  You MUST follow your Chain-of-Thought instructions
  during each thought phase.
  1. REVIEW SITUATION: Check network status and step
     history. Which hosts need attention? What actions
     have been tried?
  2. IDENTIFY TARGET: Select the most critical host or
     threat to address this step
  3. GATHER INFO: If needed, use
     get_analysis_of_host_update for detailed analysis
     of a changed host
  4. GET SUGGESTIONS: Call
     get_suggestion_for_next_action with JSON:
     {"target_host": "hostname", "situation":
     "description", "severity": "level", "context":
     "relevant history"}
  5. DECIDE: Choose ONE action from suggestions. You
     may override based on strategic reasoning

rules:
  - You must select ONLY ONE action for your final
    Answer from the list of suggestions provided by
    the 'get_suggestion_for_next_action' tool
  - Your final Answer MUST be a verbatim copy of the
    action-string from ONE of the suggestions
  - TOOLS CANNOT HANDLE MULTIPLE HOSTS, YOU MUST
    SELECT ONLY ONE SPECIFIC HOST AT A TIME

tools:
  - name: "get_analysis_of_host_update"
    description: "Provides a detailed analysis and
      comparison of the state change of the specific
      host to its baseline"
    example_calling:
      "get_analysis_of_host_update: User4"

  - name: "get_suggestion_for_next_action"
    description: |
      Suggests a ranked list of final actions, each
      with a justification and confidence score based
      on recent analysis and action history.

      CRITICAL INPUT REQUIREMENT: Your input MUST be
      a valid JSON object with these required keys:
      - "target_host": The specific single hostname
      - "situation": Brief description of the threat
      - "severity": Threat level
        (low|medium|high|critical)
      - "context": Relevant information from
        previous steps
    example_calling: >
      get_suggestion_for_next_action:
      {"target_host": "Enterprise1", "situation":
      "critically compromised with active C2",
      "severity": "critical", "context": "Remove
      action failed previously"}
    is_critical: true
\end{verbatim}

The Planner's two domain-specific tools,\\ \texttt{get\_analysis\_of\_host\_update} \\ and \texttt{get\_suggestion\_for\_next\_action}, are the interfaces through which the \textsc{Coordinator} spawns the Analyst and ActionChooser sub-agents in hierarchical configurations (\texttt{hier-base} and \texttt{hier-delib}). When delegation is disabled, the Planner does not invoke these tools and instead emits an environment action directly.

\paragraph{planner/initial\_prompt.yaml.}
\begin{verbatim}
prompt:
  opening: |
    You are at step {step_number} now.

    {network_status}

    {history}

  closing: |
    What action should be taken next?
\end{verbatim}

The placeholders \texttt{\{network\_status\}}, \texttt{\{history\}}, and
(in context variants that include it) \texttt{\{observation\}} are populated
deterministically at each step by the environment-state layer described in
Section~\ref{sec:env-model}. The anchor configuration includes
\texttt{\{network\_status\}} and \texttt{\{history\}} but omits
\texttt{\{observation\}}.

\paragraph{Context output examples.}\label{app:context-examples}
The \texttt{\{network\_status\}} placeholder is rendered as a JSON list of all
non-baseline hosts, annotated with current status, recency, and action history:

{\begin{verbatim}
Network Status: The following hosts have
updates or are in a non-baseline state:
[
  {"host_name": "Enterprise1",
   "current_status": "changed",
   "time_of_update": "Current",
   "applied_actions_so_far": "Analyse->Remove"},
  {"host_name": "User4",
   "current_status": "unknown",
   "time_of_update": "Past",
   "applied_actions_so_far": "Remove"}
]
\end{verbatim}
}

\noindent When all hosts are healthy, a single sentence reports baseline status,
keeping the prompt compact. The \texttt{\{history\}} placeholder is rendered as
a compressed action log with smart collapsing: consecutive quiet steps are
folded into ranges when no state changes are detected, while steps involving
interventions retain full detail:

{\begin{verbatim}
Steps 1-3: Action: Monitor/No action needed.
  (No state changes observed)
Step 4: {"action": "Analyse Enterprise1",
  "analysis": "Host shows suspicious processes"}
Step 5: {"action": "Remove Enterprise1"}
Step 6: Action: Monitor
\end{verbatim}
}

\paragraph{planner/persistent\_knowledge.yaml.}
\begin{verbatim}
reflection_knowledge:
  - content:
    - Description: monitoring network for malicious
        activity.
      Example use: Monitor
      Name: Monitor
      Type: Passive observing action
    - Description: it does not remove the host but
        attempts to remove malicious infection from
        a host.
      Example use: Remove hostname=someName
      Name: Remove
      Type: Reactive intervention
    - Description: analyzing a host for malicious
        activity at deeper system level
      Example use: Analyse hostname=someName
      Name: Analyse
      Type: Passive observing action
    - Description: restoring a host to a clean state
        with very high penalty, this action removes
        all previously deployed decoys from the host
      Example use: Restore hostname=someName
      Name: Restore
      Type: Reactive intervention
    - Description: deploying a service to act as a
        decoy to a host to distract the attacker
        in future steps
      Example use: DecoySERVICE hostname=someName
      Name: DecoySERVICE
      Type: Proactive Protective action
    header: >
      INTERPRETATION OF ACTIONS FROM
      <AVAILABLE_ACTIONS_LIST>
    root: actions
    type: json
\end{verbatim}

This is the only domain knowledge provided to the Planner (and, identically,
to the ActionChooser): a five-entry action-type glossary with names, types,
and usage syntax. No tactical heuristics, no threat-assessment rubrics, and no
worked examples are included.

\paragraph{Empty knowledge files.}
The following files are empty for all Planner experiments:
\texttt{examples.yaml}, \texttt{reflection\_examples.yaml}, and
\texttt{reflection\_knowledge.yaml}.

\subsection{Analyst}

\paragraph{analyst/core.yaml.}
\begin{verbatim}
agent_type: "ReAct"
include_tool_raise_a_question: false
include_tool_critique_the_answer: false
include_tool_improve_based_on_critique: false
include_COT_instruction: false

add_examples: false
add_reflection_examples: true

system_message: |
  You are a cybersecurity analyst.
  You MUST follow your <reflection_rules> during
  each thought phase.
  Provide a final answer to the question asked
  by the USER.

COT_instruction: |
  You MUST follow your Chain-of-Thought instructions
  during each thought phase.
  1. GET CURRENT STATE: Use get_host_current_state
     for the target host
  2. GET BASELINE: Use get_host_baseline_state to
     compare against initial state
  3. IDENTIFY ANOMALIES: What changed? New processes,
     connections, missing services?
  4. ASSESS SEVERITY: How critical is this compromise?
     Is there C2 activity?
  5. RECOMMEND ACTION: Should we contain, investigate
     further, or just monitor?

tools:
  - name: "get_host_current_state"
    description: "Get the current state details for a
      specific host. The input must be a single
      hostname."
    example_calling: "get_host_current_state:
      Enterprise1"

  - name: "get_host_baseline_state"
    description: "Get the baseline state details for
      a specific host. The input must be a single
      hostname."
    example_calling: "get_host_baseline_state:
      Enterprise1"

answer_format: |
  Your response MUST STRICTLY be a JSON array of
  objects that follows the following schema:
  {
    "host": "<n>",
    "status": "baseline|changed|unknown",
    "severity": "low|medium|high|critical",
    "anomalies": ["<short bullet>", "..."],
    "suspected_c2": ["<ip/host>"],
    "recommended_focus":
      "containment|investigation|monitor",
    "rationale": "<2-3 sentences>"
  }
\end{verbatim}

The Analyst receives a one-sentence role identity and two read-only tools
(\texttt{get\_host\_current\_state}, \texttt{get\_host\_baseline\_state}) that
return deterministic JSON from the environment-state layer. Its output is a
structured JSON assessment with a constrained schema. No domain heuristics are
provided: its \texttt{persistent\_knowledge.yaml} and all example files are
empty.

\paragraph{analyst/initial\_prompt.yaml.}
\begin{verbatim}
prompt:
  opening: |
    You must provide an analysis for the host:
    {hostname}.
  closing: |
    What is your understanding of the situation
    with the host?
\end{verbatim}

\subsection{ActionChooser}

\paragraph{action\_chooser/core.yaml.}
\begin{verbatim}
agent_type: "ReAct"
include_tool_raise_a_question: false
include_tool_critique_the_answer: false
include_tool_improve_based_on_critique: false
include_COT_instruction: false

add_examples: false
add_reflection_examples: true

system_message: |
  You are a tactical cybersecurity advisor for the
  Blue team.

  You MUST follow your <reflection_rules> during
  each thought phase.

  Provide a final answer to the question asked
  by the USER.

COT_instruction: |
  You MUST follow your Chain-of-Thought instructions
  during each thought phase.
  1. READ SITUATION: Check SITUATION_JSON for
     target_host, threat description, severity,
     and context
  2. EVALUATE ACTIONS: Consider available action
     types and their costs vs benefits
  3. RANK THREE: Provide three suggestions with
     confidence scores (0.0-1.0), highest confidence
     first

answer_format: |
  Your response MUST STRICTLY be a JSON array of
  objects, where each object represents a suggested
  action. Each object must have ONLY the following
  keys: "action", "confidence".
\end{verbatim}

The ActionChooser has \emph{no tools}, it is a pure generation agent that
receives a situation JSON and returns a ranked action list. Its
\texttt{persistent\_knowledge.yaml} contains the same five-entry action glossary
as the Planner (reproduced above). All example files are empty.

\paragraph{action\_chooser/initial\_prompt.yaml.}
\begin{verbatim}
prompt: |
  <SITUATION_JSON>
  {situational_understanding}
  </SITUATION_JSON>

  The JSON above contains:
  - "target_host": The specific hostname requiring
    action (CRITICAL: All your suggested actions
    MUST target this host)
  - "situation": Description of the current threat
  - "severity": The threat level
    (low|medium|high|critical)
  - "context": Additional relevant information
    from previous steps

  Provide THREE action suggestions for the
  target_host specified in the JSON, ordered from
  highest confidence to lowest.
\end{verbatim}

\section{Deliberation Tool Schemas}\label{app:sr-tools}

The deliberation tools are three generic self-critique operations implemented
in the shared \texttt{BaseToolExecutor} class and inherited by all agent types.
They are toggled via boolean flags in each agent's \texttt{core.yaml}; the
cumulative activation sequence defines the four deliberation levels in
Axis~2 (Section~\ref{sec:reasoning-axis}).

\paragraph{Tool activation sequence.}
Table~\ref{tab:sr-activation} shows the cumulative activation of each flag.

\begin{table}[htbp]
  \centering
  \caption{Deliberation tool activation by experimental level. Each level
    cumulatively adds capabilities; \texttt{+COT} adds all three tools plus
    an explicit chain-of-thought system prompt injection.}
  \label{tab:sr-activation}
  \scriptsize
  \begin{tabular}{l cccc}
    \toprule
    \textbf{Flag} & \textbf{+ques.} & \textbf{+crit.} & \textbf{+impr.} & \textbf{+COT} \\
    \midrule
    \texttt{include\_tool\_raise\_a\_question} & \checkmark & \checkmark & \checkmark & \checkmark \\
    \texttt{include\_tool\_critique\_the\_answer} & & \checkmark & \checkmark & \checkmark \\
    \texttt{include\_tool\_improve\_based\_on\_critique} & & & \checkmark & \checkmark \\
    \texttt{include\_COT\_instruction} & & & & \checkmark \\
    \bottomrule
  \end{tabular}
\end{table}

\paragraph{raise\_a\_question.}
When invoked, the agent formulates a question directed at itself. The tool
returns the question text as an "observation" prompting the agent to answer
it in the next reasoning turn. This creates a self-interrogation loop: the
agent pauses its action-selection process to surface uncertainties or
alternative framings. The tool accepts free-text input (the question) and
returns a formatted prompt: \emph{"You have chosen ask yourself: \{question\}.
What can you answer to yourself?"}

\paragraph{critique\_the\_answer.}
The agent submits its current reasoning as input. The tool returns this text
as a critique prompt, asking the agent to evaluate its own logic:
\emph{"You have chosen to critique your reasoning: `\{reasoning\}'. Now,
provide your critical assessment."} This forces a second pass over the
agent's tentative conclusion before action commitment.

\paragraph{improve\_based\_on\_critique.}
After critiquing, the agent submits an improved version of its reasoning. The
tool returns: \emph{"You have decided to improve your answer based on the
critique: \\`\{improved\_reasoning\}'. Now, provide your final answer."} This
creates a three-phase deliberation cycle: question $\to$ critique $\to$ improve.

\paragraph{COT instruction injection.}
The \texttt{+COT} level does not add a fourth tool. Instead, it sets \texttt{include\_COT\_instruction=true},
which injects the \texttt{COT\_instruction} block from the agent's \texttt{core.yaml}
(shown in Appendix~\ref{app:agent-definitions}) into the system prompt. This provides explicit step-by-step
reasoning guidance tailored to each agent's role. Combined with the three tools, \texttt{+COT} represents the
maximum deliberation configuration.

\paragraph{Scope of distribution.}
In \texttt{hier-base}, only the Planner's deliberation flags are toggled (the
Analyst and ActionChooser retain \texttt{false} for all flags). In
\texttt{hier-delib}, the same flags are toggled for \emph{all three agents},
creating the distributed deliberation condition studied in
Section~\ref{sec:compound-meta}.

\section{Complete Results}\label{app:results}

\begin{figure}[h]
\centering
\includegraphics[width=\linewidth]{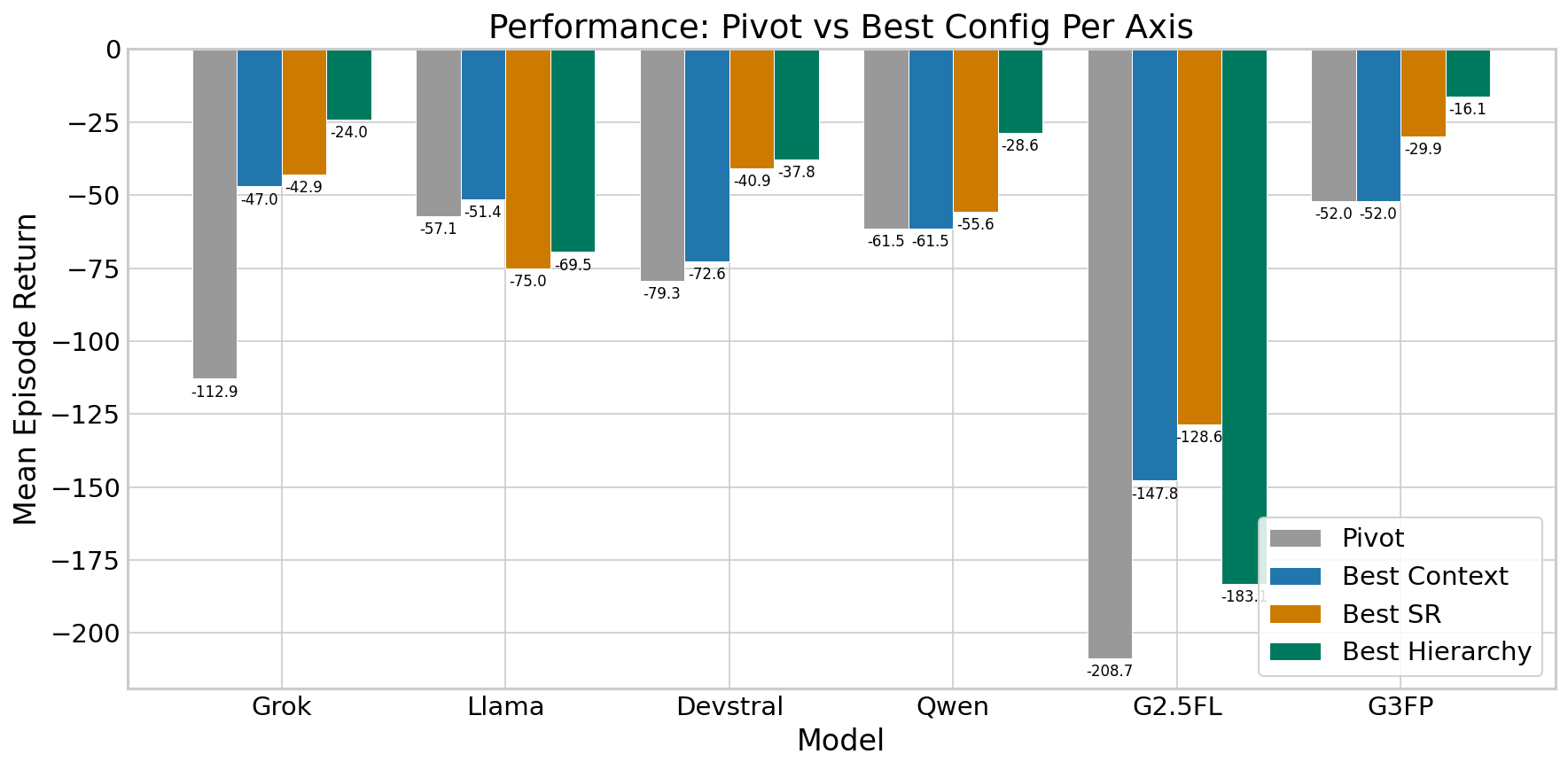}
\caption{Best configuration per axis compared to the shared anchor configuration. For most models, the largest absolute improvement comes from hierarchy (Axis~3), but context engineering (Axis~1) achieves competitive gains at a fraction of the token cost.}
\Description{Grouped bar chart comparing the anchor configuration return against the best configuration on each axis for all six models.}
\label{fig:headline-overview}
\end{figure}

This appendix presents supplementary results for all 72 model--configuration
pairs, organized by experimental axis. Figure~\ref{fig:headline-overview}
compares each axis's best configuration against the shared anchor.

\subsection{Context Engineering}\label{app:results-context}

\begin{figure}[h]
\centering
\includegraphics[width=\linewidth]{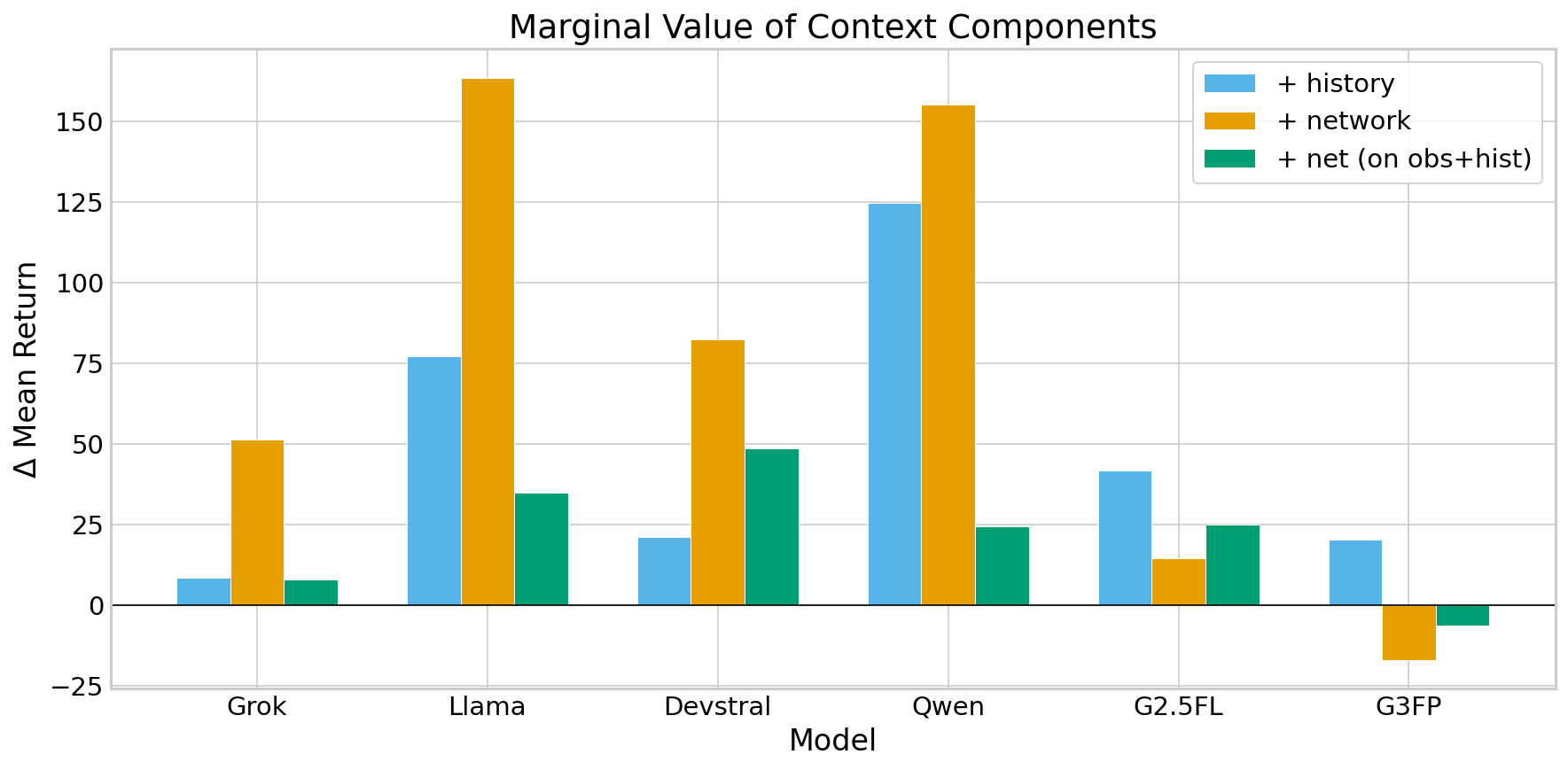}
\caption{Marginal value of adding individual context components. Positive values indicate improvement. Adding \texttt{\{network\_status\}} to raw observation delivers the largest consistent gains.}
\Description{Bar chart of marginal improvement from adding each context component, grouped by model.}
\label{fig:marginal-value}
\end{figure}

\begin{figure}[h]
\centering
\includegraphics[width=\linewidth]{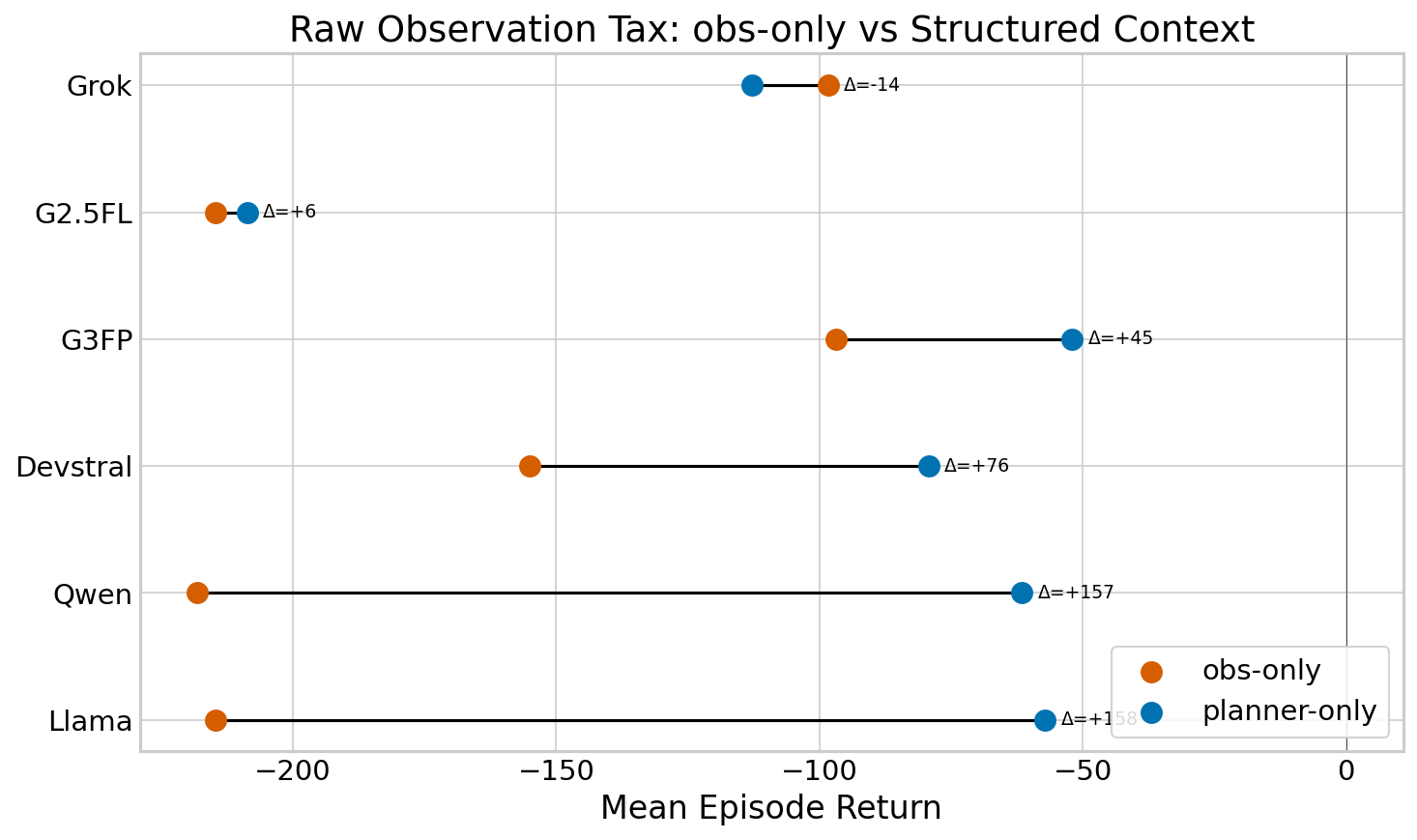}
\caption{Raw observation penalty. Gap between \texttt{obs}-only and the structured \texttt{hist+net} anchor configuration per model. Longer bars indicate larger benefit from replacing raw observations with programmatic context.}
\Description{Horizontal bar chart comparing obs-only and hist+net returns for each model.}
\label{fig:raw-obs-penalty}
\end{figure}

\begin{figure*}[htbp]
\centering
\includegraphics[width=\linewidth]{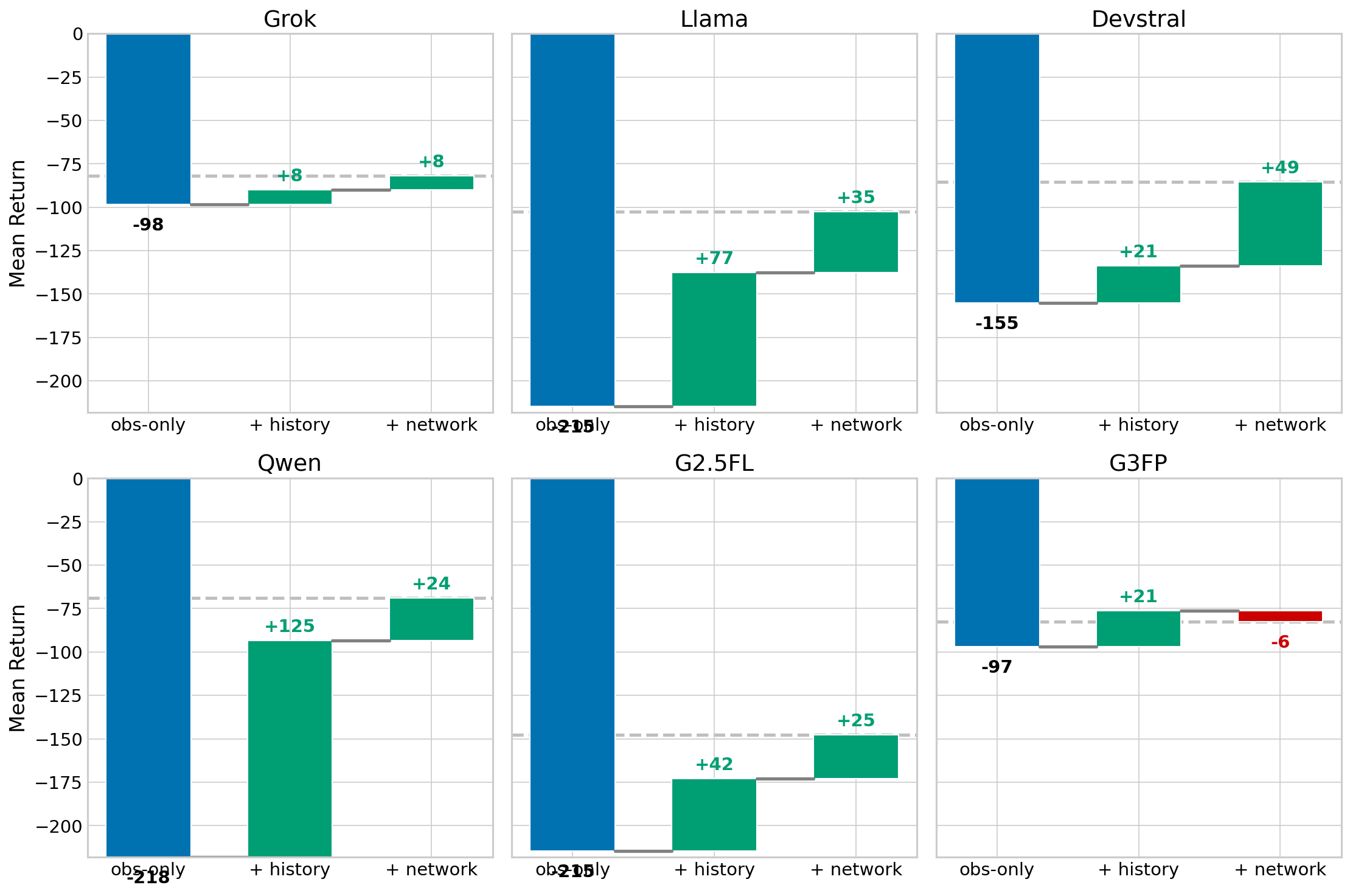}
\caption{Context component waterfall. Additive effect of history and network status on top of raw observation. Green = improvement, red = degradation.}
\Description{Context component waterfall per model.}
\label{fig:context-waterfall}
\end{figure*}

\begin{figure}[htbp]
\centering
\includegraphics[width=\linewidth]{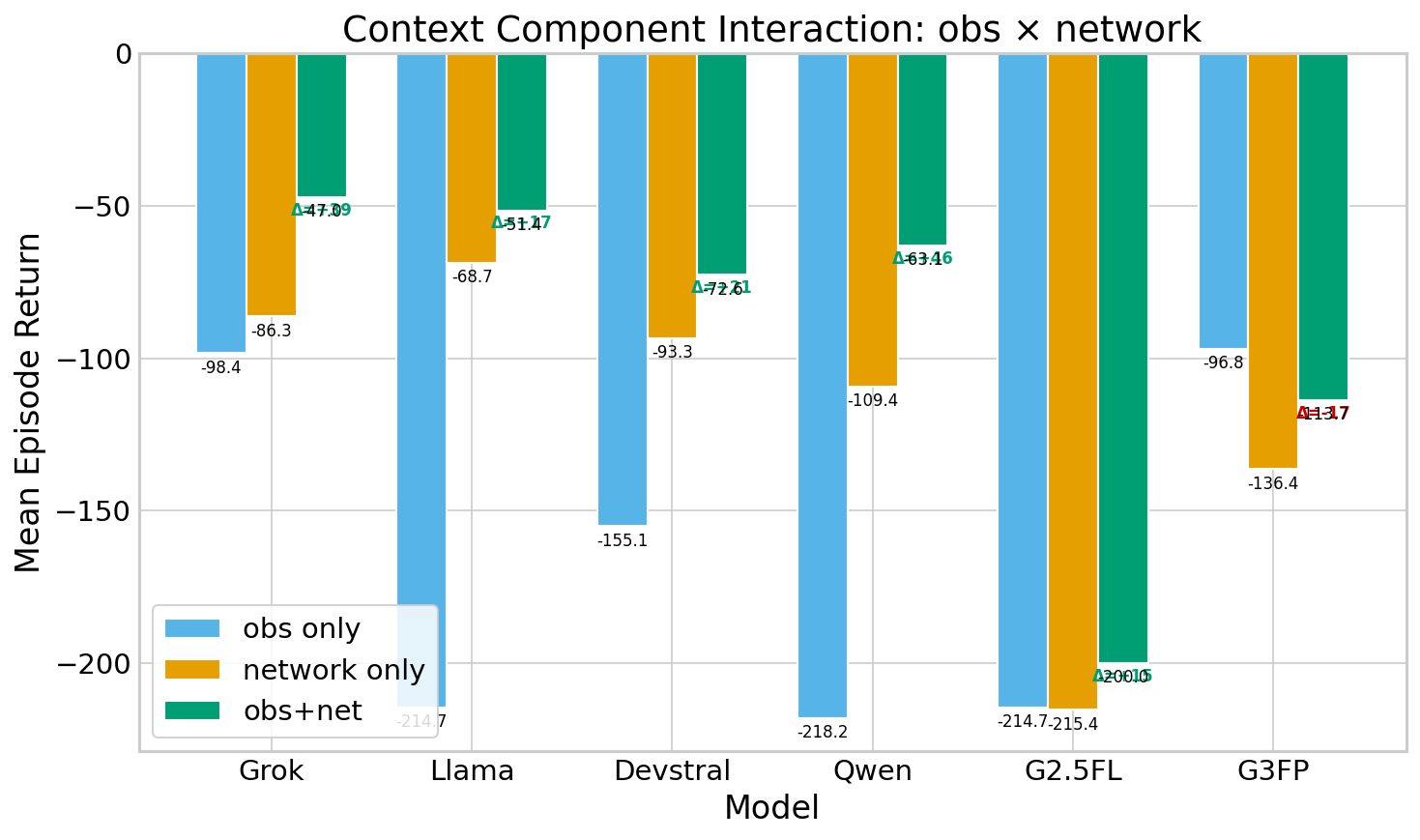}
\caption{Context component interaction. obs+net compared to obs-only and network-only. Synergy ($\Delta$) shows whether combining exceeds the better individual.}
\Description{Context component synergy analysis.}
\label{fig:context-interaction}
\end{figure}

Table~\ref{tab:results-std} provides standard deviations complementing the
main-text hero table (Table~\ref{tab:full-results}).
Table~\ref{tab:marginal-context} reports the marginal value of adding or
removing individual context components.
Figures~\ref{fig:marginal-value} and~\ref{fig:raw-obs-penalty} visualize
context component marginal gains and the raw-observation penalty.
Figures~\ref{fig:context-waterfall} and~\ref{fig:context-interaction} show
context component interactions.

\begin{table*}[htbp]
  \centering
  \caption{Full results with standard deviation. Mean episode return
    ($\pm$ standard deviation) across all 72 model--configuration pairs,
    complementing the token-cost data in Table~\ref{tab:full-results}.
    Configurations are grouped by experimental axis.}
  \label{tab:results-std}
  \begin{tabular}{l l r r r r r r}
    \toprule
    \textbf{Group} & \textbf{Config} & \textbf{Grok} & \textbf{Llama} & \textbf{Devstral} & \textbf{Qwen} & \textbf{G2.5FL} & \textbf{G3FP} \\
    \midrule
    Context & obs & $-98.4$$\pm$69.8 & $-214.7$$\pm$22.5 & $-155.1$$\pm$64.7 & $-218.2$$\pm$19.8 & $-214.7$$\pm$22.8 & $-96.8$$\pm$70.7 \\
     & obs+hist & $-89.9$$\pm$75.3 & $-137.4$$\pm$63.3 & $-133.9$$\pm$85.5 & $-93.4$$\pm$71.7 & $-172.9$$\pm$69.4 & $-76.3$$\pm$67.2 \\
     & obs+hist+net & $-81.9$$\pm$78.0 & $-102.6$$\pm$73.2 & $-85.3$$\pm$77.5 & $-69.0$$\pm$53.2 & $-147.8$$\pm$69.5 & $-82.6$$\pm$63.8 \\
     & obs+net & $\textbf{-47.0}$$\pm$40.3 & $\textbf{-51.4}$$\pm$19.9 & $\textbf{-72.6}$$\pm$47.7 & $-63.1$$\pm$25.9 & $-200.0$$\pm$40.8 & $-113.7$$\pm$68.5 \\
     & network & $-86.3$$\pm$30.9 & $-68.7$$\pm$37.5 & $-93.3$$\pm$44.2 & $-109.4$$\pm$58.8 & $-215.4$$\pm$18.5 & $-136.4$$\pm$49.1 \\
     & hist+net & $-112.9$$\pm$78.7 & $-57.1$$\pm$54.1 & $-79.3$$\pm$76.5 & $\textbf{-61.5}$$\pm$51.8 & $-208.7$$\pm$39.3 & $\textbf{-52.0}$$\pm$57.6 \\
    Delib. & +question & $-66.5$$\pm$62.2 & $-104.1$$\pm$83.8 & $-53.9$$\pm$41.8 & $-92.3$$\pm$69.2 & $-206.2$$\pm$37.0 & $-100.6$$\pm$57.3 \\
     & +critique & $-44.9$$\pm$44.6 & $-93.4$$\pm$77.3 & $-62.8$$\pm$58.0 & $-93.6$$\pm$67.1 & $\textbf{-128.6}$$\pm$94.8 & $-66.4$$\pm$63.4 \\
     & +improve & $-53.4$$\pm$46.6 & $-75.0$$\pm$57.3 & $-80.6$$\pm$70.5 & $-92.4$$\pm$55.0 & $-168.9$$\pm$59.8 & $-64.0$$\pm$44.9 \\
     & +COT & $-42.9$$\pm$32.8 & $-100.8$$\pm$70.1 & $-40.9$$\pm$31.1 & $-55.6$$\pm$43.2 & $-157.0$$\pm$69.7 & $-29.9$$\pm$19.7 \\
    Hierarchy & hier-base & $-24.0$$\pm$27.0 & $-69.5$$\pm$60.0 & $-37.8$$\pm$37.2 & $-28.6$$\pm$36.6 & $-183.1$$\pm$62.4 & $-16.1$$\pm$2.7 \\
     & hier-delib & $-40.4$$\pm$26.1 & $-108.0$$\pm$74.3 & $-127.4$$\pm$71.8 & $-30.1$$\pm$32.7 & $-186.4$$\pm$58.1 & $-23.6$$\pm$12.4 \\
    \bottomrule
  \end{tabular}
\end{table*}

\begin{table*}[htbp]
  \centering
  \caption{Context component marginal value. Each row shows the percentage
    improvement from adding or removing one context component. Positive values
    indicate improvement (return moves toward zero). Transitions with
    $\geq$30\% improvement are \textbf{bolded}.}
  \label{tab:marginal-context}
  \begin{tabular}{l l r r r r r r}
    \toprule
    \textbf{Transition} & \textbf{Change}
      & \textbf{Grok} & \textbf{Llama} & \textbf{Devstral}
      & \textbf{Qwen} & \textbf{G2.5FL} & \textbf{G3FP} \\
    \midrule
    obs $\to$ obs+hist            & Adding history        & 8.6\%  & \textbf{36.0\%}  & 13.7\%           & \textbf{57.2\%}  & 19.5\%  & 21.2\% \\
    obs $\to$ obs+net             & Adding network status & \textbf{52.2\%} & \textbf{76.0\%} & \textbf{53.2\%} & \textbf{71.1\%} & 6.9\%  & $-17.4$\% \\
    obs+hist $\to$ obs+hist+net   & Adding net to obs+hist & 8.9\%  & 25.3\%           & \textbf{36.3\%}  & 26.2\%           & 14.5\%  & $-8.2$\% \\
    network $\to$ hist+net        & Adding history to net  & $-30.8$\% & 16.8\%        & 15.1\%           & \textbf{43.8\%}  & 3.1\%   & \textbf{61.9\%} \\
    obs+net $\to$ obs+hist+net    & Adding hist to obs+net & $-74.3$\% & $-99.6$\%     & $-17.6$\%        & $-9.3$\%         & 26.1\%  & 27.4\% \\
    obs+hist+net $\to$ hist+net   & Dropping raw obs       & $-37.8$\% & \textbf{44.4\%} & 7.1\%          & 10.8\%           & $-41.2$\% & \textbf{37.0\%} \\
    \bottomrule
  \end{tabular}
\end{table*}

\subsection{Hierarchy and Architecture Summary}\label{app:results-hierarchy}

\begin{figure*}[htbp]
\centering
\includegraphics[width=\linewidth]{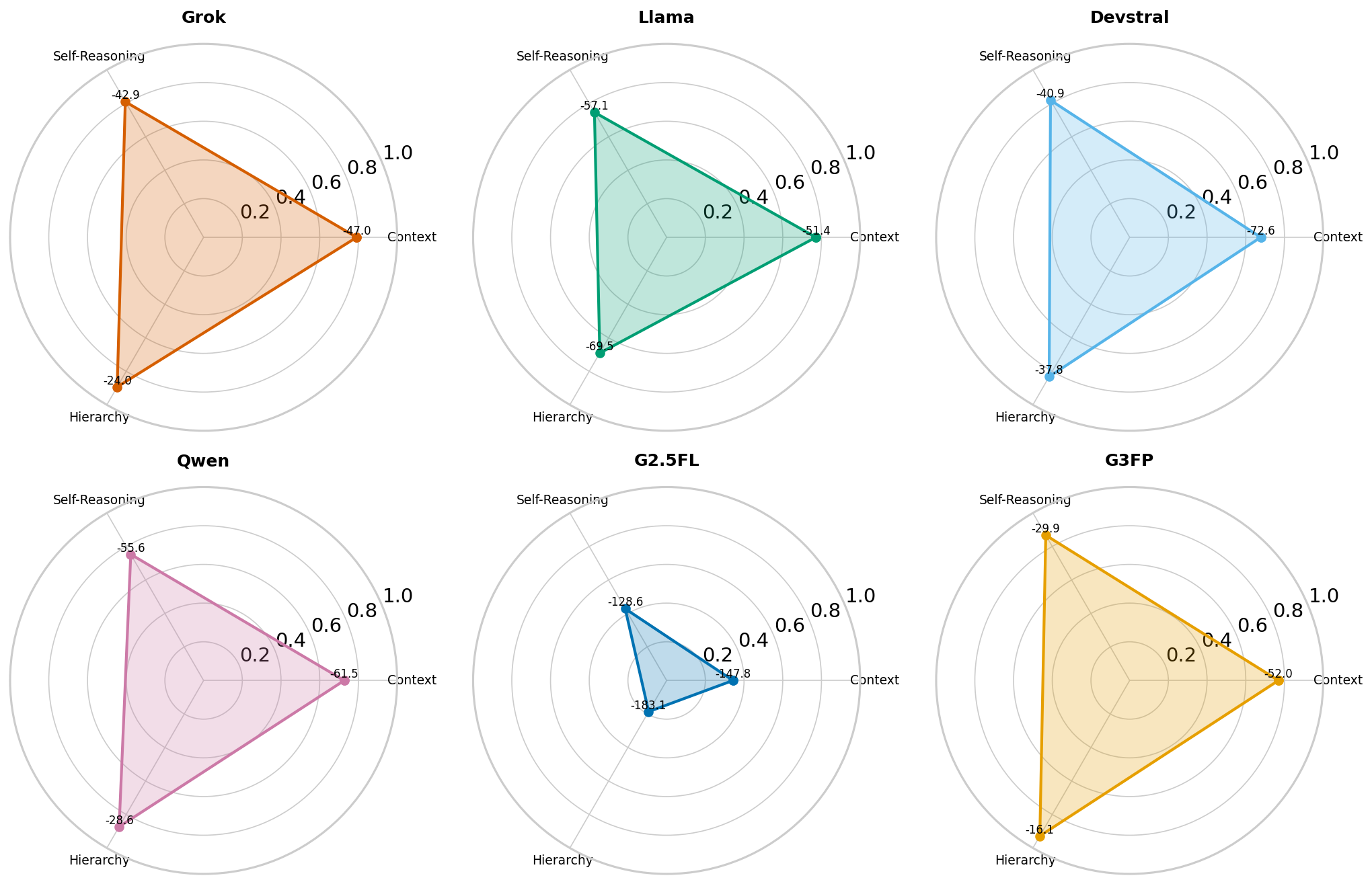}
\caption{Model fingerprints. Each radar shows normalized best performance on three axes (context, deliberation, hierarchy). Larger area indicates better overall performance. Models exhibit distinct capability profiles: Grok and G3FP excel across axes, while G2.5FL is uniformly weak.}
\Description{Six radar chart subplots, one per model. Each has three spokes for context, deliberation, and hierarchy. Filled areas show normalized best scores. Grok and G3FP have the largest areas; G2.5FL the smallest.}
\label{fig:radar}
\end{figure*}

\begin{figure}[htbp]
\centering
\includegraphics[width=\linewidth]{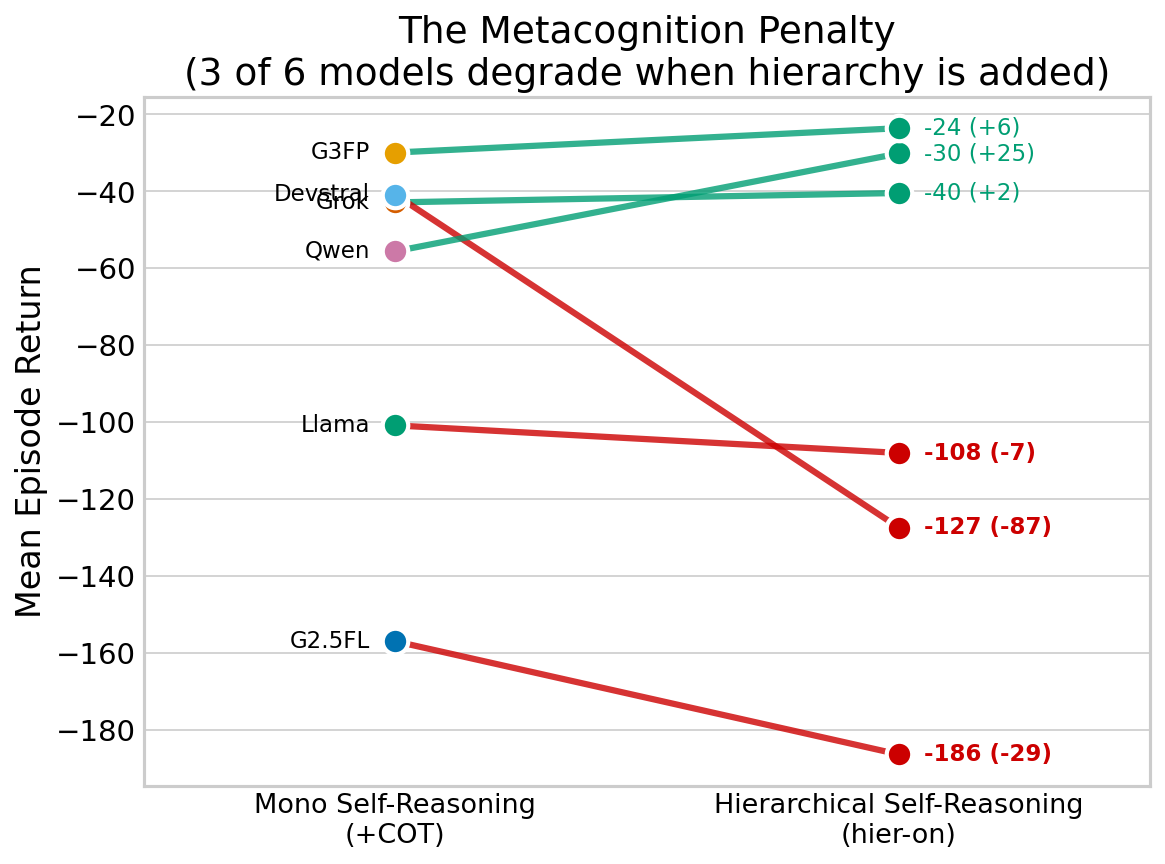}
\caption{The Deliberation Cascade Penalty. Slope chart showing the shift in mean return when moving from monolithic deliberation to hierarchical deliberation. Red = degradation.}
\Description{Deliberation cascade penalty slope chart.}
\label{fig:metacognition-penalty}
\end{figure}

\begin{figure}[htbp]
\centering
\includegraphics[width=\linewidth]{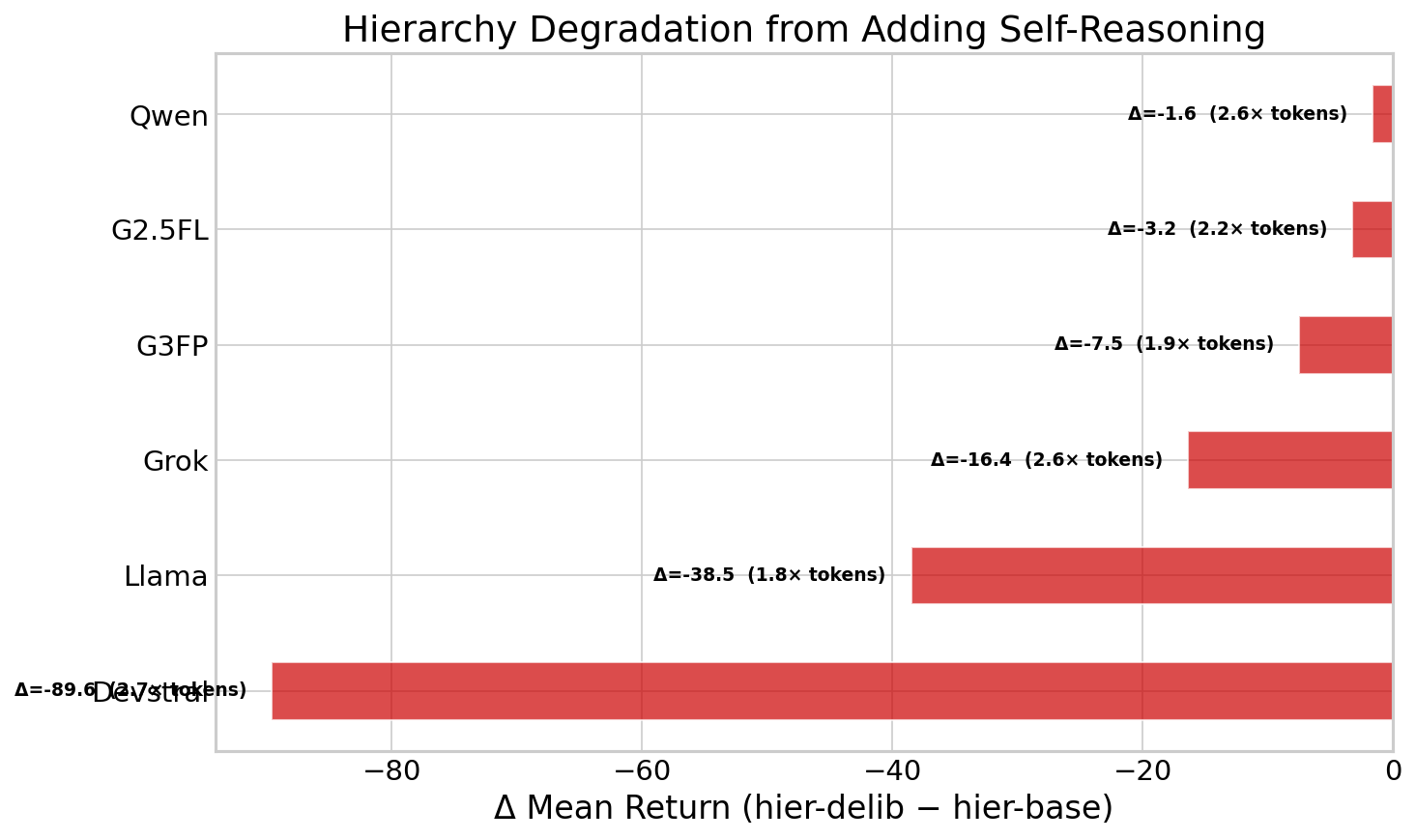}
\caption{Hierarchy degradation. Performance change when distributing deliberation tools across the hierarchy, showing model-specific sensitivity to the deliberation cascade.}
\Description{Hierarchy degradation visualization comparing hier-base and hier-delib performance across models.}
\label{fig:hier-degradation}
\end{figure}

Table~\ref{tab:hierarchy-results} details the hierarchy degradation ratios.
Table~\ref{tab:mono-vs-hier-delib} compares deliberation across monolithic
and hierarchical settings. Table~\ref{tab:best-worst} quantifies the
architectural impact range per model. Figure~\ref{fig:radar} shows model
fingerprints across axes. Figure~\ref{fig:metacognition-penalty} visualizes
the deliberation cascade penalty and Figure~\ref{fig:hier-degradation} shows
the hierarchy degradation pattern.

\begin{table}[htbp]
  \centering
  \caption{Hierarchy results. Mean return, standard deviation, and tokens per
    episode for both hierarchy configurations, plus the degradation ratio
    (\texttt{hier-delib}/\texttt{hier-base}; values $>$1 indicate degradation
    from adding distributed deliberation).}
  \label{tab:hierarchy-results}
  \small
  \begin{tabular}{l rrr rrr c}
    \toprule
    & \multicolumn{3}{c}{hier-base} & \multicolumn{3}{c}{hier-delib} & \\
    \cmidrule(lr){2-4}\cmidrule(lr){5-7}
    Model & Mean & Std & Tok & Mean & Std & Tok & Ratio \\
    \midrule
    G3FP     & $-16.1$ & 2.7  & 56.4K  & $-23.6$  & 12.4 & 104.8K & 1.46$\times$ \\
    Grok     & $-24.0$ & 27.0 & 141.9K & $-40.4$  & 26.1 & 364.1K & 1.68$\times$ \\
    Qwen     & $-28.6$ & 36.6 & 79.6K  & $-30.1$  & 32.7 & 209.9K & 1.06$\times$ \\
    Devstral & $-37.8$ & 37.2 & 97.0K  & $-127.4$ & 71.8 & 257.7K & 3.37$\times$ \\
    Llama    & $-69.5$ & 60.0 & 87.7K  & $-108.0$ & 74.3 & 158.1K & 1.55$\times$ \\
    G2.5FL   & $-183.1$ & 62.4 & 120.6K & $-186.4$ & 58.1 & 270.5K & 1.02$\times$ \\
    \bottomrule
  \end{tabular}
\end{table}

\begin{table*}[htbp]
  \centering
  \caption{Deliberation in monolithic vs.\ hierarchical settings. Compares
    the shared anchor configuration, best monolithic deliberation level, and both hierarchy
    configurations. For most models, \texttt{hier-base} matches or exceeds the
    best monolithic deliberation at comparable cost, while \texttt{hier-delib}
    degrades it.}
  \label{tab:mono-vs-hier-delib}
  \begin{tabular}{l rr rr rr rr}
    \toprule
    & \multicolumn{2}{c}{Anchor (hist+net)}
    & \multicolumn{2}{c}{Best Mono Delib.}
    & \multicolumn{2}{c}{hier-base}
    & \multicolumn{2}{c}{hier-delib} \\
    \cmidrule(lr){2-3}\cmidrule(lr){4-5}\cmidrule(lr){6-7}\cmidrule(lr){8-9}
    Model & Return & Tok & Return & Tok & Return & Tok & Return & Tok \\
    \midrule
    Grok     & $-112.9$ & 29.1K  & $-42.9$  & 144.5K & $-24.0$  & 141.9K & $-40.4$  & 364.1K \\
    Llama    & $-57.1$  & 28.6K  & $-75.0$  & 115.3K & $-69.5$  & 87.7K  & $-108.0$ & 158.1K \\
    Devstral & $-79.3$  & 21.3K  & $-40.9$  & 157.4K & $-37.8$  & 97.0K  & $-127.4$ & 257.7K \\
    Qwen     & $-61.5$  & 16.4K  & $-55.6$  & 162.4K & $-28.6$  & 79.6K  & $-30.1$  & 209.9K \\
    G2.5FL   & $-208.7$ & 81.7K  & $-128.6$ & 118.2K & $-183.1$ & 120.6K & $-186.4$ & 270.5K \\
    G3FP     & $-52.0$  & 18.0K  & $-29.9$  & 75.1K  & $-16.1$  & 56.4K  & $-23.6$  & 104.8K \\
    \bottomrule
  \end{tabular}
\end{table*}

\begin{table}[htbp]
  \centering
  \caption{Architectural impact range per model. Best and worst configurations
    across all twelve, with the return gap quantifying the maximum leverage
    of architectural choices within each model family.}
  \label{tab:best-worst}
  \small
  \begin{tabular}{l l r l r r}
    \toprule
    \textbf{Model} & \textbf{Best Config} & \textbf{Best}
      & \textbf{Worst Config} & \textbf{Worst} & \textbf{Gap} \\
    \midrule
    Grok     & hier-base   & $-24.0$  & hist+net & $-112.9$ & 88.8 \\
    Llama    & obs+net    & $-51.4$  & obs      & $-214.7$ & 163.3 \\
    Devstral & hier-base   & $-37.8$  & obs      & $-155.1$ & 117.3 \\
    Qwen     & hier-base   & $-28.6$  & obs      & $-218.2$ & 189.6 \\
    G2.5FL   & +critique  & $-128.6$ & network  & $-215.4$ & 86.8 \\
    G3FP     & hier-base   & $-16.1$  & network  & $-136.4$ & 120.3 \\
    \bottomrule
  \end{tabular}
\end{table}

\subsection{Deliberation}\label{app:results-deliberation}

\begin{figure}[htbp]
\centering
\includegraphics[width=\linewidth]{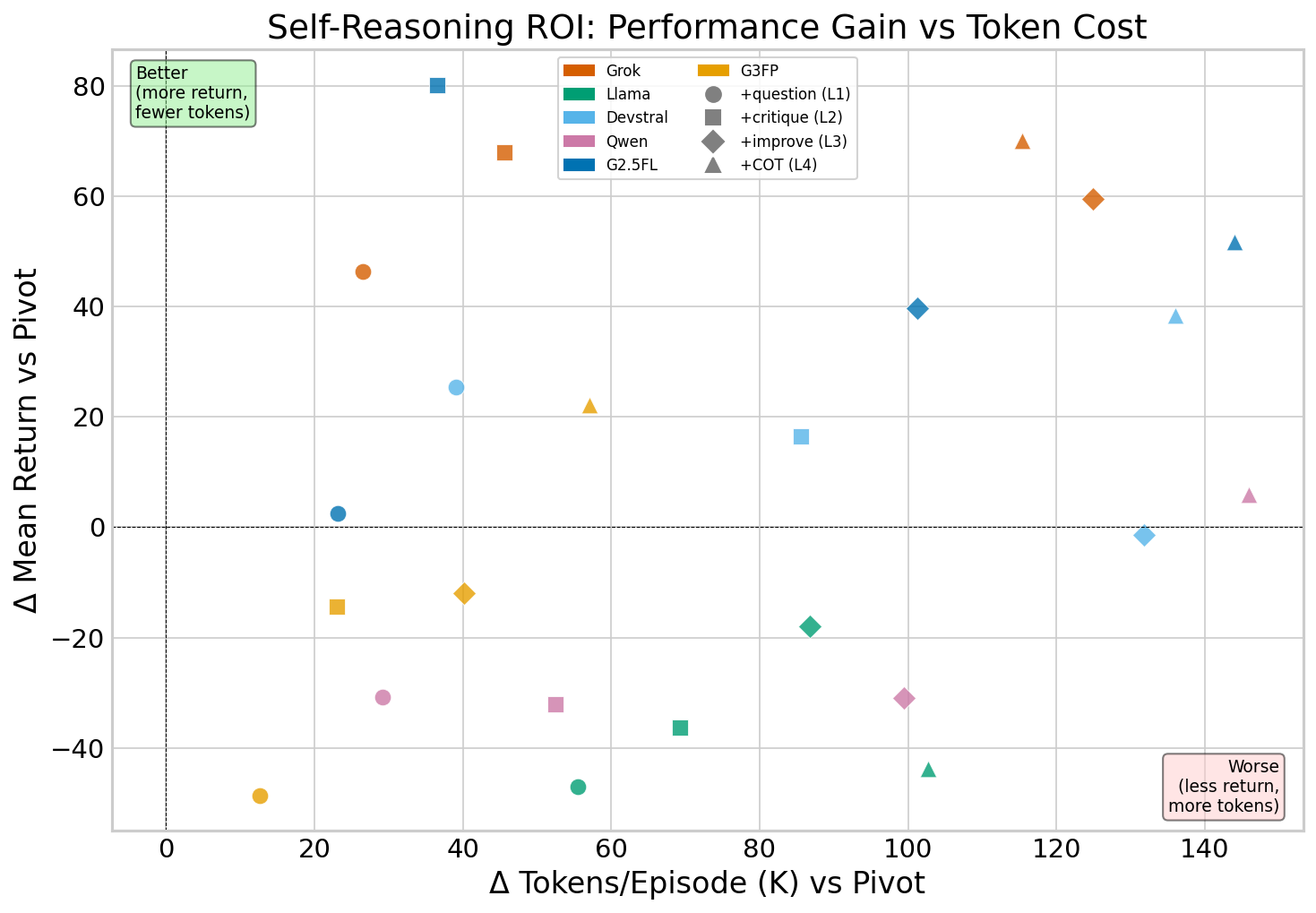}
\caption{Deliberation ROI. Each point shows one model--level pair's change in return and tokens relative to anchor. Upper-left = efficient improvement.}
\Description{Deliberation ROI scatter plot.}
\label{fig:sr-roi-scatter}
\end{figure}

\begin{figure}[htbp]
\centering
\includegraphics[width=\linewidth]{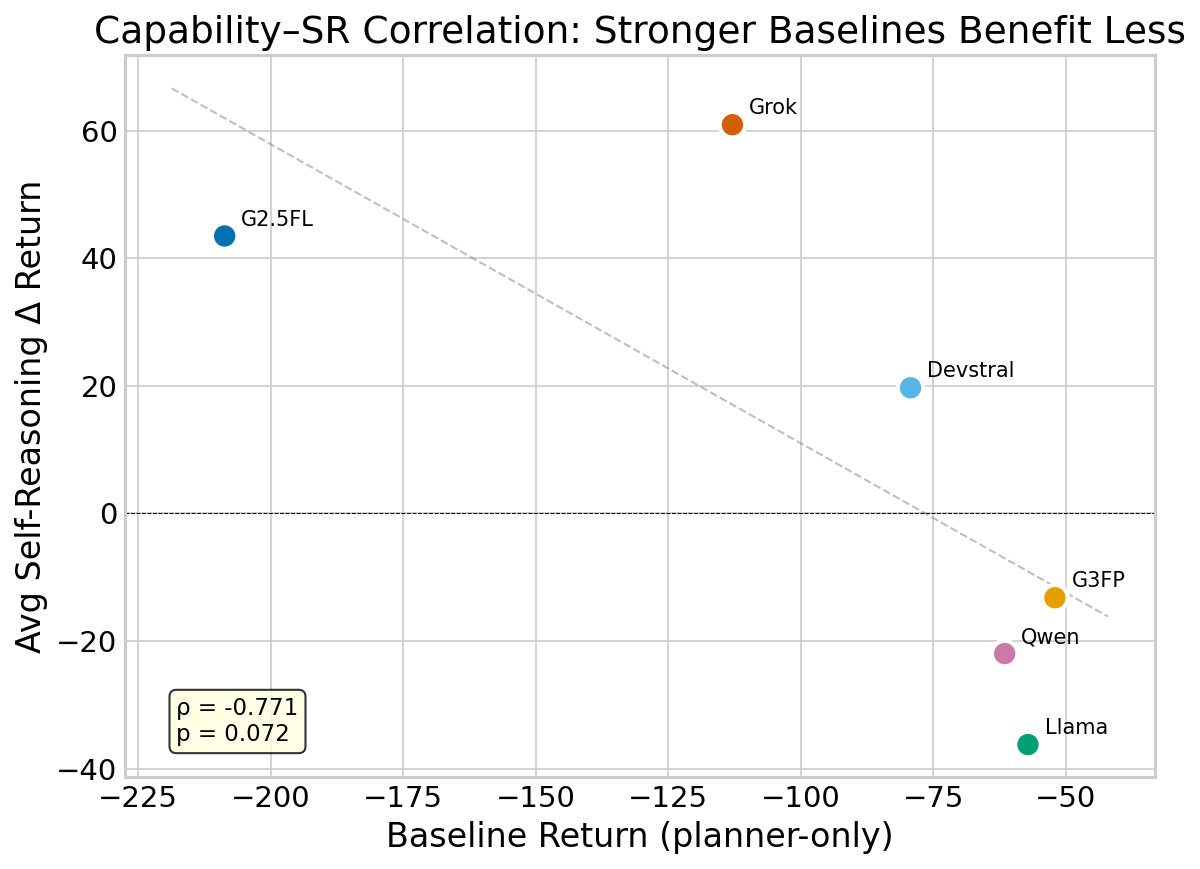}
\caption{Capability correlation. Models with stronger baselines (right) benefit less from deliberation on average. Spearman correlation shown.}
\Description{Baseline capability vs deliberation benefit.}
\label{fig:capability-correlation}
\end{figure}

\begin{figure}[htbp]
\centering
\includegraphics[width=\linewidth]{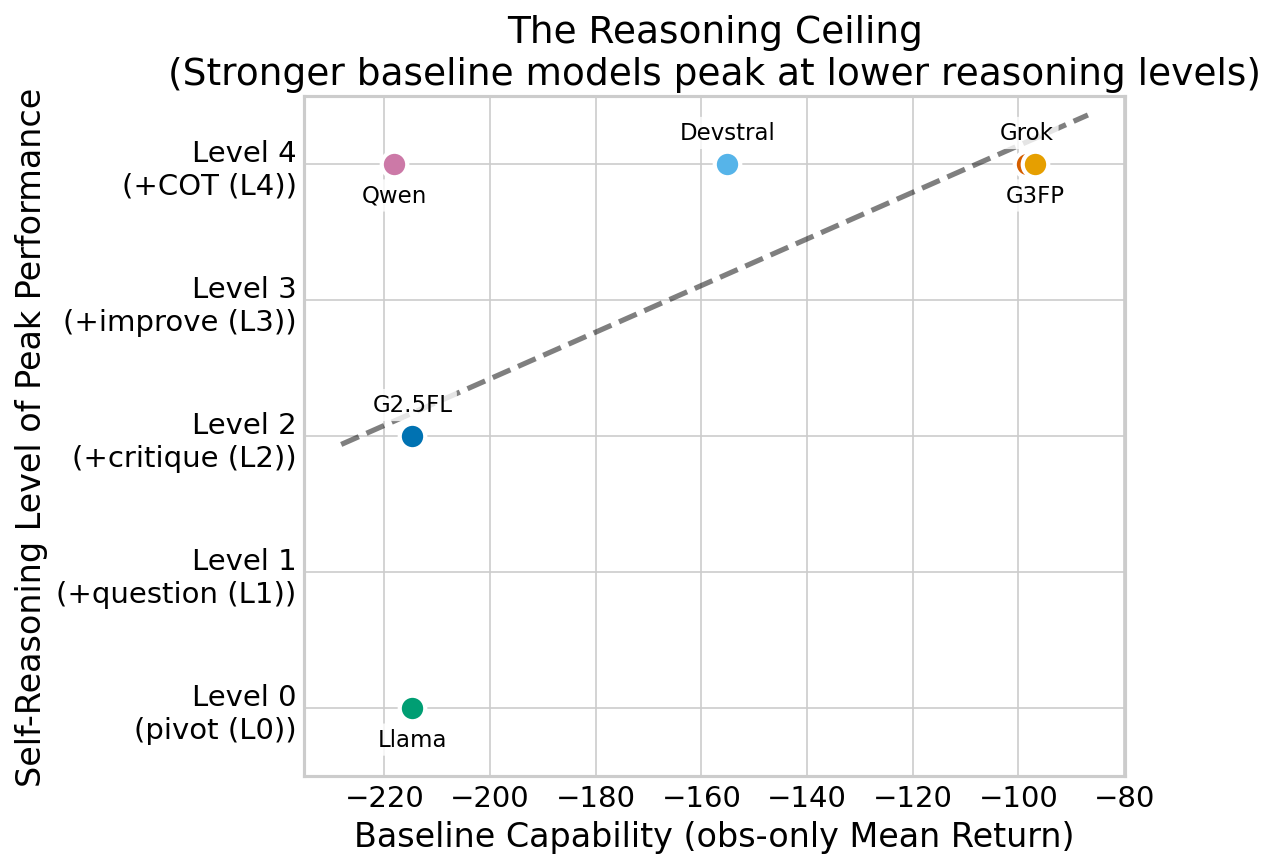}
\caption{The reasoning ceiling. Models with higher baseline capabilities tend to peak at lower levels of deliberation before degrading.}
\Description{Reasoning ceiling effect.}
\label{fig:reasoning-ceiling}
\end{figure}

\begin{figure}[htbp]
\centering
\includegraphics[width=\linewidth]{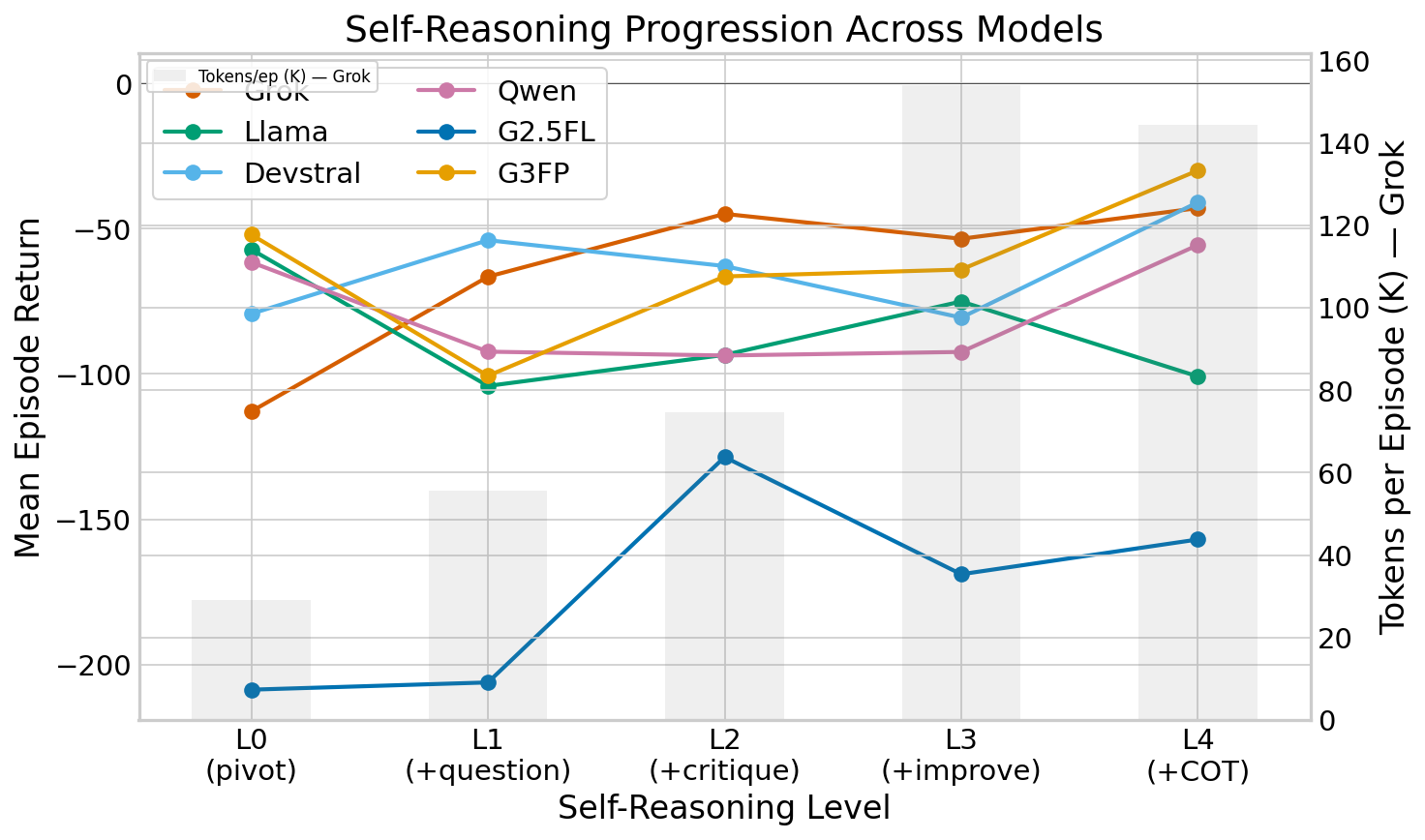}
\caption{Deliberation progression. Performance trajectory across cumulative deliberation levels for each model, showing non-monotonic patterns and model-dependent ceilings.}
\Description{Deliberation progression across cumulative levels for all six models.}
\label{fig:reasoning-progression}
\end{figure}

Tables~\ref{tab:reasoning-results-1} and~\ref{tab:reasoning-results-2}
break down the deliberation axis with per-level token consumption.
Table~\ref{tab:reasoning-roi} shows the return-on-investment for each
deliberation level. Figures~\ref{fig:sr-roi-scatter},
\ref{fig:capability-correlation}, and~\ref{fig:reasoning-ceiling} visualize
ROI, capability correlation, and the reasoning ceiling effect.
Figure~\ref{fig:reasoning-progression} shows the performance trajectory
across cumulative deliberation levels.

\begin{table*}[htbp]
  \centering
  \caption{Deliberation results (Grok, Llama, Devstral). Mean episode return, standard
    deviation, and tokens per episode across five deliberation levels. Best return
    per model is \textbf{bolded}.}
  \label{tab:reasoning-results-1}
  \begin{tabular}{l rrr rrr rrr}
    \toprule
    & \multicolumn{3}{c}{Grok} & \multicolumn{3}{c}{Llama} & \multicolumn{3}{c}{Devstral} \\
    \cmidrule(lr){2-4}    \cmidrule(lr){5-7}    \cmidrule(lr){8-10}
    Config & Mean & Std & Tok & Mean & Std & Tok & Mean & Std & Tok \\
    \midrule
    hist+net & $-112.9$ & 78.7 & 29.1K & $\textbf{-57.1}$ & 54.1 & 28.6K & $-79.3$ & 76.5 & 21.3K \\
    +question & $-66.5$ & 62.2 & 55.6K & $-104.1$ & 83.8 & 84.1K & $-53.9$ & 41.8 & 60.4K \\
    +critique & $-44.9$ & 44.6 & 74.7K & $-93.4$ & 77.3 & 97.8K & $-62.8$ & 58.0 & 106.9K \\
    +improve & $-53.4$ & 46.6 & 154.0K & $-75.0$ & 57.3 & 115.3K & $-80.6$ & 70.5 & 153.1K \\
    +COT & $\textbf{-42.9}$ & 32.8 & 144.5K & $-100.8$ & 70.1 & 131.3K & $\textbf{-40.9}$ & 31.1 & 157.4K \\
    \bottomrule
  \end{tabular}
\end{table*}

\begin{table*}[htbp]
  \centering
  \caption{Deliberation results (Qwen, G2.5FL, G3FP). Mean episode return, standard
    deviation, and tokens per episode across five deliberation levels. Best return
    per model is \textbf{bolded}.}
  \label{tab:reasoning-results-2}
  \begin{tabular}{l rrr rrr rrr}
    \toprule
    & \multicolumn{3}{c}{Qwen} & \multicolumn{3}{c}{G2.5FL} & \multicolumn{3}{c}{G3FP} \\
    \cmidrule(lr){2-4}    \cmidrule(lr){5-7}    \cmidrule(lr){8-10}
    Config & Mean & Std & Tok & Mean & Std & Tok & Mean & Std & Tok \\
    \midrule
    hist+net & $-61.5$ & 51.8 & 16.4K & $-208.7$ & 39.3 & 81.7K & $-52.0$ & 57.6 & 18.0K \\
    +question & $-92.3$ & 69.2 & 45.7K & $-206.2$ & 37.0 & 104.8K & $-100.6$ & 57.3 & 30.7K \\
    +critique & $-93.6$ & 67.1 & 68.9K & $\textbf{-128.6}$ & 94.8 & 118.2K & $-66.4$ & 63.4 & 41.0K \\
    +improve & $-92.4$ & 55.0 & 115.9K & $-168.9$ & 59.8 & 182.9K & $-64.0$ & 44.9 & 58.1K \\
    +COT & $\textbf{-55.6}$ & 43.2 & 162.4K & $-157.0$ & 69.7 & 225.7K & $\textbf{-29.9}$ & 19.7 & 75.1K \\
    \bottomrule
  \end{tabular}
\end{table*}

\begin{table*}[htbp]
  \centering
  \caption{Deliberation return on investment. Shows the reward change
    ($\Delta$) and token increase ($\Delta$ Tok) relative to the planner-only
    anchor for each reasoning level. Positive $\Delta$ = improvement.}
  \label{tab:reasoning-roi}
  \begin{tabular}{l rr rr rr rr rr rr}
    \toprule
    & \multicolumn{2}{c}{Grok} & \multicolumn{2}{c}{Llama} & \multicolumn{2}{c}{Devstral} & \multicolumn{2}{c}{Qwen} & \multicolumn{2}{c}{G2.5FL} & \multicolumn{2}{c}{G3FP} \\
    \cmidrule(lr){2-3}    \cmidrule(lr){4-5}    \cmidrule(lr){6-7}    \cmidrule(lr){8-9}    \cmidrule(lr){10-11}    \cmidrule(lr){12-13}
    Level & $\Delta$Ret & $\Delta$Tok & $\Delta$Ret & $\Delta$Tok & $\Delta$Ret & $\Delta$Tok & $\Delta$Ret & $\Delta$Tok & $\Delta$Ret & $\Delta$Tok & $\Delta$Ret & $\Delta$Tok \\
    \midrule
    +question & \textbf{+46.4} & +26.5K & -47.0 & +55.5K & \textbf{+25.3} & +39.0K & -30.8 & +29.2K & \textbf{+2.5} & +23.2K & -48.6 & +12.7K \\
    +critique & \textbf{+68.0} & +45.6K & -36.3 & +69.3K & \textbf{+16.4} & +85.6K & -32.1 & +52.5K & \textbf{+80.1} & +36.5K & -14.4 & +23.0K \\
    +improve & \textbf{+59.4} & +124.9K & -17.9 & +86.7K & -1.3 & +131.7K & -30.9 & +99.4K & \textbf{+39.8} & +101.2K & -12.0 & +40.2K \\
    +COT & \textbf{+70.0} & +115.4K & -43.7 & +102.7K & \textbf{+38.4} & +136.1K & \textbf{+5.9} & +146.0K & \textbf{+51.7} & +144.0K & \textbf{+22.1} & +57.1K \\
    \bottomrule
  \end{tabular}
\end{table*}

\subsection{Cross-Axis Comparisons}\label{app:results-cross}

\begin{figure}[htbp]
\centering
\includegraphics[width=\linewidth]{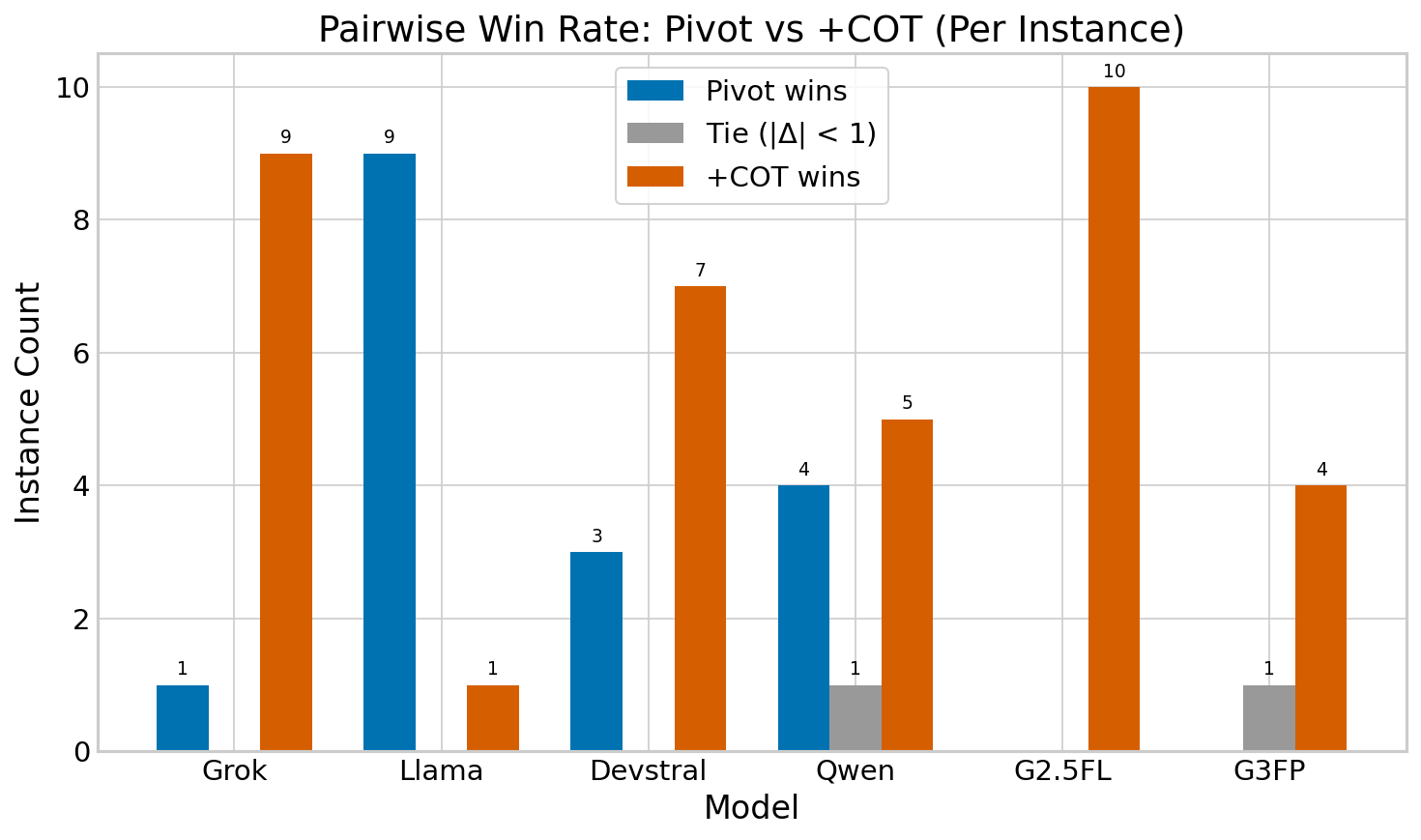}
\caption{Pairwise win rates. Anchor vs +COT compared instance-by-instance. Blue = anchor wins, orange = +COT wins, gray = ties.}
\Description{Pairwise win rates visualization.}
\label{fig:win-rates}
\end{figure}

\begin{figure}[htbp]
\centering
\includegraphics[width=\linewidth]{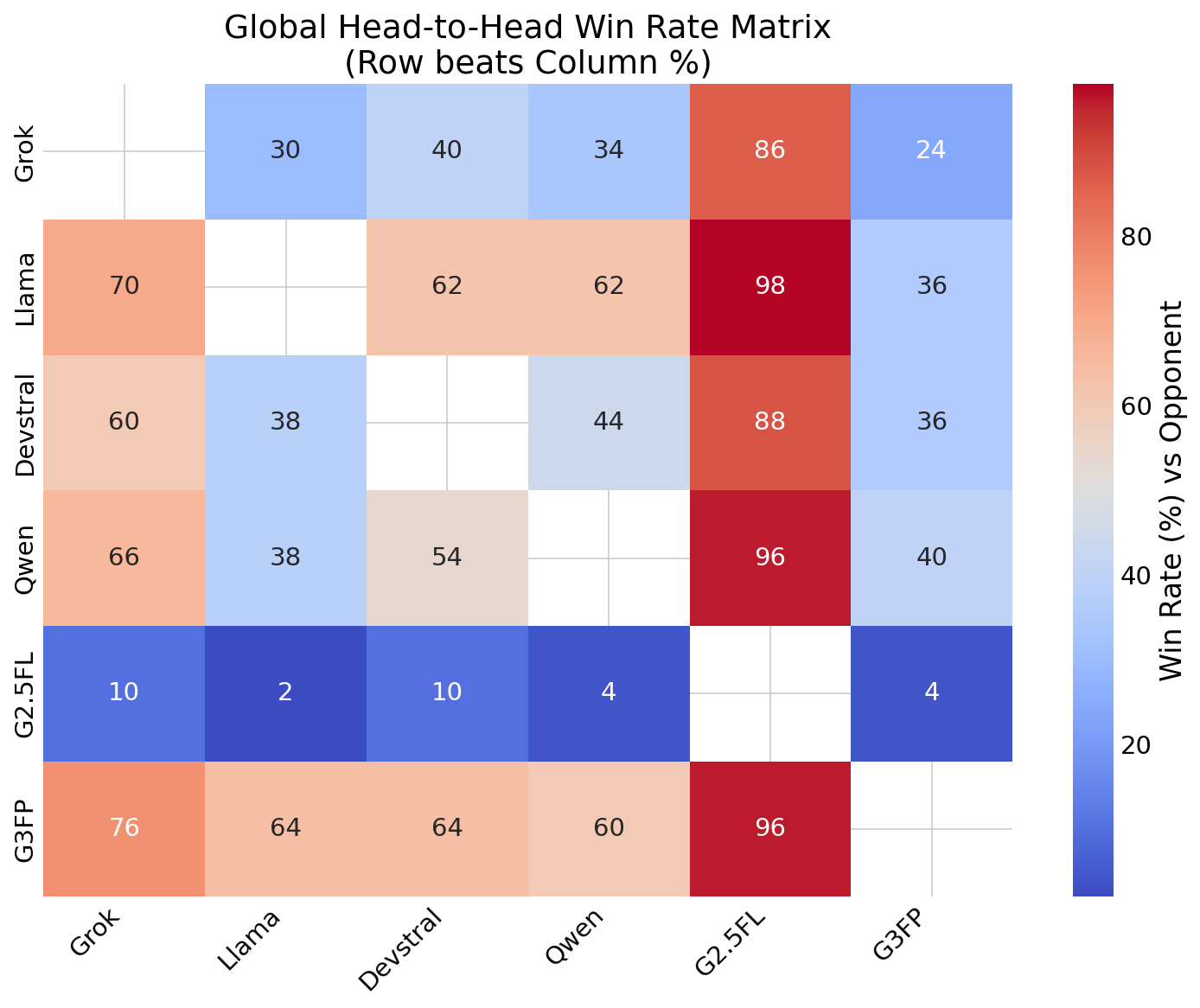}
\caption{Global head-to-head win-rate matrix. Compares the peak configuration of every model against every other model on a per-instance basis.}
\Description{Global win-rate matrix heatmap.}
\label{fig:global-win-matrix}
\end{figure}

\begin{figure}[htbp]
\centering
\includegraphics[width=\linewidth]{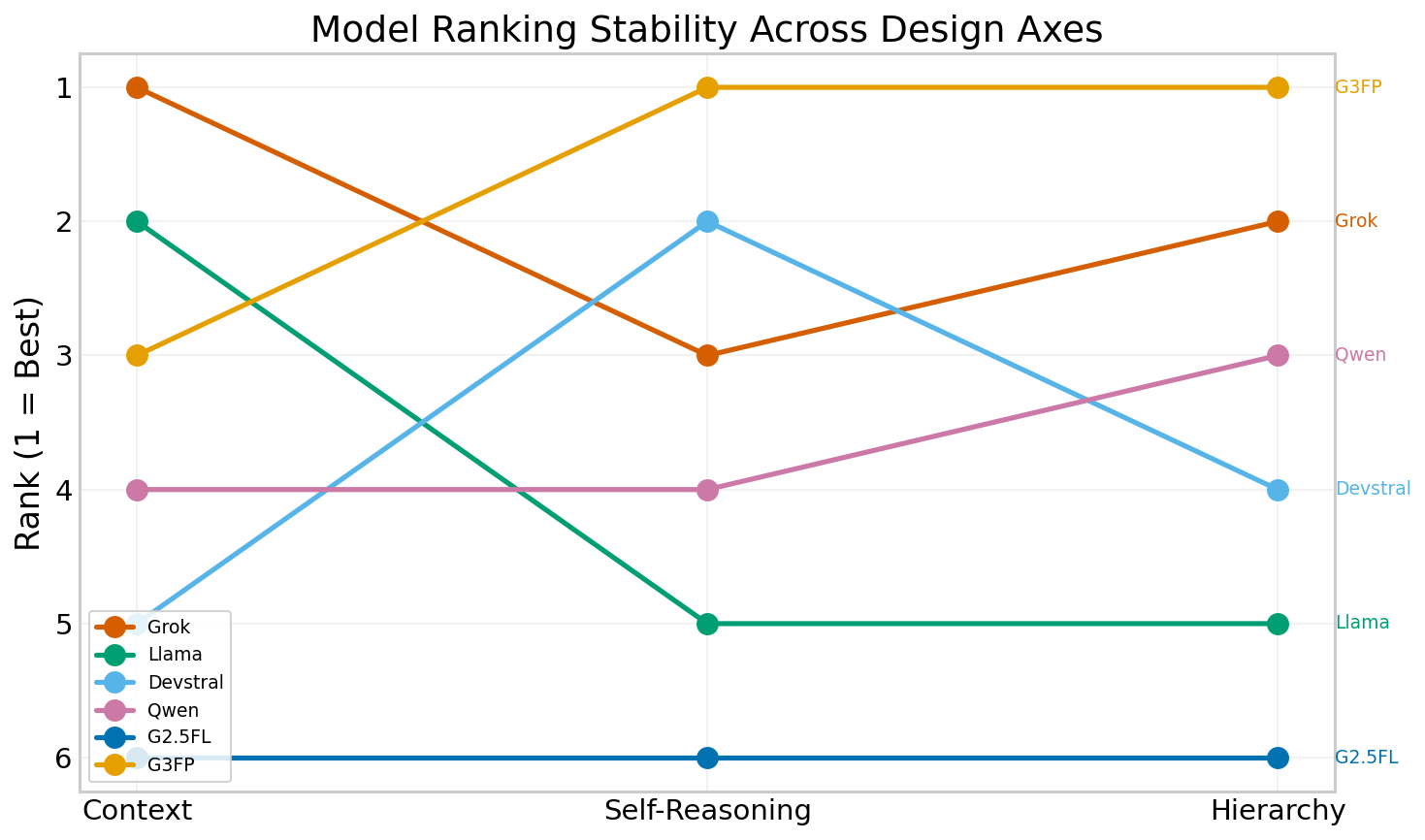}
\caption{Model ranking stability. Lines connect each model's best-config rank across the three axes. Flat lines = consistent relative performance.}
\Description{Model ranking stability across axes.}
\label{fig:ranking-stability}
\end{figure}

Table~\ref{tab:baseline} compares the observation-only baseline against each
model's best and worst configuration.
Tables~\ref{tab:context-best-worst}, \ref{tab:reasoning-best-worst},
and~\ref{tab:hierarchy-best-worst} identify best and worst configurations
within each axis. Table~\ref{tab:win-rates} and Figure~\ref{fig:win-rates}
present pairwise win rates; Figure~\ref{fig:global-win-matrix} provides a
head-to-head win-rate matrix using each model's peak configuration.
Table~\ref{tab:group-comparison} summarizes performance by configuration group.
Figure~\ref{fig:ranking-stability} and Table~\ref{tab:model-rankings}
demonstrate ranking stability across axes. Table~\ref{tab:anchor} reports
anchor-configuration performance.

\begin{table*}[htbp]
  \caption{Baseline Performance (Planner + Observation Only). The simplest
    configuration compared against each model's best and worst overall
    configuration across all 12 options.}
  \label{tab:baseline}
  \centering
  \begin{tabular}{lrrrr lr lr r}
    \toprule
    & \multicolumn{4}{c}{\textbf{Baseline (obs only)}} & \multicolumn{2}{c}{\textbf{Worst Config}} & \multicolumn{2}{c}{\textbf{Best Config}} & \\
    \cmidrule(lr){2-5}\cmidrule(lr){6-7}\cmidrule(lr){8-9}
    \textbf{Model} & \textbf{Runs} & \textbf{Return} & \textbf{Std} & \textbf{Min} & \textbf{Config} & \textbf{Return} & \textbf{Config} & \textbf{Return} & \textbf{Improv.} \\
    \midrule
    G3FP & 25 & $-96.8$ & $\pm$70.7 & $-200.8$ & network & $-136.4$ & hier-base & $-16.1$ & 88\% \\
    Grok & 50 & $-98.4$ & $\pm$69.8 & $-225.8$ & hist+net & $-112.9$ & hier-base & $-24.0$ & 79\% \\
    Devstral & 50 & $-155.1$ & $\pm$64.7 & $-225.8$ & obs & $-155.1$ & hier-base & $-37.8$ & 76\% \\
    Llama & 50 & $-214.7$ & $\pm$22.5 & $-225.2$ & obs & $-214.7$ & obs+net & $-51.4$ & 76\% \\
    G2.5FL & 50 & $-214.7$ & $\pm$22.8 & $-225.8$ & network & $-215.4$ & +critique & $-128.6$ & 40\% \\
    Qwen & 50 & $-218.2$ & $\pm$19.8 & $-225.9$ & obs & $-218.2$ & hier-base & $-28.6$ & 87\% \\
    \bottomrule
  \end{tabular}
\end{table*}

\begin{table}[htbp]
  \caption{Pairwise Win Rates: Anchor vs.\ Full Deliberation (+COT).}
  \label{tab:win-rates}
  \centering
  \begin{tabular}{lrrrr}
    \toprule
    \textbf{Model} & \textbf{+COT Wins} & \textbf{Ties} & \textbf{+COT Losses} & \textbf{Total} \\
    \midrule
    Grok & 37 (74\%) & 1 & 12 (24\%) & 50 \\
    Llama & 15 (30\%) & 0 & 35 (70\%) & 50 \\
    Devstral & 33 (66\%) & 0 & 17 (34\%) & 50 \\
    Qwen & 27 (54\%) & 0 & 23 (46\%) & 50 \\
    G2.5FL & 34 (68\%) & 5 & 11 (22\%) & 50 \\
    G3FP & 14 (56\%) & 0 & 11 (44\%) & 25 \\
    \bottomrule
  \end{tabular}
\end{table}

\begin{table}[htbp]
  \centering
  \caption{Performance by config group. Mean and range of mean episode return
    across all models for each configuration group.}
  \label{tab:group-comparison}
  \begin{tabular}{l rrrr}
    \toprule
    \textbf{Group} & \textbf{Mean} & \textbf{Best} & \textbf{Worst} & \textbf{Avg Tok/ep} \\
    \midrule
    Anchor (net+hist) & $-95.2$ & $-52.0$ & $-208.7$ & 32.5K \\
    Context & $-113.9$ & $-47.0$ & $-218.2$ & 30.6K \\
    Deliberation & $-86.4$ & $-29.9$ & $-206.2$ & 106.8K \\
    Hierarchy & $-72.9$ & $-16.1$ & $-186.4$ & 162.4K \\
    \bottomrule
  \end{tabular}
\end{table}

\begin{table*}[htbp]
  \caption{Context Engineering: Best and Worst Configurations per model.}
  \label{tab:context-best-worst}
  \centering
  \begin{tabular}{l rr | rr | r}
    \toprule
    & \multicolumn{2}{c}{\textbf{Worst Context Config}} & \multicolumn{2}{c}{\textbf{Best Context Config}} & \\
    \cmidrule(lr){2-3}\cmidrule(lr){4-5}
    \textbf{Model} & \textbf{Config} & \textbf{Return} & \textbf{Config} & \textbf{Return} & \textbf{Improv.} \\
    \midrule
    Grok & hist+net & $-112.9$ & obs+net & $-47.0$ & 58\% \\
    Llama & obs & $-214.7$ & obs+net & $-51.4$ & 76\% \\
    Devstral & obs & $-155.1$ & obs+net & $-72.6$ & 53\% \\
    Qwen & obs & $-218.2$ & hist+net & $-61.5$ & 72\% \\
    G2.5FL & network & $-215.4$ & obs+hist+net & $-147.8$ & 31\% \\
    G3FP & network & $-136.4$ & hist+net & $-52.0$ & 62\% \\
    \bottomrule
  \end{tabular}
\end{table*}

\begin{table*}[htbp]
  \caption{Deliberation: Best and Worst Configurations per model.}
  \label{tab:reasoning-best-worst}
  \centering
  \begin{tabular}{l rr | rr | r}
    \toprule
    & \multicolumn{2}{c}{\textbf{Worst Reasoning Config}} & \multicolumn{2}{c}{\textbf{Best Reasoning Config}} & \\
    \cmidrule(lr){2-3}\cmidrule(lr){4-5}
    \textbf{Model} & \textbf{Config} & \textbf{Return} & \textbf{Config} & \textbf{Return} & \textbf{Improv.} \\
    \midrule
    Grok & hist+net & $-112.9$ & +COT & $-42.9$ & 62\% \\
    Llama & +question & $-104.1$ & hist+net & $-57.1$ & 45\% \\
    Devstral & +improve & $-80.6$ & +COT & $-40.9$ & 49\% \\
    Qwen & +critique & $-93.6$ & +COT & $-55.6$ & 41\% \\
    G2.5FL & hist+net & $-208.7$ & +critique & $-128.6$ & 38\% \\
    G3FP & +question & $-100.6$ & +COT & $-29.9$ & 70\% \\
    \bottomrule
  \end{tabular}
\end{table*}

\begin{table*}[htbp]
  \caption{Hierarchy: Best and Worst Configurations per model.}
  \label{tab:hierarchy-best-worst}
  \centering
  \begin{tabular}{l rr | rr | r}
    \toprule
    & \multicolumn{2}{c}{\textbf{Worst Hierarchy Config}} & \multicolumn{2}{c}{\textbf{Best Hierarchy Config}} & \\
    \cmidrule(lr){2-3}\cmidrule(lr){4-5}
    \textbf{Model} & \textbf{Config} & \textbf{Return} & \textbf{Config} & \textbf{Return} & \textbf{Improv.} \\
    \midrule
    Grok & hier-delib & $-40.4$ & hier-base & $-24.0$ & 41\% \\
    Llama & hier-delib & $-108.0$ & hier-base & $-69.5$ & 36\% \\
    Devstral & hier-delib & $-127.4$ & hier-base & $-37.8$ & 70\% \\
    Qwen & hier-delib & $-30.1$ & hier-base & $-28.6$ & 5\% \\
    G2.5FL & hier-delib & $-186.4$ & hier-base & $-183.1$ & 2\% \\
    G3FP & hier-delib & $-23.6$ & hier-base & $-16.1$ & 32\% \\
    \bottomrule
  \end{tabular}
\end{table*}

\begin{table}[htbp]
  \centering
  \caption{Model rankings by config group (1 = best).}
  \label{tab:model-rankings}
  \begin{tabular}{l cccc}
    \toprule
    \textbf{Model} & \textbf{Context} & \textbf{Delib.} & \textbf{Hierarchy} & \textbf{Overall} \\
    \midrule
    Grok & \textbf{1} & \textbf{1} & 3 & \textbf{1} \\
    Llama & 5 & 5 & 5 & 5 \\
    Devstral & 4 & 2 & 4 & 4 \\
    Qwen & 3 & 4 & 2 & 3 \\
    G2.5FL & 6 & 6 & 6 & 6 \\
    G3FP & 2 & 3 & \textbf{1} & 2 \\
    \bottomrule
  \end{tabular}
\end{table}

\begin{table}[htbp]
  \centering
  \caption{Anchor performance across all models.}
  \label{tab:anchor}
  \begin{tabular}{l rrrr}
    \toprule
    \textbf{Model} & \textbf{Mean} & \textbf{Std} & \textbf{Min} & \textbf{Tok/ep} \\
    \midrule
    G3FP & $\textbf{-52.0}$ & 57.6 & $-224.8$ & 18.0K \\
    Llama & $-57.1$ & 54.1 & $-222.5$ & 28.6K \\
    Qwen & $-61.5$ & 51.8 & $-199.8$ & 16.4K \\
    Devstral & $-79.3$ & 76.5 & $-224.8$ & 21.3K \\
    Grok & $-112.9$ & 78.7 & $-224.7$ & 29.1K \\
    G2.5FL & $-208.7$ & 39.3 & $-224.8$ & 81.7K \\
    \bottomrule
  \end{tabular}
\end{table}

\subsection{Distributional Analysis}\label{app:results-distributional}

\begin{figure}[htbp]
\centering
\includegraphics[width=\linewidth]{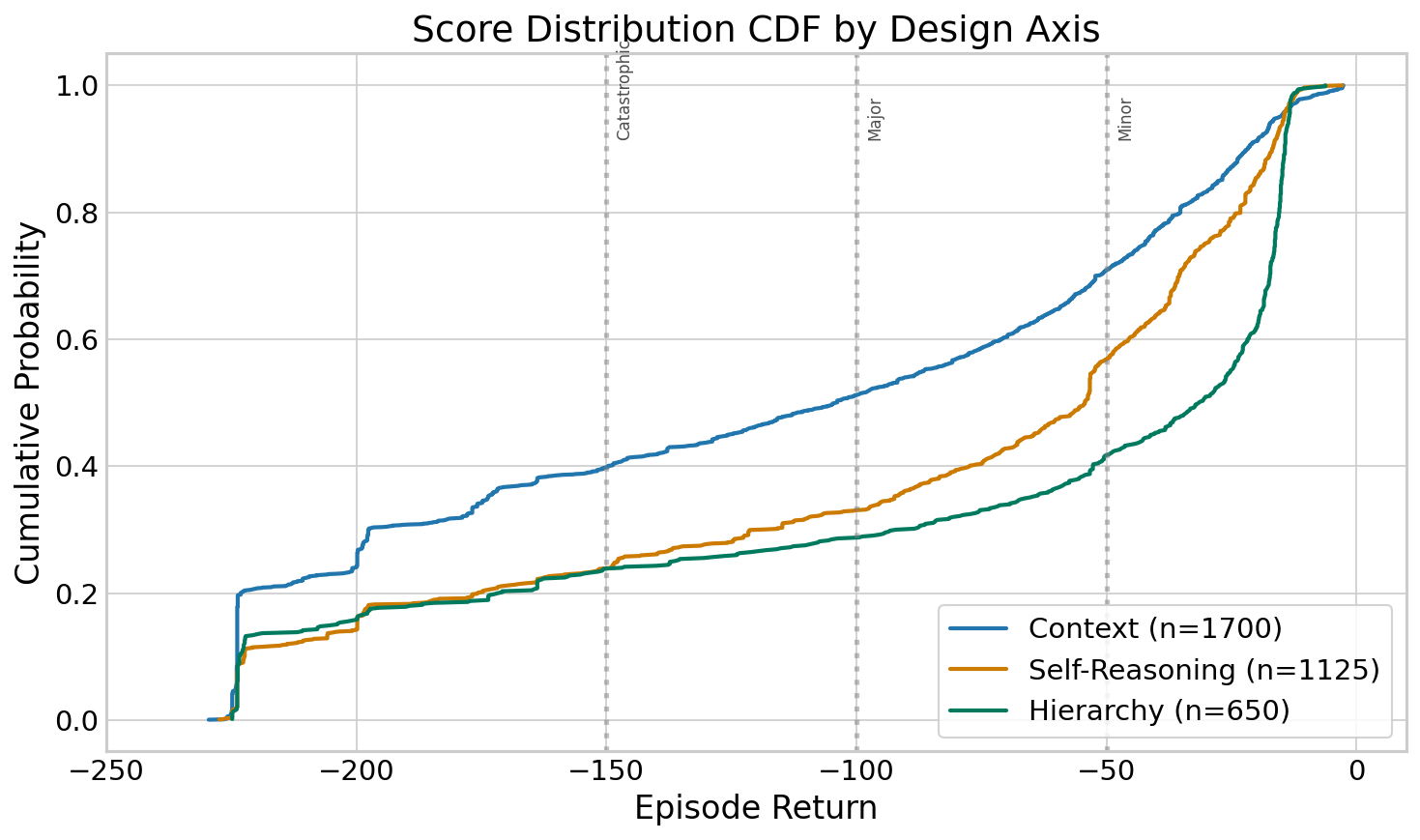}
\caption{Cumulative distributions by axis. Threshold lines mark failure severity. Right-shifted curves indicate better tail behavior.}
\Description{CDF by design axis.}
\label{fig:score-cdf}
\end{figure}

\begin{figure}[htbp]
\centering
\includegraphics[width=\linewidth]{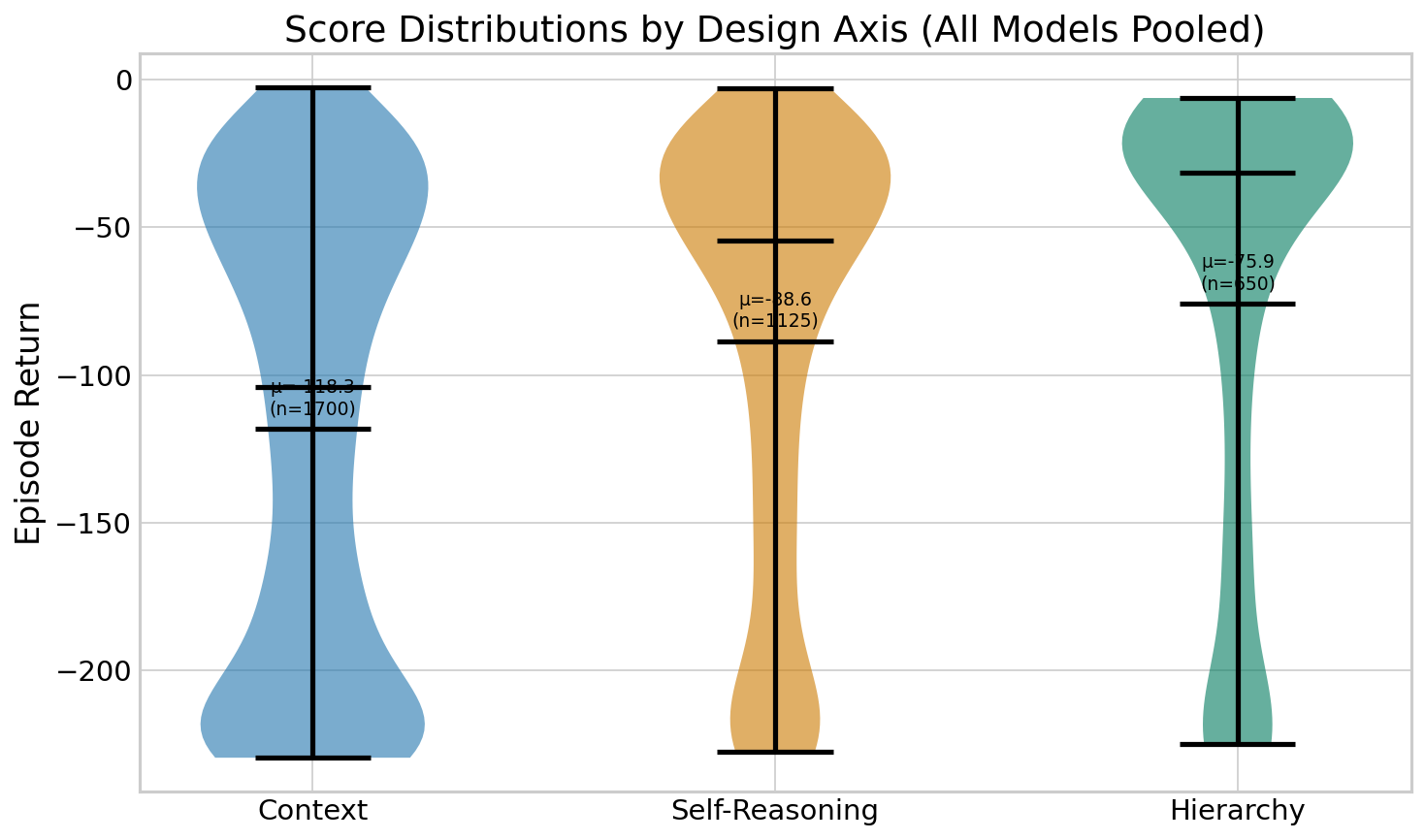}
\caption{Score distributions by design axis. Violins show full distribution with mean and median. Context configs have the widest spread.}
\Description{Score distributions by design axis.}
\label{fig:group-boxplot}
\end{figure}

\begin{figure}[htbp]
\centering
\includegraphics[width=\linewidth]{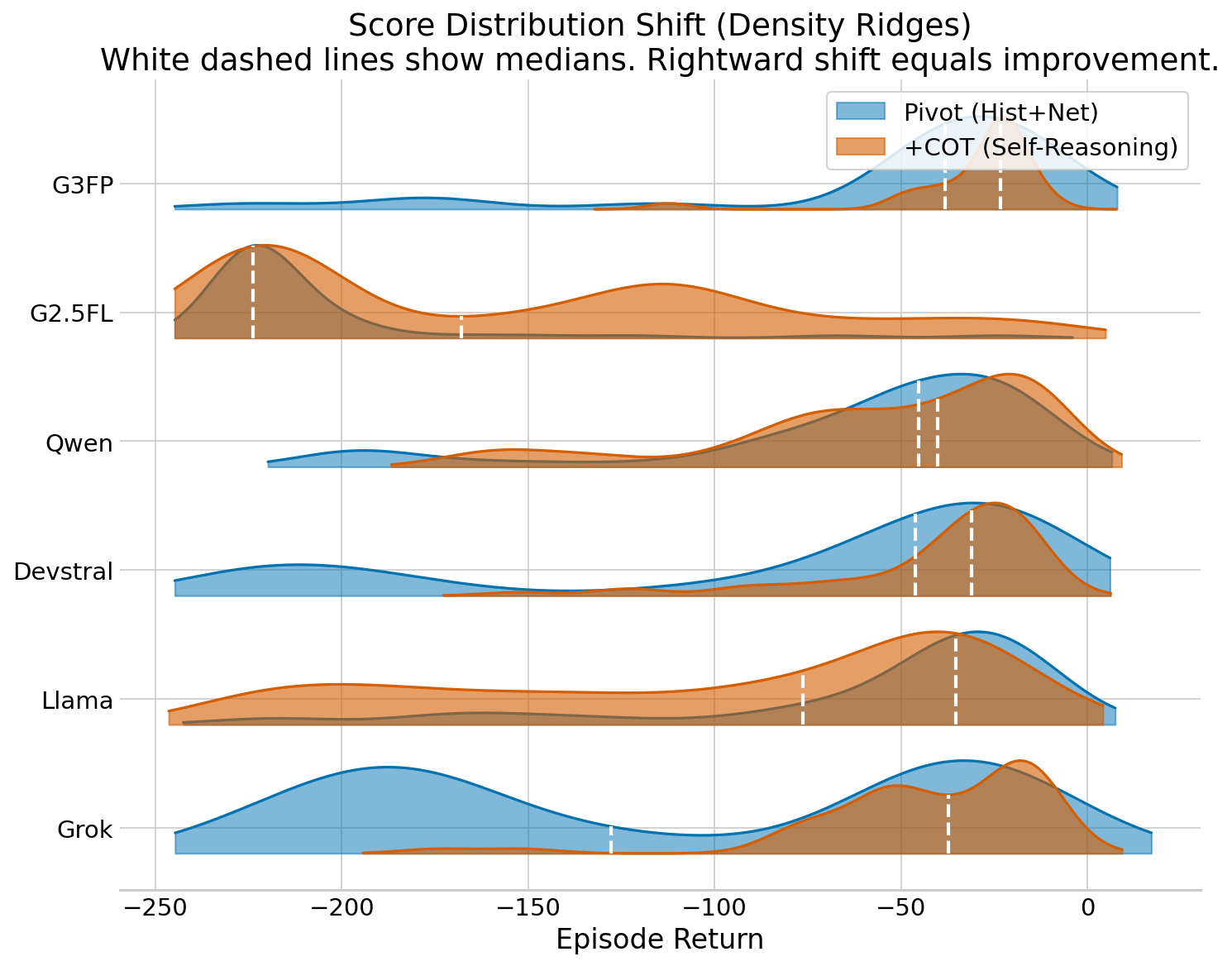}
\caption{Score distribution shift. Density ridges comparing the continuous probability distribution of episode returns for the Anchor baseline (blue) versus +COT (orange).}
\Description{Density ridges anchor vs COT.}
\label{fig:score-distribution-ridges}
\end{figure}

\begin{figure}[htbp]
\centering
\includegraphics[width=\linewidth]{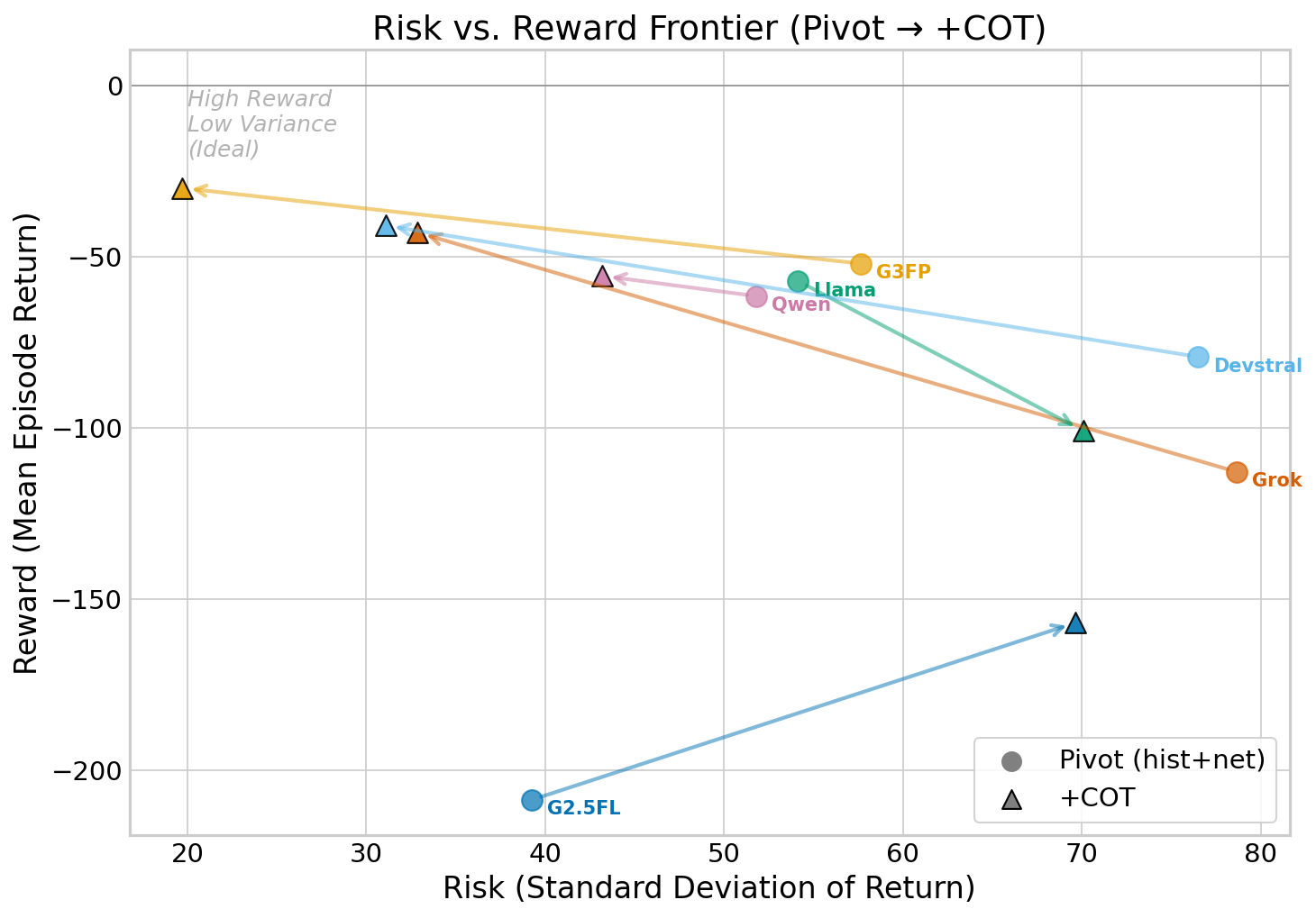}
\caption{Risk vs.\ Reward stability frontier. Shifts show how adding +COT changes both the mean return and variance. Ideally, arrows move up and to the left.}
\Description{Risk-reward frontier.}
\label{fig:risk-reward}
\end{figure}

\begin{figure}[htbp]
\centering
\includegraphics[width=\linewidth]{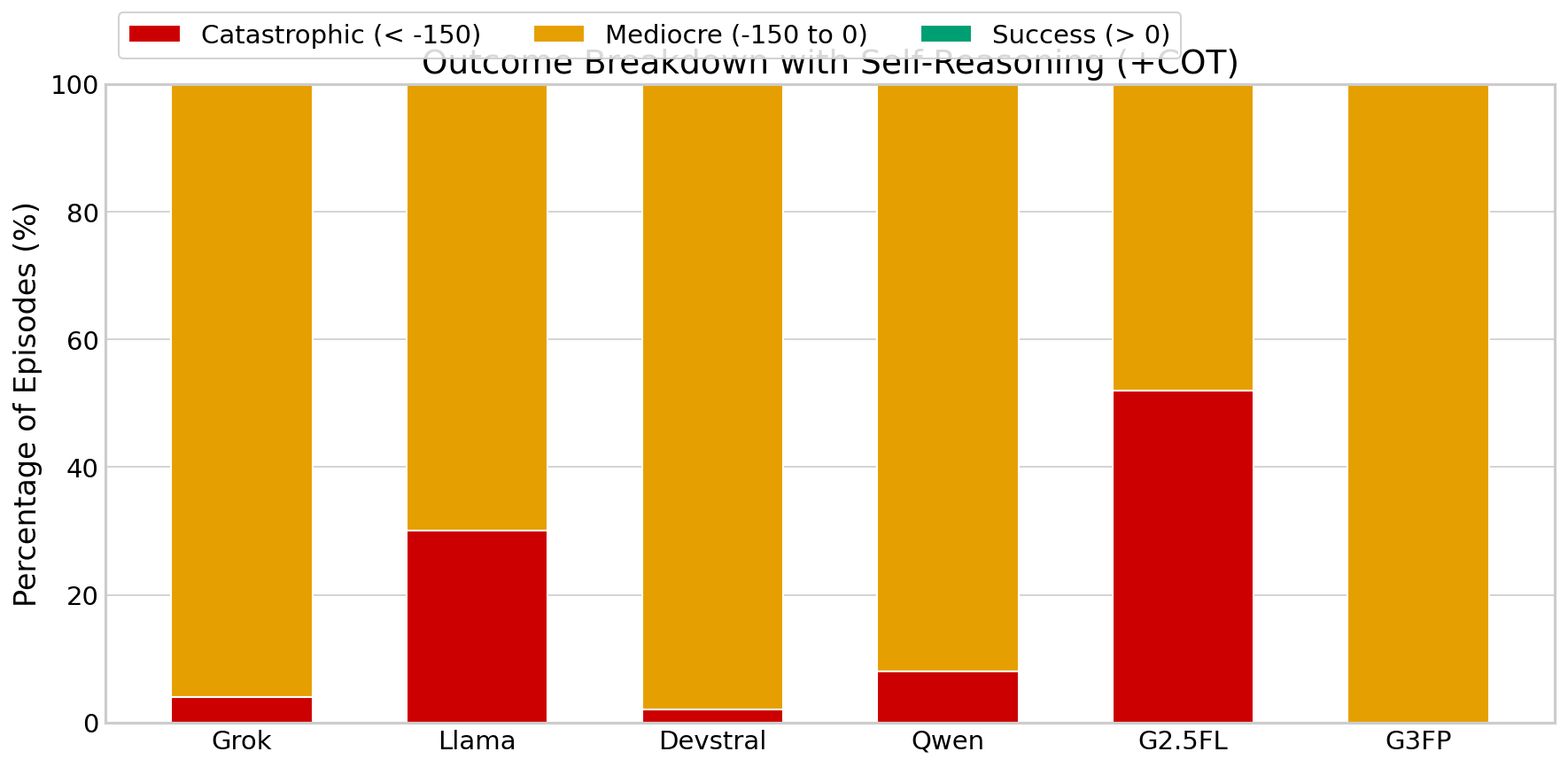}
\caption{Outcome breakdown for the +COT configuration. Shows the percentage of episodes resulting in success, mediocre failure, or catastrophic failure.}
\Description{Outcome breakdown per model for +COT.}
\label{fig:outcome-breakdown}
\end{figure}

Table~\ref{tab:variance-tail} reports standard deviation and worst-case
(minimum) episode return for every model--configuration pair, supporting the
robustness analysis in Section~\ref{sec:robustness}.
Table~\ref{tab:catastrophic-failures} reports catastrophic failure rates
(return $<-150$) for each configuration.
Figures~\ref{fig:score-cdf}, \ref{fig:group-boxplot},
\ref{fig:score-distribution-ridges}, \ref{fig:risk-reward},
and~\ref{fig:outcome-breakdown} provide distributional views of episode returns
across design axes.

\begin{table*}[htbp]
  \centering
  \caption{Variance and tail risk. Standard deviation and minimum (worst-case)
    episode return for each model--configuration pair.}
  \label{tab:variance-tail}
  \footnotesize
  \begin{tabular}{l l rr rr rr rr rr rr}
    \toprule
    & \multicolumn{2}{c}{Grok} & \multicolumn{2}{c}{Llama} & \multicolumn{2}{c}{Devstral} & \multicolumn{2}{c}{Qwen} & \multicolumn{2}{c}{G2.5FL} & \multicolumn{2}{c}{G3FP} \\
    \cmidrule(lr){3-4}    \cmidrule(lr){5-6}    \cmidrule(lr){7-8}    \cmidrule(lr){9-10}    \cmidrule(lr){11-12}    \cmidrule(lr){13-14}
    \textbf{Group} & \textbf{Config} & Std & Min & Std & Min & Std & Min & Std & Min & Std & Min & Std & Min \\
    \midrule
    Ctx & obs & 69.8 & $-225.8$ & 22.5 & $-225.2$ & 64.7 & $-225.8$ & 19.8 & $-225.9$ & 22.8 & $-225.8$ & 70.7 & $-200.8$ \\
     & obs+hist & 75.3 & $-224.8$ & 63.3 & $-224.7$ & 85.5 & $-225.3$ & 71.7 & $-224.8$ & 69.4 & $-229.5$ & 67.2 & $-223.8$ \\
     & obs+hist+net & 78.0 & $-224.8$ & 73.2 & $-222.4$ & 77.5 & $-224.8$ & 53.2 & $-224.8$ & 69.5 & $-225.7$ & 63.8 & $-224.8$ \\
     & obs+net & 40.3 & $-145.8$ & 19.9 & $-117.5$ & 47.7 & $-212.8$ & 25.9 & $-176.8$ & 40.8 & $-225.1$ & 68.5 & $-224.8$ \\
     & network & 30.9 & $-169.7$ & 37.5 & $-167.2$ & 44.2 & $-208.6$ & 58.8 & $-223.6$ & 18.5 & $-225.8$ & 49.1 & $-220.8$ \\
     & hist+net & 78.7 & $-224.7$ & 54.1 & $-222.5$ & 76.5 & $-224.8$ & 51.8 & $-199.8$ & 39.3 & $-224.8$ & 57.6 & $-224.8$ \\
    Delib. & +question & 62.2 & $-223.8$ & 83.8 & $-223.8$ & 41.8 & $-224.8$ & 69.2 & $-224.8$ & 37.0 & $-224.8$ & 57.3 & $-200.8$ \\
     & +critique & 44.6 & $-226.1$ & 77.3 & $-227.4$ & 58.0 & $-223.7$ & 67.1 & $-223.8$ & 94.8 & $-224.8$ & 63.4 & $-199.8$ \\
     & +improve & 46.6 & $-222.4$ & 57.3 & $-223.8$ & 70.5 & $-223.8$ & 55.0 & $-223.9$ & 59.8 & $-224.8$ & 44.9 & $-147.6$ \\
     & +COT & 32.8 & $-174.3$ & 70.1 & $-226.4$ & 31.1 & $-152.7$ & 43.2 & $-166.7$ & 69.7 & $-224.8$ & 19.7 & $-112.2$ \\
    Hier & hier-base & 27.0 & $-150.8$ & 60.0 & $-222.3$ & 37.2 & $-218.8$ & 36.6 & $-184.7$ & 62.4 & $-224.8$ & 2.7 & $-20.5$ \\
     & hier-delib & 26.1 & $-115.9$ & 74.3 & $-223.8$ & 71.8 & $-224.7$ & 32.7 & $-173.6$ & 58.1 & $-224.6$ & 12.4 & $-52.7$ \\
    \bottomrule
  \end{tabular}
\end{table*}

\begin{table*}[htbp]
  \centering
  \caption{Catastrophic failure rates. Percentage of episodes with return
    below $-150$ (indicating near-total network compromise). Lower is better.}
  \label{tab:catastrophic-failures}
  \footnotesize
  \begin{tabular}{l r r r r r r}
    \toprule
    \textbf{Config} & \textbf{Grok} & \textbf{Llama} & \textbf{Devstral} & \textbf{Qwen} & \textbf{G2.5FL} & \textbf{G3FP} \\
    \midrule
    obs & 28.0\% & \textbf{96.0\%} & \textbf{64.0\%} & \textbf{98.0\%} & \textbf{96.0\%} & 32.0\% \\
    obs+hist & 26.0\% & 50.0\% & \textbf{54.0\%} & 24.0\% & \textbf{72.0\%} & 16.0\% \\
    obs+hist+net & 24.0\% & 38.0\% & 26.0\% & 14.0\% & \textbf{58.0\%} & 12.0\% \\
    obs+net & 0.0\% & 0.0\% & 10.0\% & 2.0\% & \textbf{88.0\%} & 36.0\% \\
    network & 6.0\% & 4.0\% & 10.0\% & 32.0\% & \textbf{98.0\%} & 24.0\% \\
    hist+net & 44.0\% & 10.0\% & 24.0\% & 10.0\% & \textbf{92.0\%} & 12.0\% \\
    +question & 16.0\% & 38.0\% & 4.0\% & 24.0\% & \textbf{96.0\%} & 22.0\% \\
    +critique & 6.0\% & 32.0\% & 12.0\% & 20.0\% & 48.0\% & 12.0\% \\
    +improve & 6.0\% & 12.0\% & 22.0\% & 18.0\% & \textbf{60.0\%} & 0.0\% \\
    +COT & 4.0\% & 30.0\% & 2.0\% & 8.0\% & \textbf{52.0\%} & 0.0\% \\
    hier-base & 2.0\% & 14.0\% & 2.0\% & 5.0\% & \textbf{78.7\%} & 0.0\% \\
    hier-delib & 0.0\% & 36.0\% & 44.0\% & 4.0\% & \textbf{80.0\%} & 0.0\% \\
    \bottomrule
  \end{tabular}
\end{table*}

\subsection{Token Cost Progression}\label{app:results-tokens}

\begin{figure}[h]
\centering
\includegraphics[width=\linewidth]{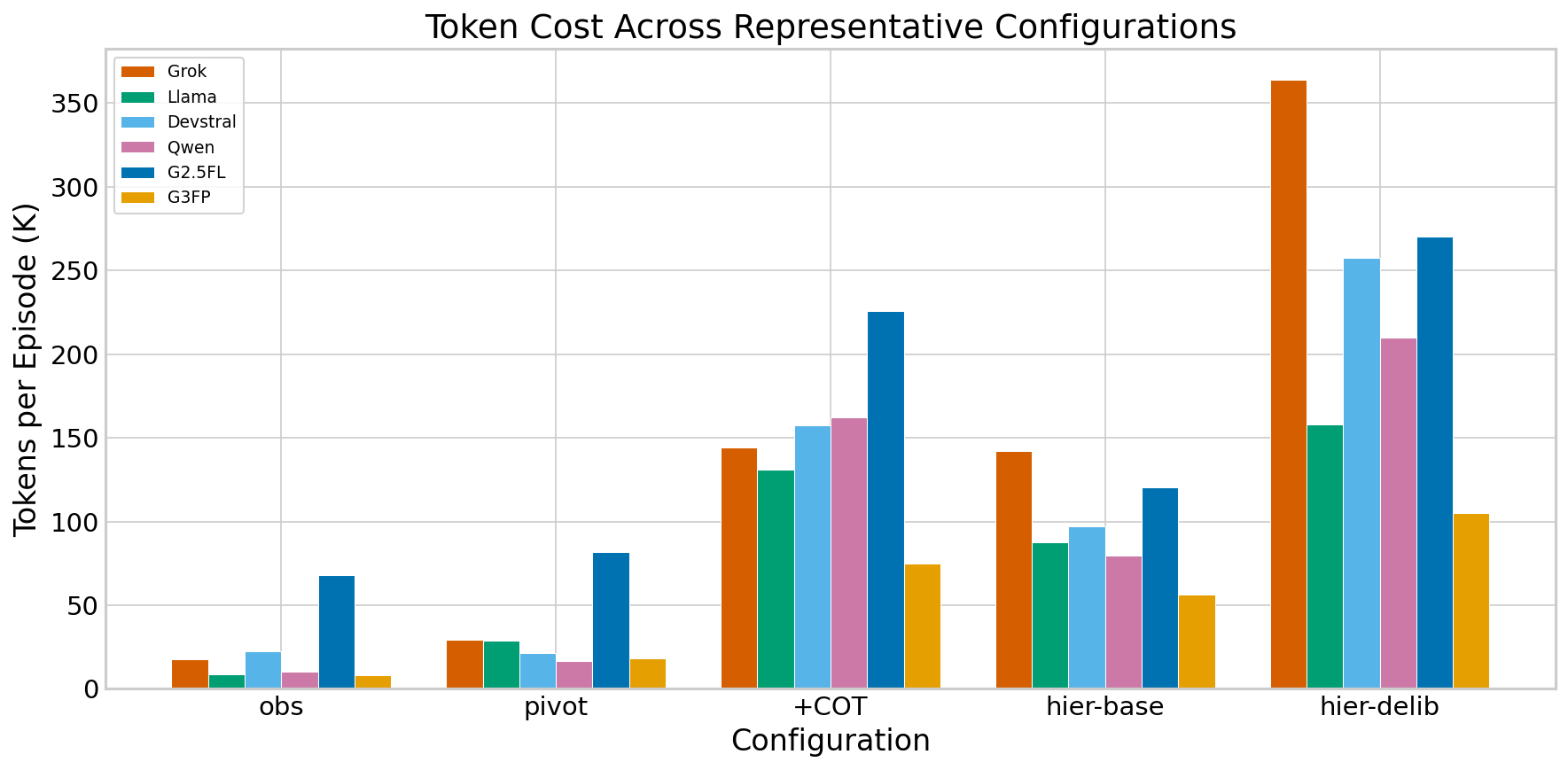}
\caption{Token cost progression from cheapest (\texttt{obs}) to most expensive (\texttt{hier-delib}). Deliberation and hierarchy dramatically increase token consumption; the deliberation cascade represents the cost ceiling.}
\Description{Bar chart showing token cost per episode from cheapest to most expensive configuration type.}
\label{fig:token-waterfall}
\end{figure}

Figure~\ref{fig:token-waterfall} shows the token cost progression from
cheapest (\texttt{obs}) to most expensive (\texttt{hier-delib}) configuration.

\section{Statistical Support}\label{app:statistical}

This appendix provides 95\% confidence intervals for mean episode returns (Table~\ref{tab:full-results-ci}) and paired mean-return differences for key comparisons (Table~\ref{tab:paired-diff-ci}).


\begin{table*}[htbp]
    \centering
    \caption{Full results matrix with 95\% confidence intervals.
      Mean episode return ($\pm$ CI half-width) across all 72
      model--configuration pairs.  Best point-estimate return per model is
      \textbf{bolded}.  Configurations are grouped by experimental axis.}
    \label{tab:full-results-ci}
    \footnotesize
    \begin{tabular}{l l r r r r r r}
      \toprule
      \textbf{Group} & \textbf{Config} & \textbf{Grok} & \textbf{Llama} & \textbf{Devstral} & \textbf{Qwen} & \textbf{G2.5FL} & \textbf{G3FP} \\
      \midrule
      Context & obs & $-98.4$$\pm$19.8 & $-214.7$$\pm$6.4 & $-155.1$$\pm$18.4 & $-218.2$$\pm$5.6 & $-214.7$$\pm$6.5 & $-96.8$$\pm$29.2 \\
       & obs+hist & $-89.9$$\pm$21.4 & $-137.4$$\pm$18.0 & $-133.9$$\pm$24.3 & $-93.4$$\pm$20.4 & $-172.9$$\pm$19.7 & $-76.3$$\pm$27.8 \\
       & obs+hist+net & $-81.9$$\pm$22.2 & $-102.6$$\pm$20.8 & $-85.3$$\pm$22.0 & $-69.0$$\pm$15.1 & $-147.8$$\pm$19.8 & $-82.6$$\pm$26.3 \\
       & obs+net & $-47.0$$\pm$11.4 & $\textbf{-51.4}$$\pm$5.7 & $-72.6$$\pm$13.6 & $-63.1$$\pm$7.4 & $-200.0$$\pm$8.1 & $-113.7$$\pm$28.3 \\
       & network & $-86.3$$\pm$8.8 & $-68.7$$\pm$10.7 & $-93.3$$\pm$12.6 & $-109.4$$\pm$16.7 & $-215.4$$\pm$5.3 & $-136.4$$\pm$20.3 \\
       & hist+net & $-112.9$$\pm$22.4 & $-57.1$$\pm$15.4 & $-79.3$$\pm$21.7 & $-61.5$$\pm$14.7 & $-208.7$$\pm$11.2 & $-52.0$$\pm$23.8 \\
      Self-Reas. & +question & $-66.5$$\pm$17.7 & $-104.1$$\pm$23.8 & $-53.9$$\pm$11.9 & $-92.3$$\pm$19.7 & $-206.2$$\pm$10.5 & $-100.6$$\pm$16.3 \\
       & +critique & $-44.9$$\pm$12.7 & $-93.4$$\pm$22.0 & $-62.8$$\pm$16.5 & $-93.6$$\pm$19.1 & $\textbf{-128.6}$$\pm$27.0 & $-66.4$$\pm$26.2 \\
       & +improve & $-53.4$$\pm$13.2 & $-75.0$$\pm$16.3 & $-80.6$$\pm$20.0 & $-92.4$$\pm$15.6 & $-168.9$$\pm$17.0 & $-64.0$$\pm$18.5 \\
       & +COT & $-42.9$$\pm$9.3 & $-100.8$$\pm$19.9 & $-40.9$$\pm$8.8 & $-55.6$$\pm$12.3 & $-157.0$$\pm$19.8 & $-29.9$$\pm$8.1 \\
      Hierarchy & hier-base & $\textbf{-24.0}$$\pm$7.7 & $-69.5$$\pm$17.1 & $\textbf{-37.8}$$\pm$10.6 & $\textbf{-28.6}$$\pm$7.3 & $-183.1$$\pm$14.4 & $\textbf{-16.1}$$\pm$1.1 \\
       & hier-delib & $-40.4$$\pm$7.4 & $-108.0$$\pm$21.1 & $-127.4$$\pm$20.4 & $-30.1$$\pm$9.3 & $-186.4$$\pm$16.5 & $-23.6$$\pm$3.5 \\
      \bottomrule
    \end{tabular}
  \end{table*}

  
  \begin{table*}[htbp]
    \centering
    \caption{Paired mean-return differences with 95\% confidence intervals.
      Each cell shows $\Delta$\,=\,mean paired difference $\pm$ CI half-width
      (matched by instance$\times$run; duplicate episodes averaged before pairing).
      \textbf{Bold} indicates the 95\% CI excludes zero;
      positive $\Delta$ means the first-named configuration yields higher
      return.  $^\dagger$Post-hoc selected:
      \textsf{best-ctx} is the highest-return structured context per model
      (excluding raw \texttt{obs});
      \textsf{best-SR} is the highest-return monolithic self-reasoning level
      (\texttt{+question}\ldots\texttt{+COT}, excluding the \texttt{hist+net}
      anchor).}
    \label{tab:paired-diff-ci}
    \footnotesize
    \begin{tabular}{l r r r r r r}
      \toprule
      \textbf{Comparison} & \textbf{Grok} & \textbf{Llama} & \textbf{Devstral} & \textbf{Qwen} & \textbf{G2.5FL} & \textbf{G3FP} \\
      \midrule
      hier-delib $-$ hier-base & $\textbf{-16.4}$$\pm$11.5 & $\textbf{-38.5}$$\pm$28.7 & $\textbf{-89.6}$$\pm$23.1 & $-1.6$$\pm$11.9 & $-0.1$$\pm$22.7 & $\textbf{-7.5}$$\pm$3.6 \\
      obs+net $-$ obs & $\textbf{+51.4}$$\pm$23.3 & $\textbf{+163.3}$$\pm$7.9 & $\textbf{+82.5}$$\pm$24.5 & $\textbf{+155.1}$$\pm$9.1 & $\textbf{+14.8}$$\pm$8.5 & $-16.8$$\pm$41.1 \\
      best-ctx$^\dagger$ $-$ obs & $\textbf{+51.4}$$\pm$23.3 & $\textbf{+163.3}$$\pm$7.9 & $\textbf{+82.5}$$\pm$24.5 & $\textbf{+156.7}$$\pm$15.8 & $\textbf{+66.9}$$\pm$19.0 & $\textbf{+44.8}$$\pm$32.1 \\
      hier-base $-$ hist+net & $\textbf{+88.8}$$\pm$25.1 & $-12.4$$\pm$23.2 & $\textbf{+41.4}$$\pm$22.8 & $\textbf{+32.9}$$\pm$17.7 & $\textbf{+22.5}$$\pm$13.5 & $\textbf{+35.9}$$\pm$24.0 \\
      hier-delib $-$ best-SR$^\dagger$ & $+2.4$$\pm$10.3 & $\textbf{-33.0}$$\pm$30.1 & $\textbf{-86.5}$$\pm$22.3 & $\textbf{+25.4}$$\pm$15.0 & $\textbf{-57.8}$$\pm$28.6 & $+6.3$$\pm$9.0 \\
      \bottomrule
    \end{tabular}
  \end{table*}

  \clearpage
\section{Token Consumption}\label{app:tokens}

This appendix details per-model token profiles and token efficiency across configurations.
Tables~\ref{tab:token-eff-grok} through~\ref{tab:token-eff-g3fp} report a scalar
\emph{shifted return-per-kilotoken} efficiency, $\tilde{G}/\mathrm{KTok}$, where
$\tilde{G}=225+G$ converts the non-positive episodic return $G$ into a non-negative
"defense score" (higher is better), and $\mathrm{KTok}$ is tokens per episode in thousands.
This scalar is provided as a compact summary; our primary cost--performance comparisons
use Pareto frontiers in the main text. We also performed a pricing sensitivity check by re-weighting token costs using provider-specific input/output pricing ratios. This re-weighting did not reverse any qualitative conclusion; it narrowed the relative cost advantage of context over hierarchy because simpler context configurations contain a higher share of output tokens, but the qualitative ordering was preserved.
 
Table~\ref{tab:token-profile} breaks down the prompt vs.\ completion token
split for the anchor baseline and the +COT configuration. Deliberation
dramatically increases prompt tokens (due to multi-turn tool-call
exchanges) and moderately increases completion tokens.
The prompt-to-completion ratio shifts from roughly $2$--$24\times$ at baseline
to $4$--$31\times$ under +COT, indicating that deliberation overhead is
dominated by the expansion of the conversational context rather than by longer
model outputs.

\begin{figure}[htbp]
\centering
\includegraphics[width=\linewidth]{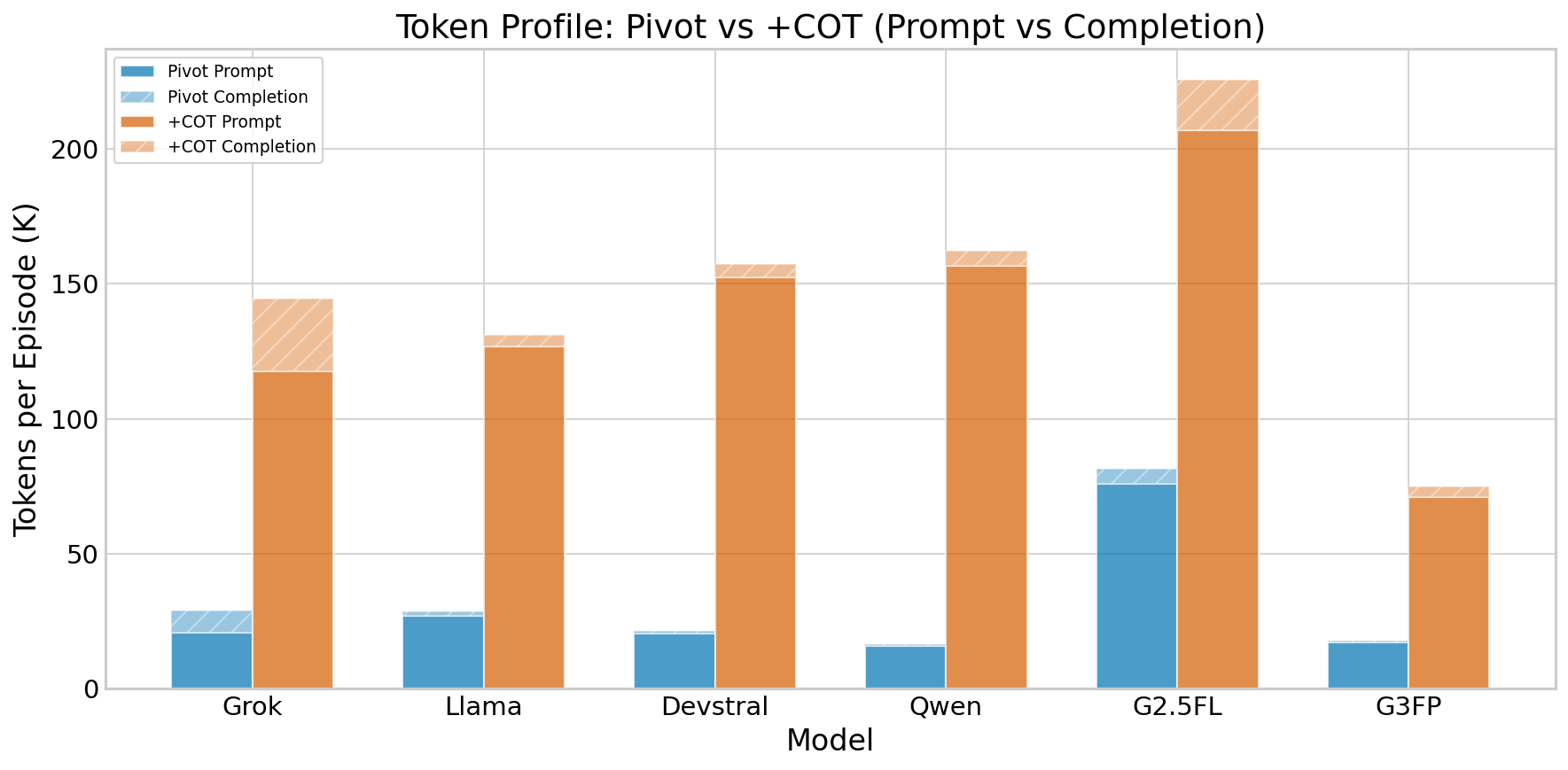}
\caption{Token profile shift. Stacked bars show prompt (solid) vs completion (hatched) for anchor and +COT. Deliberation increases both components.}
\Description{Token profile shift visualization.}
\label{fig:token-profile}
\end{figure}

\begin{figure}[htbp]
\centering
\includegraphics[width=\linewidth]{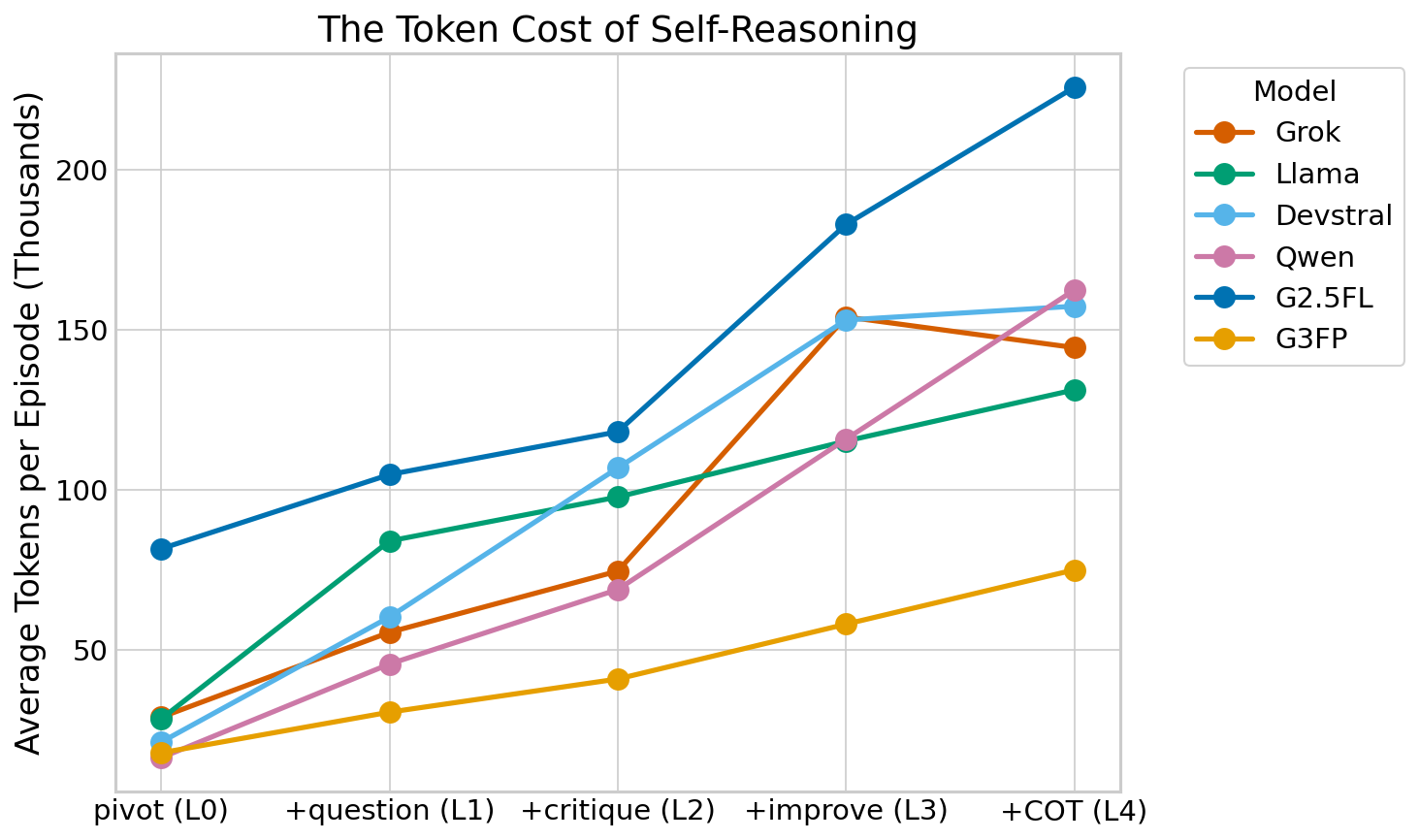}
\caption{Token generation velocity. Shows the exponential increase in token usage as cumulative deliberation capabilities are added.}
\Description{Token velocity across reasoning levels.}
\label{fig:token-velocity}
\end{figure}

\begin{figure}[htbp]
\centering
\includegraphics[width=\linewidth]{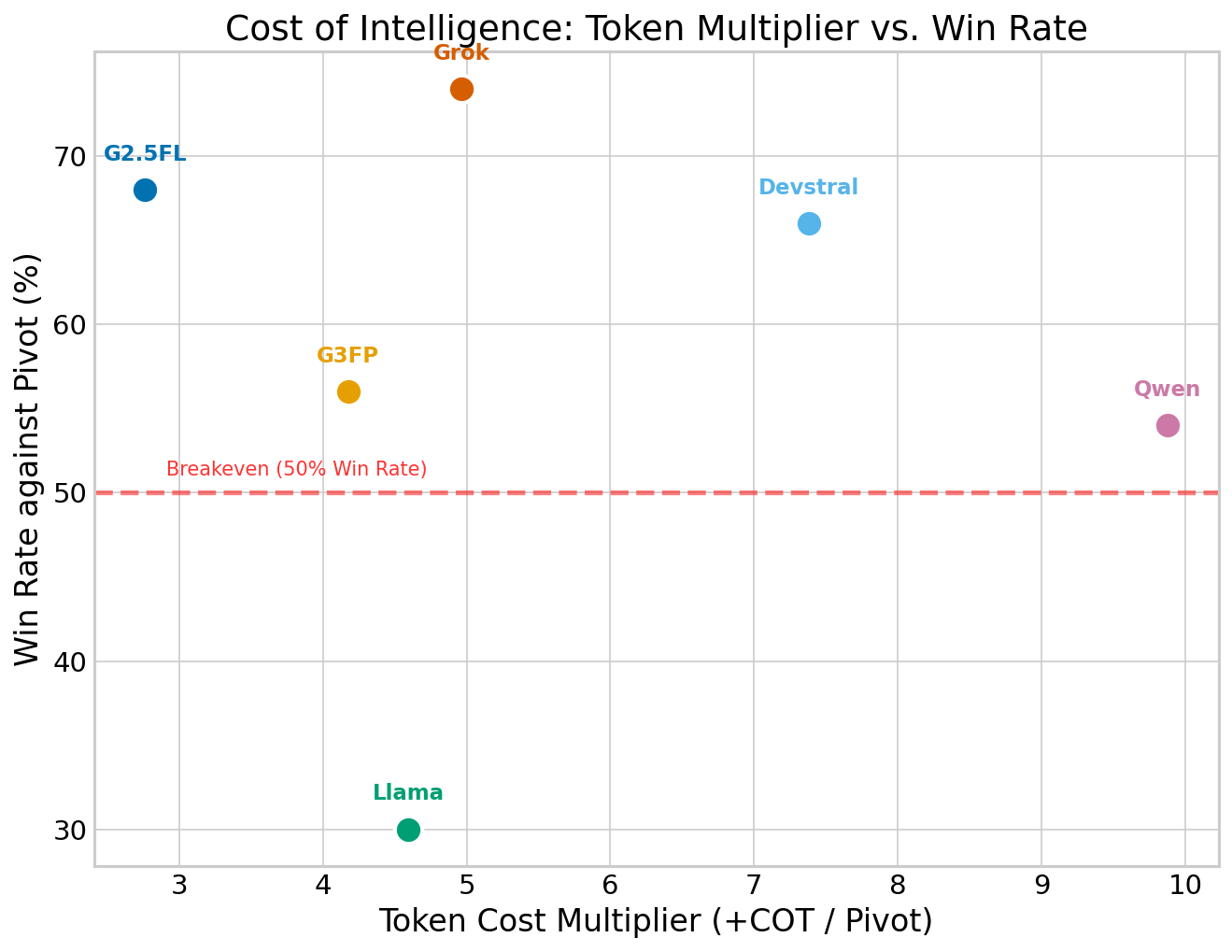}
\caption{Cost vs.\ Win Rate. Plots the token cost multiplier of using +COT against the resulting win rate against the Anchor baseline.}
\Description{Cost vs win rate scatter.}
\label{fig:cost-vs-winrate}
\end{figure}

Figure~\ref{fig:token-profile} visualizes the prompt/completion breakdown. Figure~\ref{fig:token-velocity} shows the
exponential increase in token consumption as deliberation levels are added.
Figure~\ref{fig:cost-vs-winrate} plots the token-cost multiplier
of +COT against its per-instance win rate over the anchor: models where +COT achieves high win rates (Grok,
G2.5FL) pay 4--5$\times$ more tokens, while models where +COT is harmful
(Llama) pay a similar multiplier for worse outcomes.

\begin{table}[htbp]
  \centering
  \caption{Token efficiency: Grok. Shifted return-per-KToken ($\tilde{G}/\mathrm{KTok}$, higher is better), where $\tilde{G}=225+G$ converts the non-positive return $G$ into a non-negative defense score. The most efficient configuration by this scalar is \textbf{bolded} (Pareto efficiency is analyzed in the main text).}
  \label{tab:token-eff-grok}
  \begin{tabular}{l rrr}
    \toprule
    \textbf{Config} & \textbf{Return} & \textbf{Tok/ep} & \textbf{$\tilde{G}$/KTok} \\
    \midrule
    obs & $-98.4$ & 17.7K & 7.1525 \\
    obs+hist & $-89.9$ & 34.6K & 3.9046 \\
    obs+hist+net & $-81.9$ & 33.5K & 4.2716 \\
    obs+net & $-47.0$ & 20.9K & \textbf{8.5167} \\
    network & $-86.3$ & 20.2K & 6.8663 \\
    hist+net & $-112.9$ & 29.1K & 3.8522 \\
    +question & $-66.5$ & 55.6K & 2.8507 \\
    +critique & $-44.9$ & 74.7K & 2.411 \\
    +improve & $-53.4$ & 154.0K & 1.1143 \\
    +COT & $-42.9$ & 144.5K & 1.2602 \\
    hier-base & $-24.0$ & 141.9K & 1.4165 \\
    hier-delib & $-40.4$ & 364.1K & 0.507 \\
    \bottomrule
  \end{tabular}
\end{table}

  \begin{table}[htbp]
  \centering
  \caption{Token efficiency: Llama. Shifted return-per-KToken ($\tilde{G}/\mathrm{KTok}$, higher is better), where $\tilde{G}=225+G$ converts the non-positive return $G$ into a non-negative defense score. The most efficient configuration by this scalar is \textbf{bolded} (Pareto efficiency is analyzed in the main text).}
  \label{tab:token-eff-llama}
  \begin{tabular}{l rrr}
    \toprule
    \textbf{Config} & \textbf{Return} & \textbf{Tok/ep} & \textbf{$\tilde{G}$/KTok} \\
    \midrule
    obs & $-214.7$ & 8.8K & 1.1705 \\
    obs+hist & $-137.4$ & 19.2K & 4.5625 \\
    obs+hist+net & $-102.6$ & 30.6K & 4 \\
    obs+net & $-51.4$ & 12.3K & \textbf{14.1138} \\
    network & $-68.7$ & 13.0K & 12.0231 \\
    hist+net & $-57.1$ & 28.6K & 5.8706 \\
    +question & $-104.1$ & 84.1K & 1.4376 \\
    +critique & $-93.4$ & 97.8K & 1.3456 \\
    +improve & $-75.0$ & 115.3K & 1.301 \\
    +COT & $-100.8$ & 131.3K & 0.9459 \\
    hier-base & $-69.5$ & 87.7K & 1.7731 \\
    hier-delib & $-108.0$ & 158.1K & 0.74 \\
    \bottomrule
  \end{tabular}
\end{table}

  \begin{table}[htbp]
  \centering
  \caption{Token efficiency: Devstral. Shifted return-per-KToken ($\tilde{G}/\mathrm{KTok}$, higher is better), where $\tilde{G}=225+G$ converts the non-positive return $G$ into a non-negative defense score. The most efficient configuration by this scalar is \textbf{bolded} (Pareto efficiency is analyzed in the main text).}
  \label{tab:token-eff-devstral}
  \begin{tabular}{l rrr}
    \toprule
    \textbf{Config} & \textbf{Return} & \textbf{Tok/ep} & \textbf{$\tilde{G}$/KTok} \\
    \midrule
    obs & $-155.1$ & 22.2K & 3.1486 \\
    obs+hist & $-133.9$ & 26.4K & 3.4508 \\
    obs+hist+net & $-85.3$ & 25.1K & 5.5657 \\
    obs+net & $-72.6$ & 15.4K & \textbf{9.8961} \\
    network & $-93.3$ & 14.9K & 8.8389 \\
    hist+net & $-79.3$ & 21.3K & 6.8404 \\
    +question & $-53.9$ & 60.4K & 2.8328 \\
    +critique & $-62.8$ & 106.9K & 1.5173 \\
    +improve & $-80.6$ & 153.1K & 0.9432 \\
    +COT & $-40.9$ & 157.4K & 1.1696 \\
    hier-base & $-37.8$ & 97.0K & 1.9299 \\
    hier-delib & $-127.4$ & 257.7K & 0.3787 \\
    \bottomrule
  \end{tabular}
\end{table}

\begin{table}[htbp]
  \centering
  \caption{Token efficiency: Qwen. Shifted return-per-KToken ($\tilde{G}/\mathrm{KTok}$, higher is better), where $\tilde{G}=225+G$ converts the non-positive return $G$ into a non-negative defense score. The most efficient configuration by this scalar is \textbf{bolded} (Pareto efficiency is analyzed in the main text).}
  \label{tab:token-eff-qwen}
  \begin{tabular}{l rrr}
    \toprule
    \textbf{Config} & \textbf{Return} & \textbf{Tok/ep} & \textbf{$\tilde{G}$/KTok} \\
    \midrule
    obs & $-218.2$ & 10.3K & 0.6602 \\
    obs+hist & $-93.4$ & 19.0K & 6.9263 \\
    obs+hist+net & $-69.0$ & 18.7K & 8.3422 \\
    obs+net & $-63.1$ & 11.4K & \textbf{14.2018} \\
    network & $-109.4$ & 10.6K & 10.9057 \\
    hist+net & $-61.5$ & 16.4K & 9.9695 \\
    +question & $-92.3$ & 45.7K & 2.9037 \\
    +critique & $-93.6$ & 68.9K & 1.9071 \\
    +improve & $-92.4$ & 115.9K & 1.1441 \\
    +COT & $-55.6$ & 162.4K & 1.0431 \\
    hier-base & $-28.6$ & 79.6K & 2.4673 \\
    hier-delib & $-30.1$ & 209.9K & 0.9285 \\
    \bottomrule
  \end{tabular}
\end{table}

\begin{table}[htbp]
  \centering
  \caption{Token efficiency: G2.5FL. Shifted return-per-KToken ($\tilde{G}/\mathrm{KTok}$, higher is better), where $\tilde{G}=225+G$ converts the non-positive return $G$ into a non-negative defense score. The most efficient configuration by this scalar is \textbf{bolded} (Pareto efficiency is analyzed in the main text).}
  \label{tab:token-eff-g25fl}
  \begin{tabular}{l rrr}
    \toprule
    \textbf{Config} & \textbf{Return} & \textbf{Tok/ep} & \textbf{$\tilde{G}$/KTok} \\
    \midrule
    obs & $-214.7$ & 68.1K & 0.1512 \\
    obs+hist & $-172.9$ & 125.2K & 0.4161 \\
    obs+hist+net & $-147.8$ & 94.0K & \textbf{0.8213} \\
    obs+net & $-200.0$ & 79.4K & 0.3149 \\
    network & $-215.4$ & 94.4K & 0.1017 \\
    hist+net & $-208.7$ & 81.7K & 0.1995 \\
    +question & $-206.2$ & 104.8K & 0.1794 \\
    +critique & $-128.6$ & 118.2K & 0.8156 \\
    +improve & $-168.9$ & 182.9K & 0.3067 \\
    +COT & $-157.0$ & 225.7K & 0.3013 \\
    hier-base & $-183.1$ & 120.6K & 0.3474 \\
    hier-delib & $-186.4$ & 270.5K & 0.1427 \\
    \bottomrule
  \end{tabular}
\end{table}

\begin{table}[htbp]
  \centering
  \caption{Token efficiency: G3FP. Shifted return-per-KToken ($\tilde{G}/\mathrm{KTok}$, higher is better), where $\tilde{G}=225+G$ converts the non-positive return $G$ into a non-negative defense score. The most efficient configuration by this scalar is \textbf{bolded} (Pareto efficiency is analyzed in the main text).}
  \label{tab:token-eff-g3fp}
  \begin{tabular}{l rrr}
    \toprule
    \textbf{Config} & \textbf{Return} & \textbf{Tok/ep} & \textbf{$\tilde{G}$/KTok} \\
    \midrule
    obs & $-96.8$ & 8.0K & \textbf{16.025} \\
    obs+hist & $-76.3$ & 15.4K & 9.6558 \\
    obs+hist+net & $-82.6$ & 19.4K & 7.3402 \\
    obs+net & $-113.7$ & 9.0K & 12.3667 \\
    network & $-136.4$ & 7.9K & 11.2152 \\
    hist+net & $-52.0$ & 18.0K & 9.6111 \\
    +question & $-100.6$ & 30.7K & 4.0521 \\
    +critique & $-66.4$ & 41.0K & 3.8683 \\
    +improve & $-64.0$ & 58.1K & 2.7711 \\
    +COT & $-29.9$ & 75.1K & 2.5979 \\
    hier-base & $-16.1$ & 56.4K & 3.7039 \\
    hier-delib & $-23.6$ & 104.8K & 1.9218 \\
    \bottomrule
  \end{tabular}
\end{table}

\begin{table*}[htbp]
  \caption{Token Profile Shift. Average prompt and completion tokens per
    episode for the Anchor baseline vs.\ +COT.}
  \label{tab:token-profile}
  \centering
  \begin{tabular}{l rr|rr | rr|rr}
    \toprule
    & \multicolumn{4}{c}{\textbf{Anchor Baseline}} & \multicolumn{4}{c}{\textbf{+COT}} \\
    \cmidrule(lr){2-5}\cmidrule(lr){6-9}
    \textbf{Model} & \textbf{Prompt} & \textbf{Compl} & \textbf{Total} & \textbf{P/C} & \textbf{Prompt} & \textbf{Compl} & \textbf{Total} & \textbf{P/C} \\
    \midrule
    Grok & 20.7K & 8.4K & 29.1K & 2.5$\times$ & 117.7K & 26.8K & 144.5K & 4.4$\times$ \\
    Llama & 27.0K & 1.6K & 28.6K & 16.9$\times$ & 126.7K & 4.5K & 131.3K & 28.0$\times$ \\
    Devstral & 20.4K & 1.0K & 21.3K & 21.3$\times$ & 152.5K & 4.9K & 157.4K & 31.2$\times$ \\
    Qwen & 15.8K & 0.7K & 16.4K & 23.8$\times$ & 156.8K & 5.6K & 162.4K & 28.1$\times$ \\
    G2.5FL & 75.8K & 5.8K & 81.7K & 13.1$\times$ & 206.8K & 18.9K & 225.7K & 11.0$\times$ \\
    G3FP & 17.2K & 0.7K & 18.0K & 23.5$\times$ & 71.0K & 4.0K & 75.1K & 17.6$\times$ \\
    \bottomrule
  \end{tabular}
\end{table*}

\section{Trajectory Examples}\label{app:trajectories}

We present a paired trajectory comparison from Devstral on
instance~7, illustrating the deliberation cascade failure mode described in
Section~\ref{sec:compound-meta}. Both episodes use identical model weights,
environment seed, and \texttt{hist+net} context. The only difference is whether
deliberation tools are distributed across the hierarchy.

\paragraph{Episode identification.}
The \texttt{hier-base} episode achieves a return of $-13.3$ (near-optimal
defense); the \texttt{hier-delib} episode achieves $-211.2$ (near-total network
loss), a $15.9\times$ degradation on the same environment instance.
Table~\ref{tab:trajectory-summary} summarizes the behavioral contrast.

\begin{table}[htbp]
  \centering
  \caption{Behavioral comparison: Devstral instance~7, \texttt{hier-base} vs.\
    \texttt{hier-delib}. The only architectural difference is the distribution of
    deliberation tools across the hierarchy.}
  \label{tab:trajectory-summary}
  \begin{tabular}{l r r}
    \toprule
    \textbf{Metric} & \textbf{hier-base} & \textbf{hier-delib} \\
    \midrule
    Episode return          & $-13.3$  & $-211.2$ \\
    Restore actions         & 12       & 1 \\
    Remove actions          & 5        & 0 \\
    Decoy deployments       & 0        & 13 \\
    Monitor actions         & 3        & 7 \\
    Delib. tool calls (Planner) & 0        & 11 \\
    AC msgs / invocation    & 2.0      & 6.6 \\
    Analyst invocations     & 14       & 5 \\
    Steps at $\leq\!-13$/step & 0       & 16 \\
    First Restore action    & step 4   & step 25 \\
    \bottomrule
  \end{tabular}
\end{table}

\paragraph{Mechanism: distributed caution bias.}
The trajectories reveal a failure mode we term \emph{distributed caution bias}, a concrete
instantiation of the cascading-uncertainty mechanism discussed in Section~\ref{sec:compound-meta}.
When each agent independently runs a deliberation cycle
(raise\_a\_question $\to$ critique $\to$ improve), each cycle
independently converges on the conservative option: investigate further,
deploy passive defenses, defer reactive intervention. No individual agent's
reasoning is incorrect, the Analyst's hedge is reasonable, the ActionChooser's
caution is defensible, the Planner's self-questioning is thoughtful. But the
\emph{composition} of three independently cautious agents produces a system
that never intervenes until catastrophic damage has accumulated. The failure
is structural: the absence of an urgency-propagation mechanism that would
override local caution when system-level evidence demands immediate action.

\paragraph{Showcase: Step 18 (hier-delib).}
At step~18, the network has been losing $-13.1$ reward per step since step~15
(Op\_Server0 is compromised). The Planner's reasoning trace proceeds as
follows. (Excerpts are lightly trimmed for space)

\begin{verbatim}
[Step 18, hier-delib, reward: -13.1]

Planner (thought): User4 has been analyzed twice
  and has decoys deployed. The other hosts have
  similar setups. The next strategic step is to
  ensure all hosts are consistently protected...
Planner -> raise_a_question: "Should I focus on
  monitoring the network or re-analyzing User4?"
Planner (thought): I need to critique my own
  reasoning to ensure I'm making the best decision.
Planner -> critique_the_answer: "My initial thought
  is to either monitor or re-analyze User4.
  However, I should consider whether there are
  other hosts that might need attention..."
Planner -> improve_based_on_critique: "Monitoring
  the network is the more strategic action. It
  will provide a broader view of potential threats
  and ensure all hosts are observed for anomalies."
Planner -> get_suggestion_for_next_action:
  {"target_host": "User4", "severity": "low", ...}
  ActionChooser (8 msgs, own deliberation cycle):
    -> raise_a_question: "What would be the
       appropriate strategy for User4?"
    -> critique: "Have I considered the potential
       need for reactive interventions?"
    -> improve: "Balance passive and proactive..."
    Returns: [Monitor (high), Analyse (medium),
              Restore (low)]
Planner (final): Answer: Monitor
\end{verbatim}

The Planner uses three deliberation tools before consulting the
ActionChooser. The ActionChooser runs its own three-phase deliberation cycle.
Ten messages are exchanged to arrive at "Monitor", a passive action, while
the network loses 13.1 reward points. The Analyst is not consulted at all
at this step.

\paragraph{Contrast: Step 18 (hier-base).}
In the same episode phase, the \texttt{hier-base} agent operates with a clean
network (all hosts at baseline, reward $= 0.0$) because it contained threats
via timely Restore actions at steps 4, 6, 8, 11, and 14. Its step-18 trace
is four messages:

\begin{verbatim}
[Step 18, hier-base, reward: 0.0]

Planner (thought): Network is stable, all hosts in
  baseline state. Enterprise2 was recently restored;
  analyzing it again could confirm its integrity.
Planner -> get_suggestion_for_next_action:
  {"target_host": "Enterprise2",
   "situation": "restored, confirm integrity",
   "severity": "low", ...}
  ActionChooser (2 msgs): Returns:
    [Analyse (high), DecoySSHD (medium),
     Restore (low)]
Planner (final):
  Answer: Analyse hostname=Enterprise2
\end{verbatim}

No deliberation tools are invoked. The Planner identifies the most useful
proactive action and executes it in a single delegation round. The
ActionChooser returns its recommendation in one exchange without a critique
cycle.

\paragraph{The critical divergence.}
The behavioral divergence begins at step~7. Both systems detect changes on
Enterprise0 and User4. The \texttt{hier-base} Planner, at steps~3--8, follows
a tight detect--analyse--restore cycle: upon finding C2 indicators (connection
to port~4444, backdoor process), it immediately requests suggestions and acts
on the high-confidence "Restore" recommendation. The \texttt{hier-delib}
Planner, facing the same situation, invokes \texttt{include\_tool\_raise\_a\_question}
("What is the most strategic action?"),\\
 then \texttt{include\_tool\_critique\_the\_answer}
("Should I deploy a DecoySSHD or prioritize Enterprise0?"), and resolves
toward prophylactic decoy deployment rather than reactive containment.
By step~14, the \texttt{hier-base} system has executed 5~Restore and
3~Remove actions; the \texttt{hier-delib} system has executed
0~Restore, 0~Remove, and 8~Decoy deployments. The reward gap at step~14 is
$-6.3$ vs.\ $-14.2$.

The gap becomes irreversible at step~15, when the red agent escalates to
high-value targets. The \texttt{hier-delib} system, having deployed passive
defenses rather than containing active compromise, faces a $-13.1$/step
penalty that persists for the remaining 16~steps. It does not execute its
first (and only) Restore until step~25.

\section{CybORG CAGE-2 Environment Details}\label{app:cage2}

CybORG CAGE-2~\cite{cage2,standen2021cyborg} models an autonomous
network-defense scenario used as Challenge~2 of the TTCP CAGE
(Cyber Autonomy Gym for Experimentation) series. We summarize the key
environment characteristics; full specifications and source code are available
at \url{https://github.com/cage-challenge/cage-challenge-2}.

\paragraph{Network topology.}
The simulated network comprises 13~hosts organized into three subnets:
\emph{User} (User0--User4, 5~hosts), \emph{Enterprise} (Enterprise0--Enterprise2
plus a single Enterprise Server, 4~hosts), and \emph{Operational}
(Op\_Host0--Op\_Host2, Op\_Server0, 4~hosts). Traffic flows from an Internet-facing subnet through User to Enterprise
to Operational. Op\_Server0 is the attacker's ultimate high-value target.

\paragraph{Attacker (Red agent).}
The red agent follows a scripted kill chain that progresses through five
stages: \emph{Discovery} (scanning for reachable hosts),
\emph{Access} (exploiting vulnerabilities to gain initial foothold),
\emph{Lateral Movement} (pivoting across subnets),
\emph{Privilege Escalation} (gaining root/admin access), and
\emph{Impact} (compromising high-value targets for maximum penalty).
The attacker advances one stage per step and selects
new targets based on reachability. The scripted policy is fixed
across all experiments; no adaptive adversary is used.

\paragraph{Defender actions.}
The blue agent (defender) selects one action per step from a five-type
repertoire:

\begin{description}
\item[Monitor] (free, observational): Observe the network. Returns the
  current observation with no side effects.
\item[Analyse \texttt{hostname=X}] (free, diagnostic): Deeper inspection
  of a host. Returns process- and connection-level details that may reveal
  compromise indicators not visible in the standard observation.
\item[Remove \texttt{hostname=X}] (attempted cleanup): Attempt to remove
  malicious artifacts from a host. May fail against persistent threats,
  leaving the host in an "unknown" state.
\item[Restore \texttt{hostname=X}] (guaranteed cleanup, high penalty):
  Reset the host to its known-good baseline. Guaranteed to succeed, but incurs
  a large one-step penalty and removes all deployed decoys from that host.
\item[Decoy\textit{SERVICE} \texttt{hostname=X}] (proactive): Deploy a
  decoy service (e.g., \texttt{DecoySSHD}, \texttt{DecoyApache},
  \texttt{DecoyTomcat}) on a host to attract and detect attacker activity
  in future steps. Multiple decoy types can be stacked on a single host.
\end{description}

\paragraph{Reward structure.}
The per-step reward $r_t \leq 0$ penalizes each compromised host according
to its value tier; Operational hosts carry the highest penalties.
Restore incurs an additional fixed penalty at the step it is applied.
The maximum cumulative return (perfect defense) is $0$; all configurations
operate in the negative-return regime.

\paragraph{Episode structure.}
Each episode runs for $T = 30$ steps. The attacker begins with access to
the Internet-facing subnet and progresses inward. The defender observes
partial, noisy information and must infer compromise from limited signals.
Early intervention is rewarded because attacker progress compounds:
a compromised User host enables lateral movement to Enterprise, then to
Operational targets where penalties are highest.

\section{Episode Counts}\label{app:episode-counts}

Table~\ref{tab:episode-counts} reports the number of evaluated episodes for
each of the 72 model--configuration pairs. The standard allocation is
$10$~instances $\times$ $5$~runs $= 50$~episodes per pair. G3FP uses a
reduced default of $5 \times 5 = 25$~episodes per configuration due to staged
data collection. Several configurations include extended batches (marked with
$\dagger$) to reduce uncertainty on key comparisons: G2.5FL \texttt{obs+net}
(100~episodes), Qwen \texttt{hier-base} (100~episodes), and G2.5FL
\texttt{hier-base} (75~episodes). The total across all models and
configurations is 3{,}475~episodes (104{,}250 agent--environment interaction
steps, consuming 283.9M tokens).

\begin{table}[htbp]
  \centering
  \caption{Episode counts per model--configuration pair. Standard allocation
    is 10 instances $\times$ 5 runs = 50 episodes per configuration.}
  \label{tab:episode-counts}
  \footnotesize
  \begin{tabular}{l l r r r r r r}
    \toprule
    \textbf{Group} & \textbf{Config} & \textbf{Grok} & \textbf{Ll} & \textbf{Devs} & \textbf{Qwen} & \textbf{G2.5FL} & \textbf{G3FP} \\
    \midrule
    Context & obs & 50 & 50 & 50 & 50 & 50 & 25 \\
     & obs+hist & 50 & 50 & 50 & 50 & 50 & 25 \\
     & obs+hist+net & 50 & 50 & 50 & 50 & 50 & 25 \\
     & obs+net & 50 & 50 & 50 & 50 & 100$^\dagger$ & 25 \\
     & network & 50 & 50 & 50 & 50 & 50 & 25 \\
     & hist+net & 50 & 50 & 50 & 50 & 50 & 25 \\
    Delib. & +question & 50 & 50 & 50 & 50 & 50 & 50 \\
     & +critique & 50 & 50 & 50 & 50 & 50 & 25 \\
     & +improve & 50 & 50 & 50 & 50 & 50 & 25 \\
     & +COT & 50 & 50 & 50 & 50 & 50 & 25 \\
    Hier & hier-base & 50 & 50 & 50 & 100$^\dagger$ & 75$^\dagger$ & 25 \\
     & hier-delib & 50 & 50 & 50 & 50 & 50 & 50 \\
    & \textbf{Total} & \textbf{600} & \textbf{600} & \textbf{600} & \textbf{650} & \textbf{675} & \textbf{350} \\
    \bottomrule
  \end{tabular}

  \smallskip
  \centering\footnotesize $^\dagger$Configuration includes additional evaluation
  batches beyond the standard allocation.
\end{table}

\end{document}